%% file: main.tex
\documentclass[sn-mathphys-num]{sn-jnl}

\usepackage{graphicx}%
\usepackage{multirow}%
\usepackage{amsmath,amssymb,amsfonts}%
\usepackage{amsthm}%
\usepackage{mathrsfs}%
\usepackage[title]{appendix}%
\usepackage{xcolor}%
\usepackage{textcomp}%
\usepackage{manyfoot}%
\usepackage{booktabs}%
\usepackage{algorithm}%
\usepackage{algorithmicx}%
\usepackage{algpseudocode}%
\usepackage{listings}%

\input{utils.tex}
\usepackage{soul} 

\newcommand \cont{\textit{CONT}}
\newcommand \lcont{\textit{L-CONT}}
\newcommand \sconv{\textit{S-CONV}}
\newcommand \diff{\textit{DIFF}}
\newcommand \conv{\textit{CONVEX}}
\newcommand{\svec}[1]{\{\ensuremath{\vec{#1}_0, \cdots, \vec{#1}_N}\}}

\newcommand{\vtheta}{{\ensuremath{\boldsymbol{\theta}}}}
\DeclareUnicodeCharacter{2212}{-}
\renewcommand\vec{\mathbf}
\algrenewcommand\algorithmicrequire{\textbf{Input:}}
\algrenewcommand\algorithmicensure{\textbf{Output:}}

\begin{document}

\title[A survey and taxonomy of loss functions in machine learning]{A survey and taxonomy of loss functions in machine learning}


\author*[1]{\fnm{Lorenzo} \sur{Ciampiconi}}\email{lorenzo.ciampiconi@lastminute.com}
\equalcont{These authors contributed equally to this work.}

\author[2]{\fnm{Adam} \sur{Elwood}}\email{adam.elwood09@gmail.com}

\author[1]{\fnm{Marco} \sur{Leonardi}}\email{marco.leonardi@lastminute.com}
\equalcont{These authors contributed equally to this work.}

\author[2]{\fnm{Ashraf} \sur{Mohamed}}\email{ashrafkasem.asu@gmail.com}

\author[1]{\fnm{Alessandro} \sur{Rozza}}\email{alessandro.rozza@lastminute.com}
\equalcont{These authors contributed equally to this work.}

\affil*[1]{\orgdiv{lastminute.com}, \orgname{group}, \orgaddress{\street{Vicolo de' Calvi 2}, \city{Chiasso}, \postcode{6830}, \state{Switzerland}
}}
\affil*[2]{Work done while working at lastminute.com group }



\abstract{Most state-of-the-art machine learning techniques revolve around the optimisation of loss functions. Defining appropriate loss functions is therefore critical to successfully solving problems in this field.  In this survey, we present a comprehensive overview of the most widely used loss functions across key applications, including regression, classification, generative modeling, ranking, and energy-based modeling. We introduce 43 distinct loss functions, structured within an intuitive taxonomy that clarifies their theoretical foundations, properties, and optimal application contexts. This survey is intended as a resource for undergraduate, graduate, and Ph.D. students, as well as researchers seeking a deeper understanding of loss functions.}





\maketitle

\section{Introduction}
    In the last few decades there has been an explosion in interest in machine learning \cite{ML_trends,ML_trends2}. This field focuses on the definition and application of algorithms that can be trained on data to model underlying patterns \cite{ML_2,ML_1,decision_trees,bishop1995neural}. 
    Machine learning approaches can be applied to many different research fields, including biomedical science \cite{survey_biomed,ml_biomed,brain_interface_survey,cancer_ml}, natural language understanding \cite{nlp_1,nlp_2}, \cite{computer_security_survey} anomaly detection \cite{anomaly_detection_survey}, image classification \cite{image_classification_survey}, database knowledge discovery \cite{frawley1992knowledge}, robot learning \cite{argall2009survey}, online advertising \cite{online_advertising}, time series forecasting \cite{time_series_forecasting_ml}, brain-computer interfacing \cite{brain_computer_interfaces} and many more \cite{other_applications}.
    To train these algorithms, it is necessary to define an objective function, which gives a scalar measure of the algorithm's performance \cite{von2007theory,ML_2}. They can then be trained by optimising the value of the objective function. 
    
    Within the machine learning literature, such objective functions are usually defined in the form of loss functions, which are optimal when they are minimised. The exact form of the loss function depends on the nature of the problem to be solved, the data available and the type of machine learning algorithm being optimised. Finding appropriate loss functions is, therefore, one of the most important research endeavors in machine learning. 

    As machine learning has progressed, many loss functions have been introduced to address various tasks and applications. Summarizing and understanding these functions is essential, yet there are few works that attempt a comprehensive overview of loss functions across the entire field~\cite{wang2020comprehensive}. Existing reviews typically either lack a robust taxonomy to effectively structure and contextualize these functions or focus narrowly on specific applications, such as image segmentation or classification~\cite{wang2021survey, jadon2020survey}. Additionally, there is no single source that presents the most commonly used loss functions within a unified formal setting, providing detailed insights into their eventual advantages, limitations, and use cases.
    
    
    For this reason, we have worked to build a proper taxonomy of loss functions, where we show the advantages and disadvantages of each technique. We hope this will be useful for new users who want to familiarise themselves with the most common loss functions used in the machine learning literature and find one that is suitable for a problem that they are trying to solve. We also hope this summary will be useful as a comprehensive reference for advanced users, allowing them to quickly find the best loss function without having to broadly search the literature.  Additionally, this can be helpful for researchers to find possible avenues for further research, or to understand where to place any new techniques that they have proposed. They could, for example, use this survey to understand if their new proposals fit somewhere inside the taxonomy we present, or if they are in a completely new category, maybe combining disparate ideas in novel ways. 
    
    Overall, we have included 43 of the most widely used loss functions. In each section of this work, we break down the losses based on the broad classification of tasks that they can be used for. Each loss function will be defined mathematically, and its most common applications listed highlighting advantages and drawbacks. 
    
    The main contribution of this work can be found in the proposed taxonomy depicted in Fig.~\ref{fig:taxionomy}. Each loss function is first divided according to the specific task on which they are exploited: regression, classification, ranking, generative and energy-based modelling.
  
    Finally, we classify each loss function by its underlying strategy, such as \textbf{error minimization}, \textbf{probabilistic formalization}, or \textbf{margin maximization}.

    \bigskip
     This work is organized as follows: In Section \ref{sec:definition}, we provide a formal definition of a loss function and introduce our taxonomy. In Section \ref{sec:regularisation}, we describe the most common regularization methods used to reduce model complexity. In Section \ref{sec:regression}, we describe the regression task and the key loss functions used to train regression models. In Section \ref{sec:classification}, we introduce the classification problem and the associated loss functions. In Section \ref{sec:generative}, we present generative models and their losses. Ranking problems and their loss functions are introduced in Section \ref{sec:ranking}, and energy based models and their losses are described in Section \ref{sec:ebm}. Finally, we draw conclusions in Section \ref{sec:conclusion}.

\section{Definition of our loss function taxonomy} \label{sec:definition}

    In a general machine learning problem, the aim is to learn a function $f$ that transforms an input, defined by the input space $\Phi$ into a desirable output, defined by the output space $\mathcal{Y}$:
    
    \begin{align*}
        f \colon &\Phi \to \mathcal{Y}
    \end{align*}
    
    Where $f$ is a function that can be approximated by a model, $f_\Theta$, parameterised by the parameters $\vec{\Theta}$.
    
    Given a set of inputs $\{\vec{x}_0, ...,\vec{x}_N\} \in \Phi$, they are used to train the model with reference to target variables in the output space, $\{\vec{y}_0, ...,\vec{y}_N\} \in \mathcal{Y}$. Notice that, in some cases (such as autoencoders) $\mathcal{Y} = \Phi$. 
    
    A loss function, $L$, is defined as a mapping of $f(\vec{x}_i)$ with it's corresponding $\vec{y}_i$ to a real number $l\in \mathbb{R}$, which captures the similarity between $f(\vec{x}_i)$ and $\vec{y}_i$. Aggregating over all the points of the dataset we find the overall loss, $\mathcal{L}$:
    \begin{align}
        \mathcal{L}(f |\{\vec{x}_0, ...,\vec{x}_N\},\{\vec{y}_0, ...,\vec{y}_N\} )  = \frac{1}{N} \sum_{i=1}^N L(f(\vec{x}_i),\vec{y}_i)
    \end{align}
    The optimisation function to be solved is defined as:  
    \begin{align}
        \min_{f} \mathcal{L}(f|\{\vec{x}_0, ...,\vec{x}_N\},\{\vec{y}_0, ...,\vec{y}_N\})
    \end{align}
    
    Notice that, it is often convenient to explicitly introduce a regularisation term ($R$) which maps $f$ to a real number $r\in \mathbb{R}$. This term is usually used for penalising the complexity of the model in the optimisation \cite{ML_2}:
    
    \begin{align}
        \min_{f} \frac{1}{N} \sum_{i=1}^N L(f(\vec{x}_i),\vec{y}_i) + R(f)
    \end{align}
    
    In practice, the family of functions chosen for the optimisation can be parameterised by a parameter vector $\vec{\Theta}$, which allows the minimisation to be defined as an exploration in the parameter space:
    \begin{align}\label{loss_regularised}
        \min_{\vec{\Theta}} \frac{1}{N} \sum_{i=1}^N L(f_\vec{\Theta}(\vec{x}_i),\vec{y}_i) + R(\vec{\Theta})
    \end{align}

\subsection{Optimisation techniques for loss functions}

\subsubsection{Loss functions and optimisation methods}
  In this section, we list out the most common mathematical properties that a loss may or may not satisfy and then we briefly discuss the main optimisation methods employed to minimise them. For the sake of simplicity, visualisation and understanding we define such properties in a two-dimensional space, but they can be easily generalised to a d-dimensional one.
 
 \begin{itemize}
    \item \textbf{Continuity} (\cont): A real function, that is a function from real numbers to real numbers, can be represented by a graph in the Cartesian plane; such a function is continuous if the graph is a single unbroken curve belonging to the real domain. A more mathematically rigorous definition can be given by defining continuity in terms of limits. A function $f$ with variable $x$ is continuous at the real number $c$, if $\lim_{x \to c} f(x) = f(c)$.
    \item \textbf{Differentiability} (\diff): A differentiable function $f$ on a real variable is a function derivable in each point of its domain. A differentiable function is smooth (the function is locally well approximated as a linear function at each interior point) and does not contain any break, angle, or cusp. A continuous function is not necessarily differentiable, but a differentiable function is necessarily continuous.
    \item \textbf{Lipschitz Continuity} (\lcont): A Lipschitz continuous function is limited in how fast it can change. More formally, there exists a real number such that, for every pair of points on the graph of this function, the absolute value of the slope of the line connecting them is not greater than this real number; this value is called the Lipschitz constant of the function. 
    
    To understand the robustness of a model, such as a neural network, some research papers \cite{virmaux2018lipschitz,gouk2021regularisation} have tried to train the underlying model by defining an input-output map 
    with a small Lipschitz constant. The intuition is that if a model is robust, it should not be too affected by perturbations in the input, $f(x + \delta x) \approx f(x) $, and this would be ensured by having $f$ be $\ell$-Lipschitz where $\ell$ is small \cite{adversarial_robustness_lipschitz_bound}.
    \item \textbf{Convexity} (\conv): a real-valued function $f$ is \textit{convex} if each segment between any two points on the graph of the function lies above the graph between the two points.
    Convexity is a key feature since the local minima of the convex function is also the global minima. Whenever the second derivative of a function exists, then the convexity is easy to check, since the Hessian of the function must be positive semi-definite.
    \item \textbf{Strict Convexity} (\sconv): a real-valued function is \textit{stricly convex} if the segment between any two points on the graph of the function lies above the graph between the two points, except for the intersection points between the straight line and the curve. Strictly convex functions have a positive definitive Hessian.
    Positive-definite matrices are invertible and the optimisation problem can be so solved in a closed form.
\end{itemize}

\subsubsection{Relevant Optimisation Methods}
An optimisation method is a technique that, given a formalised optimisation problem with an objective function, returns the solution to obtain the optimal value of that optimisation problem. Most of the optimisation methods presented in this work rely on algorithms that may not guarantee the optimality of the solution, but imply a degree of approximation.

For each optimisation method, we specify the mathematical properties that the loss function must satisfy, such as continuity, differentiability, or convexity. These requirements are listed in the headings of each method, using \textcolor{blue}{blue} to indicate the necessary properties for the method's usability and \textcolor{red}{red} for properties that provide optimality guarantees.

\begin{itemize}
    \item \textbf{Closed-Form Solutions (\textcolor{blue}{\diff}, \textcolor{blue}{\sconv})}: 
These are systems of equations solvable analytically, where values of $\vec{\Theta}$ make the derivative of the loss function equal to zero. To guarantee a unique closed-form solution, the loss function must be differentiable (\textcolor{blue}{\diff}) and strictly convex (\textcolor{blue}{\sconv}), ensuring a single global minimum. Closed-form solutions are highly efficient and desirable where feasible; however, they are often impractical for complex models or high-dimensional parameter spaces. Therefore, closed-form solutions are primarily used in simpler, linear models or settings where the loss is quadratic or log-likelihood based, as in linear regression or Gaussian MLE problems.

    \item \textbf{Gradient Descent (\textcolor{blue}{\diff}, \textcolor{red}{\conv})}:
    Gradient Descent is a first-order iterative optimization algorithm used to find a local minimum of a differentiable function. The loss function must be at least differentiable (\textcolor{blue}{\diff}) to compute gradients, and if the loss is convex (\textcolor{red}{\conv}), the local minimum is also the global minimum. Lipschitz continuity (\lcont) can improve convergence guarantees, as it limits how quickly the function can change, but it is not strictly necessary for Gradient Descent. For non-differentiable losses, techniques like subgradients or gradient approximations can be employed \cite{gd_approx_1,gd_approx_2}. The algorithm for Gradient Descent is formalized in Algorithm~\ref{alg:deep-learning-gradient-descent_a}.
    
    \begin{algorithm}[t]
        \caption{Gradient Descent}\label{alg:deep-learning-gradient-descent_a}
	\begin{algorithmic}[1]
    	\Require{initial parameters $\vec{\Theta}^{(0)}$, number of iterations $T$, learning rate $\alpha$}
    	\Ensure{final learning $\vec{\Theta}^{(T)}$}
            \State $t = 0$ 
            \While{$t < T$}
                \State estimate $\nabla \mathcal{L}(\vec{\Theta}^{(t)})$
                \State compute $\Delta \vec{\Theta}^{(t)} = - \nabla \mathcal{L}(\vec{\Theta}^{(t)})$
                \State $\vec{\Theta}^{(t + 1)} := \vec{\Theta}^{(t)} + \alpha \Delta \vec{\Theta}^{(t)}$
            \EndWhile	
    	\State \textbf{return} $\vec{\Theta}^{(T)}$
	\end{algorithmic}
    \end{algorithm}

    \item \textbf{Stochastic Gradient Descent (SGD) (\textcolor{blue}{\diff}, \textcolor{red}{\conv})}:
    SGD \cite{ML_2} is a stochastic approximation of Gradient Descent that computes the gradient from a randomly selected subset of the data instead of the entire dataset. This reduces the computational cost in high-dimensional problems, such as neural networks, and helps avoid local minima due to the stochastic nature of the updates. Like Gradient Descent, SGD requires the loss function to be differentiable (\textcolor{blue}{\diff}). Convexity (\textcolor{red}{\conv}) ensures that the global minimum is reachable, but even for non-convex functions, SGD can often find useful minima in practice. Lipschitz continuity (\lcont) can improve the convergence rate, but is not required.

    \item \textbf{Derivative-Free Optimisation}: 
In some cases, the derivative of the objective function may not exist or be difficult to compute. Derivative-Free Optimisation methods, such as simulated annealing, genetic algorithms, and particle swarm optimisation, can be employed \cite{derivative_free_opt_1,derivative_free_opt_2}. While these methods do not strictly require continuity (\textcolor{blue}{\cont}), having a continuous function typically improves the stability of the optimization process. Derivative-free methods can handle non-differentiable and non-convex losses, but they may struggle to scale to high-dimensional problems and can be computationally expensive.

\item \textbf{Zeroth-Order Optimisation (ZOO)}: 
ZOO optimisation is a subset of derivative-free optimisation that approximates gradients using function evaluations rather than direct computation of derivatives \cite{zeroth_optimisation_3}. These methods are useful in black-box scenarios where the gradient is not accessible but can be estimated through perturbations. While continuity (\textcolor{blue}{\cont}) is not required, it improves the accuracy of gradient approximations and helps achieve better convergence rates. ZOO methods are effective for non-differentiable and non-convex losses, and they have been applied in adversarial attack generation, model-agnostic explanations, and other black-box scenarios \cite{zeroth_optimisation_1,zeroth_optimisation_2}.

\end{itemize}

    \begin{figure}[H]
        \centering
        \includegraphics[width=\linewidth]{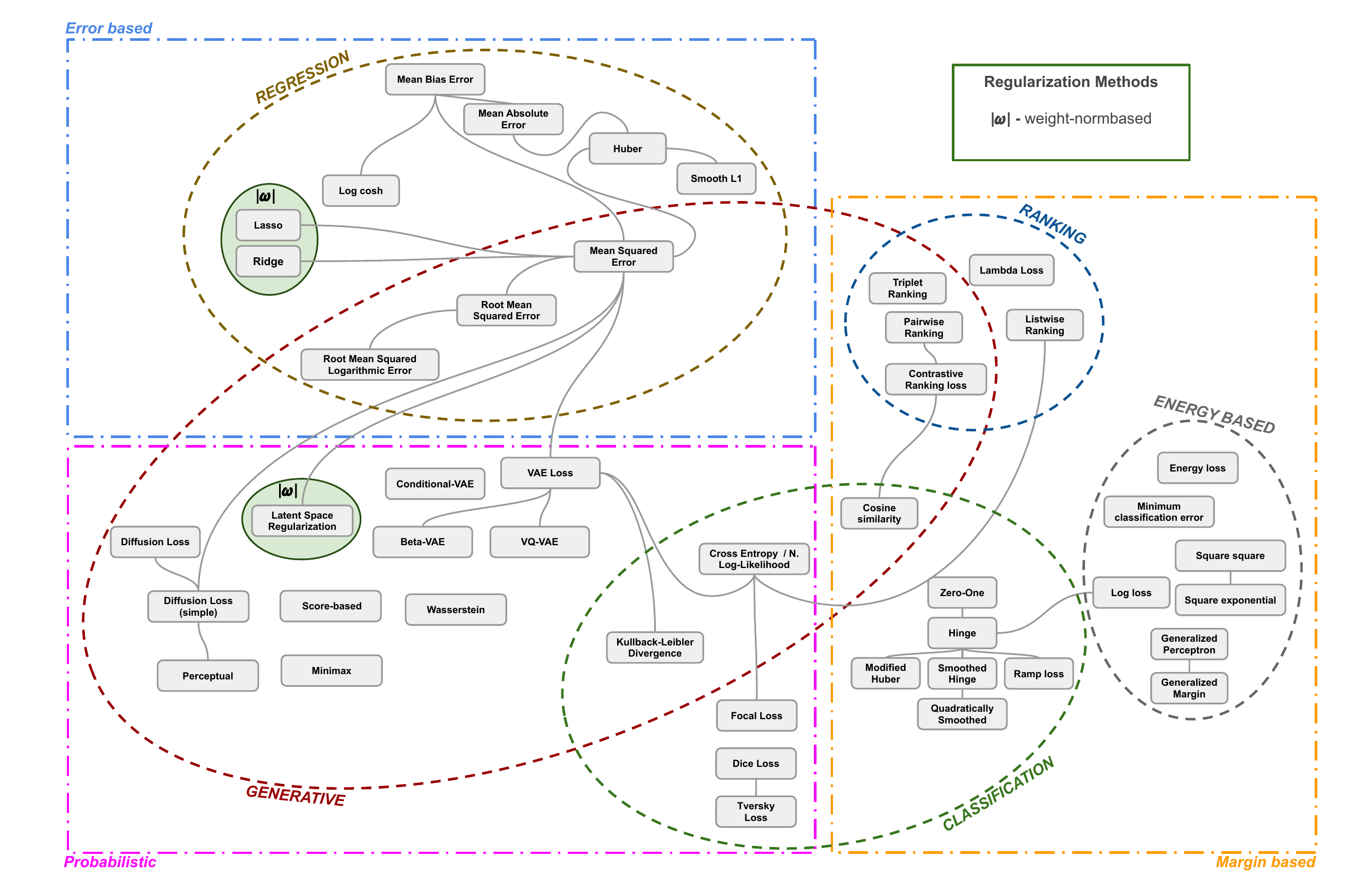}
    \caption{The proposed taxonomy. Five major tasks are identified on which loss functions are applied, namely regression, classification, ranking, generative and energy-based modeling. Finally, the underlying strategy to optimize them, namely margin-based, probabilistic, and error-based, is illustrated under each group of losses.}
    \label{fig:taxionomy}
    \end{figure}

    \subsection{Our taxonomy}
    
    Our taxonomy is summarized in Fig~\ref{fig:taxionomy}. To define it, we started by categorizing the losses depending on which machine learning problem they are best suited to solve. 
    We have identified the following categories:
    \begin{itemize}
        \item Regression (Sec.~\ref{sec:regression})
        \item Classification (Sec.~\ref{sec:classification})
        \item Generative modelling (Sec.~\ref{sec:generative})
        \item Ranking (Sec.~\ref{sec:ranking})
        \item Energy-based modelling (Sec.~\ref{sec:ebm})
    \end{itemize}  
    We also made a distinction based on the mathematical concepts used to define the loss obtaining the following sub-categories:
    \begin{itemize}
        \item Error based 
        \item Probabilistic
        \item Margin based
    \end{itemize}
    
    
    Using this approach, we developed a compact and intuitive taxonomy. We employed well-established terminology to ensure that users can intuitively navigate and understand the taxonomy.

  %
        

\section{Regularisation methods}    \label{sec:regularisation}
Regularisation methods can be applied to almost all loss functions. They are employed to reduce model complexity, simplifying the trained model and reducing its propensity to overfit the training data \cite{regularisation-cambridge-book,regularisation-for-deep-learning,model-selection-and-error-estimation}. Model complexity, is usually measured by the number of parameters and their magnitude \cite{model-selection-and-error-estimation,importance-of-complexity-in-model-selection,ML_2}. Many techniques fall under the umbrella of regularisation method and a significant number of them are based on the augmentation of the loss function \cite{regularisation-cambridge-book,ML_2}. An intuitive justification for regularization is that it imposes Occam's razor on the complexity of the final model. More theoretically, many loss-based regularization techniques are equivalent to imposing certain prior distributions on the model parameters.

\subsection{Regularisation by Loss Augmentation}
One can design the loss function to penalise the magnitude of model parameters, thus learning the best trade-off between bias and variance of the model and reducing the generalization error without affecting the training error too much. This prevents overfitting, while avoiding underfitting, and can be done by augmenting the loss function with a term that explicitly controls the magnitude of the parameters, or implicitly reduces the number of them. 
The general way of augmenting a loss function to regularise the result is formalized in the following equation:
\begin{equation}
\widehat{L}(f(\vec{x}_i),\vec{y}_i)  = L(f(\vec{x}_i),\vec{y}_i) + \lambda \rho(\vec{\Theta})
\end{equation}
where $\rho(\vec{\Theta})$ is called regularization function and $\lambda$ defines the amount of regularisation (the trade-off between fit and generalisation).

This general definition makes it clear that we can employ regularization on any of the losses proposed in this paper.

We are now going to describe the most common regularisation methods based on loss augmentation.
\subsubsection{L2-norm regularisation}\label{l2_norm_regularisation}
In $L_2$ regularization the loss is augmented to include the weighted $L_2$ norm of the weights \cite{ML_2,ML_3}, so the regularisation function is $\rho(\vec{\Theta}) = \norm{\vec{\Theta}}^2_2$:
\begin{equation}\label{L2-reg-eq}
\widehat{L}(f(\vec{x}_i),\vec{y}_i)  = L(f(\vec{x}_i),\vec{y}_i) + \lambda \norm{\vec{\Theta}}^2_2
\end{equation}
when this is employed in regression problems it is also known as Ridge regression \cite{ridge_regression,ML_2}. 

\subsubsection{$L_1$-norm regularisation}\label{L1-reg}
In $L_1$ regularization the loss is augmented to to include the weighted $L_1$ norm of the weights \cite{ML_2,ML_3}, so the regularisation function is $\rho(\vec{\Theta}) = \norm{\vec{\Theta}}$
\begin{equation}\label{L1-reg-eq}
\widehat{L}(f(\vec{x}_i),\vec{y}_i)  = L(f(\vec{x}_i),\vec{y}_i) + \lambda \norm{\vec{\Theta}}_1
\end{equation}
when this is employed in regression problems it is also known as Lasso regression \cite{lasso_regression,ML_2}.

\subsection{Comparison between $L_2$ and $L_1$ norm regularisations}
$L_1$ and $L_2$ regularisations are both based on the same concept of penalising the magnitude of the weights composing the models. Despite that, the two methods have important differences in their employability and their effects on the result. 

One of the most crucial differences is that $L_1$, when optimised, can shrink weights to 0, while $L_2$ results in non-zeros (smoothed) values \cite{ML_1,ML_2,ML_3,andrew-ng-l1-l2,l1_vs_l2}. This allows $L_1$ to reduce the dimension of a model's parameter space and perform an implicit feature selection. Indeed, it has been shown by \cite{andrew-ng-l1-l2} that by employing $L_1$ regularization on logistic regression, the sample complexity (i.e., the number of training examples
required to learn “well”) grows logarithmically in the number of irrelevant features.
On the contrary, the authors show that any rotationally invariant algorithm (including logistic regression) with $L_2$ regularization has a worst-case sample complexity that grows at least linearly in the number of irrelevant features.
Moreover, $L_2$ is more sensitive to the outliers than $L_1$-norm since it squares the error.

$L_2$ is continuous, while $L_1$ is a piece-wise function.
The main advantage of $L_2$ is that it is differentiable, while  $L_1$ is non-differentiable at $0$, which has some strong implications. 
Precisely, the $L_2$ norm can be easily trained with gradient descent, while $L_1$ sometimes cannot be efficiently applied.
The first problem is the inefficiency of applying the $L_1$ penalty to the weights of all the features, especially when the dimension of the feature space tends to be very large \cite{l1_sgd},
producing a significant slowdown of the weights updating process. Finally, the naive application of $L_1$ penalty in SGD does not always lead to compact models, because the approximate gradient used at each update could be very noisy, so the weights of the features can be easily moved away from zero by those fluctuations and $L_1$ looses its main advantages for $L_2$ \cite{l1_sgd}.

\section{Regression losses} \label{sec:regression}
\subsection{Problem Formulation and Notation}
    A regression model aims to predict the outcome of a continuous variable $y$ (the dependent variable) based on the value of one or multiple predictor variables $\vec{x}$ (the independent variables). 
    More precisely, let $f_{\Theta}$ be a generic model parameterized by $\vec{\Theta}$, which maps the independent variables $\vec{x}\in \{\vec{x}_0,...,\vec{x}_N\}, \vec{x}_i \in \mathbb{R}^D$ into the dependent variable $y\in \mathbb{R}$.
    The final goal is to estimate the parameters of the model $\vec{\Theta}$ that most closely fits the data by minimizing a loss function $\mathcal{L}$.
    
    \begin{figure}[H]
        \centering
        \includegraphics[width=0.6\linewidth]{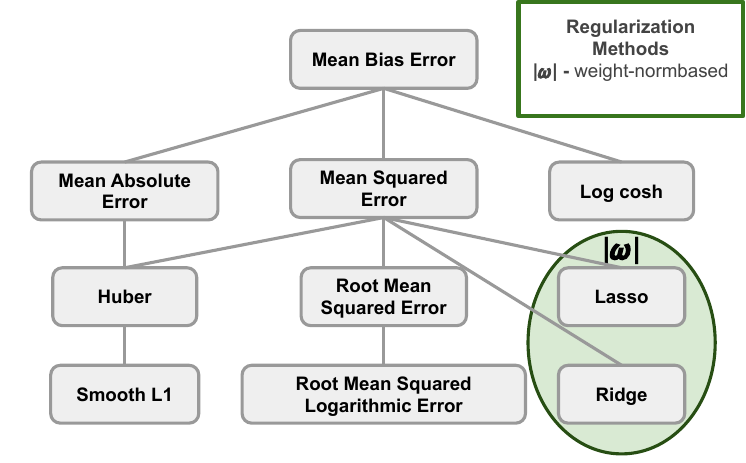}
        \caption{Schematic overview of the regression losses showing the connection }
        \label{fig:regression_losses_focus}
    \end{figure}
    All the losses considered for the regression task are based on functions of the residuals, i.e. the difference between the observed value $y$ and the predicted value $f(\vec{x})$. 
    In the following, let $f(\vec{x}_i)$ be the outcome of the prediction over $\vec{x}_i$, and $y$ be the ground truth of the $i^{th}$ variable of interest.
    
    As highlighted by Fig.~\ref{fig:regression_losses_focus} the Mean Bias Error ($MBE$) loss can be considered a base pillar for regression losses, characterized by many variations. Among them, the most relevant are Mean Absolute Error ($MAE$), Mean Squared Error ($MSE$) and the regualarised versions ($Lasso$ and $Ridge$), and Root Mean Squared Error ($RMSE$) losses. In this section, we are also going to introduce the Huber loss and the smooth L1, which are a blend between the $MAE$ and the $MSE$. Finally, the Log-cosh and the Root Mean Squared Logarithmic Error losses are presented.

    \subsection{Error Based Losses for Regression}
    Since regression models aim to minimize the prediction error between the actual and predicted values, the associated loss functions are typically classified as error-based. These losses directly measure the magnitude of residuals (i.e., the difference between observed and predicted values) to optimize model accuracy.
    
    \subsubsection{Mean Bias Error Loss (\cont, \diff)}  \label{subsubsec-regression_MBE}
    The most straightforward loss function is the Mean Bias Error loss, illustrated in Equation~\ref{eq:mbe}. It captures the average bias in the prediction but is rarely adopted as loss function to train regression models, because positive errors may cancel out the negative ones, leading to a potentially erroneous estimation of the parameters. Nevertheless, it is the starting point of the loss functions defined in the next subsections and it is commonly used to evaluate the performances of the models \cite{ullah2019short,krishnaiah2007neural,valipour2013comparison}.
    \begin{equation}
        \mathcal{L}_{MBE} = \frac{1}{N} \sum_{i=1}^{N} y_i - f(\vec{x}_i)
        \label{eq:mbe}
    \end{equation}
    Directly connected to $MBE$ are the Mean Absolute Error, the Mean Squared Error, and the Log-cosh losses, which primarily differ from $MBE$ in how they exploit the bias.
    
    \subsubsection{Mean Absolute Error Loss (\lcont,\conv)}\label{subsubsec-regression_MAE}
    The Mean Absolute Error loss or L1 loss is one of the most basic loss functions for regression, it measures the average of the absolute bias in the prediction. The absolute value overcomes the problem of the $MBE$ ensuring that positive errors do not cancel the negative ones. Therefore each error contributes to $MAE$ in proportion to the absolute value of the error.
    Notice that, the contribution of the errors follows a linear behavior, meaning that many small errors are as important as a big one. This implies that the gradient magnitude is not dependent on the error size, and thus may lead to convergence problems when the error is small. 
    A model trained to minimize the MAE is more effective when the target data conditioned on the input is symmetric.
    It is important to highlight that the derivative of the absolute value at zero is not defined. 
    
    As for MBE, MAE is also used to evaluate the performances of the models \cite{willmott2005advantages, li2010comparing}.
    \begin{equation}
        \mathcal{L}_{MAE} = \frac{1}{N} \sum_{i=1}^{N} \left | y_i - f(\vec{x}_i) \right |
    \end{equation}

    \subsubsection{Mean Squared Error Loss (\cont, \diff, \conv)}\label{subsubsec-regression_MSE}
    The Mean Squared Error (MSE) loss, or L2 loss, is the average of the squared differences between observed values $y$ and predicted values $\hat{y}$. It is widely used in regression tasks. The squared term ensures that all errors are positive and amplifies the impact of outliers, making it suitable for problems where the noise in observations follows a normal distribution.

    The MSE loss is defined as:
    \begin{equation}
        \mathcal{L}_{MSE} = \frac{1}{N} \sum_{i=1}^{N} \left ( y_i - f(\vec{x}_i) \right )^{2}
    \end{equation}

    One key drawback of MSE is its sensitivity to outliers, as large errors have a disproportionately high influence due to the squaring of residuals.

    \paragraph{Interpretation as Maximum Likelihood Estimation (MLE)}
    From a probabilistic viewpoint, MSE can be derived as a form of Maximum Likelihood Estimation (MLE) under the assumption that the errors between the predicted and observed values follow a Gaussian (normal) distribution with constant variance \cite{ML_3,murphy2013machine}. Minimizing MSE is equivalent to maximizing the likelihood of the observed data under this Gaussian noise assumption. This probabilistic interpretation explains why MSE is commonly used when the residuals are expected to follow a Gaussian distribution and highlights its role as a standard regression loss.

    \subsubsection{Lasso Regression ($L_1$ Regularization)}\label{subsubsec-lasso_regression}
Lasso regression is derived from augmenting the MSE loss with an $L_1$ regularization term, as detailed in Section \ref{L1-reg}. The regularized loss function penalizes the absolute magnitude of the model parameters, leading to sparsity in the learned model by driving some of the weights to zero. This makes Lasso particularly useful for feature selection in high-dimensional datasets \cite{lasso_regression,ML_2}.

The loss function for Lasso regression is:
\begin{equation}\label{lasso_loss}
    \mathcal{L}_{\text{Lasso}}(f(\vec{x}_i),\vec{y}_i)  = \frac{1}{N} \sum_{i=1}^{N} \left ( y_i - f(\vec{x}_i) \right )^{2} + \lambda \norm{\vec{\Theta}}_1
\end{equation}

The $L_1$ term encourages sparsity, shrinking irrelevant model weights to zero, which implicitly performs feature selection. However, as noted in Section \ref{L1-reg}, Lasso can struggle with correlated features, often selecting only one from a group of correlated variables. Additionally, Lasso is non-differentiable at zero, which may pose challenges in optimization, particularly when using gradient-based methods.

    \subsubsection{Ridge Regression ($L_2$ Regularization)}\label{subsubsec-ridge_regression}
Ridge regression is an extension of MSE with an $L_2$ regularization term, as described in Section \ref{L1-reg}. The $L_2$ term penalizes the square of the model parameters, discouraging large coefficients and helping to mitigate overfitting in regression tasks, especially when there is multicollinearity (high correlation between features) \cite{ridge_regression}.

The loss function for Ridge regression is:
\begin{equation}\label{ridge_loss}
    \mathcal{L}_{\text{Ridge}}(f(\vec{x}_i),\vec{y}_i)  = \frac{1}{N} \sum_{i=1}^{N} \left ( y_i - f(\vec{x}_i) \right )^{2} + \lambda \norm{\vec{\Theta}}_2^2
\end{equation}

\paragraph{Interpretation as Maximum A Posteriori Estimation (MAPE)}
From a Bayesian perspective, Ridge regression can be interpreted as Maximum A Posteriori Estimation (MAPE) \cite{murphy2013machine}. In this context, the $L_2$ regularization term corresponds to a Gaussian prior on the model parameters, with the objective being to maximize the posterior distribution of the parameters given the data. This probabilistic interpretation shows that Ridge regression shrinks the model coefficients towards zero without eliminating them entirely, as occurs in Lasso.

Ridge regression is particularly effective when dealing with correlated features, as it distributes the coefficient weights more smoothly across them. However, unlike Lasso, Ridge does not perform feature selection, retaining all input features with non-zero weights.

    \subsubsection{Root Mean Squared Error Loss(\cont,\diff,\conv)}\label{subsubsec-regression_RMSE}
    Directly connected to MSE, we have the Root Mean Squared Error loss, which is similar to MSE except for the square root term. 
    The main advantage is to make sure that the loss has the same units and scale of the variable of interest. 
    Since the only difference between the MSE and the RMSE consists in the application of the root term, the minimization process converges to the same optimal value. However, depending on the optimisation technique used, the RMSE may take different gradient steps. As the previously presented loss functions, it is also used as a metric to compare the performances of the model \cite{valipour2013comparison,li2010comparing}, and it shares the same limitations.
    \begin{equation}
        \mathcal{L}_{RMSE} = \sqrt{\frac{1}{N} \sum_{i=1}^{N} \left ( y_i - f(\vec{x}_i) \right )^{2}}
    \end{equation}

    \subsubsection{Huber loss  and Smooth L1 loss(\lcont,\diff,\sconv)}\label{subsubsec-regression_Huber}
    The Huber loss \cite{huber1965robust} is a variant of the MAE that becomes MSE when the residuals are small. It is parameterized by $\delta$, which defines the transition point from MAE to MSE. When $ \left | \mathbf{y_i} - f(\mathbf{x_i}) \right | \leq \delta $ the Huber loss follows the MSE, otherwise it follows the MAE.
    This allows it to combine the advantages of both the MAE and the MSE, when the difference between the prediction and the output of the model is huge errors are linear, making the Huber loss less sensitive to the outliers. Conversely, when the error is small, it follows the MSE making the convergence much faster and differentiable at 0.
    The choice of $\delta$ is fundamental and it can be constantly adjusted during the training procedure based on what is considered an outlier. The main limitation of the Huber loss resides in the additional extra hyperparameter $\delta$.
    \begin{equation}
        L_{Huberloss} = \left\{
        \begin{matrix}
            \frac{1}{2} \left ( y_i - f(\vec{x}_i) \right )^{2} &  for \left | y_i - f(\vec{x}_i) \right | \leq \delta,
            \\ \delta \left ( \left | y_i - f(\vec{x}_i) \right | -\frac{1}{2} \delta \right )
            & otherwise
        \end{matrix}\right.
    \end{equation}
    The choice of \(\delta\) is crucial and can be adjusted dynamically during the training process based on what is considered an outlier. 
    
    A specific case of the Huber loss is the Smooth L1 loss, obtained when \(\delta = 1\). This loss has been shown to be particularly useful in tasks where a balance is needed between sensitivity to small residuals and robustness to outliers, such as in object detection models \cite{girshick2015fast} or in tasks such as bounding box regression in object detection frameworks like Faster R-CNN \cite{ren2015faster}.
    
    \subsubsection{Log-cosh loss(\cont, \diff)}\label{subsubsec-regression_log-cosh}
    The log-cosh loss is the logarithm of the hyperbolic cosine of the residuals between the observed value $y$ and the predicted value $\hat{y}$. 
    As $log \left ( cosh \left ( \vec{x} \right ) \right )$ is approximately equal to $\frac{\vec{x}^{2}}{2}$ for small values of $\vec{x}$ it behave similarly to the MSE. For larger value of $\vec{x}$ instead, is nearly equivalent to $\left | \vec{x} \right | -  log \left ( 2 \right ) $ making it similar to MAE.
    \begin{equation}
        \mathcal{L}_{logcosh} = \frac{1}{N} \sum_{i=1}^{N} log \left ( cosh \left ( f(\vec{x}_i) - y_i \right ) \right )
    \end{equation}
    The log-cosh loss shares many of the advantages of the Huber loss, but without the need to manually set a threshold hyperparameter. However, it is computationally more expensive due to the logarithmic function. Additionally, it is less customizable compared to the Huber loss since it does not allow fine-tuning the transition between the quadratic and linear behavior.

    \subsubsection{Root Mean Squared Logarithmic Error Loss (\cont,\diff,\conv)}\label{subsubsec-regression_RMSLE}
The Root Mean Squared Logarithmic Error (RMSLE) loss, formalized in Eq.~\ref{eq:rmsle}, is the root mean squared error (RMSE) computed on the log-transformed observed value $y$ and predicted value $\hat{y}$. The transformation of both predicted and observed values allows for better handling of relative errors and makes the loss more robust to outliers, particularly when dealing with high-valued predictions. The addition of $1$ inside the logarithm ensures that zero values of $f(\vec{x}_i)$ can be handled without issues.

The RMSLE loss is defined as:
\begin{equation}
    \mathcal{L}_{RMSLE} = \sqrt{\frac{1}{N} \sum_{i=1}^{N} \left ( \log(y_i+1) - \log(f(\vec{x}_i)+1) \right )^{2}}
    \label{eq:rmsle}
\end{equation}

RMSLE is more robust to outliers than RMSE, as the logarithmic transformation reduces the penalization of large residuals when the predicted and actual values are high. This makes RMSLE suitable for datasets where the target values have an exponential relationship, or where underestimation should be penalized more than overestimation. However, it is unsuitable for tasks where the target values can be negative, as the logarithm is undefined for negative numbers.

Due to the properties of the logarithm, the magnitude of the RMSLE does not scale according to the magnitude of the error, meaning that large residuals are dampened when the overall target values are also large. This makes RMSLE particularly useful in regression tasks where targets exhibit large dynamic ranges, such as population growth prediction, stock prices, or financial data modeling \cite{chicco2021advantages,regression_rmsle}. However, the loss is less suitable for problems where exact magnitude matching is important, as the logarithmic transformation smooths the differences between large numbers \cite{semeniuta2017handy}.

Overall, RMSLE provides a valuable alternative to RMSE when relative differences and scaling behavior in target variables are critical, particularly in domains such as financial forecasting, population studies, or energy consumption prediction \cite{wang2006log,breiman1997arcing,lin2020frequency}.
    
\section{Classification losses} \label{sec:classification}
\subsection{Problem Formulation and Notation}\label{subsec-classification_losses_problem_formulation}
Classification is a subset of problems belonging to supervised learning.
The goal is to assign an input $\vec{x}$ to one of $K$ discrete classes. This goal can be pursued by training a model $f_{\vec{\Theta}}$ and its parameters $\vec{\Theta}$ by minimizing a loss function $L$. Let the target space of $f$ discrete and consider a model returning the output label, $f$ can be defined as:
\begin{align*}
        f \colon &\Phi \to \Lambda^K\\
        &\Lambda = \{0,1\}
\end{align*}
The above definition is working also for multi-label classification, since more than one label could be associated with a sample, e.g. $f(\vec{x})= [0, 1, 0, 1, 0, 0]$. To define single label classification we need to add the constraint that the output sum up to $1$, $\sum_k \lambda_k = 1$.\\
We can also consider models with continuous outputs, in case they return a probability $p_k(\vec{x}) \in [0,1]$ to a sample $\vec{x}$ for each possible assignable label $k \in 1, ..., K$:
\begin{align*}
        f \colon &\Phi \to P^K\\
        &P = [0,1]
\end{align*}
As before, to switch between multi-label and single-label classification,  we need to constraint the probabilities output to sum up to one, $\sum_k p_k = 1$, if we want to force a single-label assignment.

A more narrow notation for classification can be introduced to describe binary classification problems. This notation is useful in this work because margin-based losses are designed to solve binary classification problems and cannot directly be generalised to multi-class or multi-label classification. For the subset of binary classification problems the target space of $f$ is discrete and it is defined as follows:
\begin{align*}
        f \colon &\Phi \to B\\
        &B = \{-1,1\}
\end{align*}

\begin{figure}[H]
    \centering
    \includegraphics[width=0.8\linewidth]{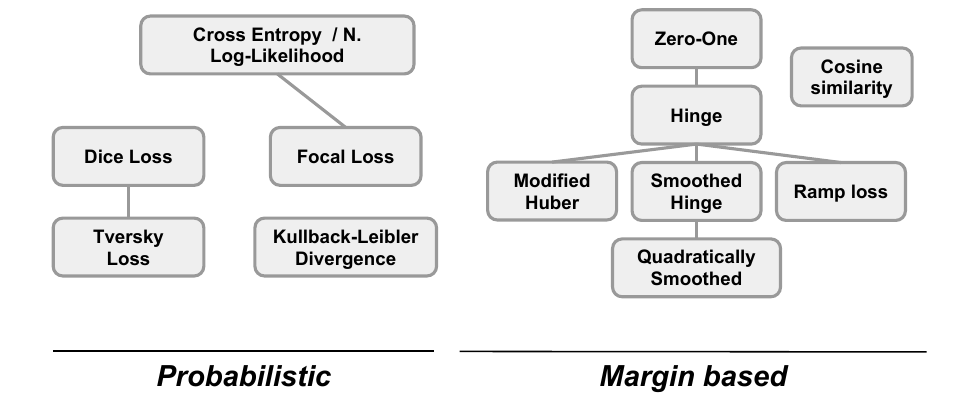}
    \caption{Overview of the classification losses divided into two major groups: margin-based losses and probabilistic ones.}
    \label{fig:classificarion_losses_focus}
\end{figure}

We define two different macro-categories of classification losses according to the underlying strategy employed to optimize them, namely the margin-based and the probabilistic ones as illustrated in Fig.~\ref{fig:classificarion_losses_focus}. In the next section, we introduce the margin-based loss function starting with the most basic and intuitive one, the Zero-One loss. Subsequently, we present the Hinge loss and its variants (the Smoothed and Quadratically Smoothed Hinge losses). Then, the Modified Huber loss, the Ramp loss, and the Cosine Similarity loss are described. 
Moreover, we introduce the probabilistic losses, beginning with the Cross-Entropy loss and Negative Log-Likelihood loss, which, from a mathematical point of view, coincide. Then, the Kullback-Leibler Divergence (KL) loss is presented, followed by several additional losses used in classification and segmentation tasks. These include the Focal Loss, which addresses class imbalance by focusing on difficult examples, and the Dice Loss, commonly used in segmentation tasks to measure the overlap between predicted and true segmentations. Lastly, the Tversky Loss, an extension of Dice Loss, is introduced to handle the trade-off between false positives and false negatives, particularly in imbalanced segmentation problems.

\subsection{Margin Based Loss Functions}\label{margin_based_loss_functions}
In this section, we introduce the most known margin-based loss functions.

\subsubsection{Zero-One loss}\label{subsubsec-classification_loss_zero_one}
The most basic and intuitive margin-based classification loss is the Zero-One loss, which assigns a value of 1 to a misclassified observation and 0 to a correctly classified one:

\begin{equation}
    L_{\text{ZeroOne}}(f(\vec{x}), y) = \begin{cases}
1 &\text{if $f(\vec{x}) \cdot y <0$}\\
0 &\text{otherwise}
\end{cases}
\end{equation}
Zero-One loss is not directly usable since it lacks convexity and differentiability. However, it is possible to derive employable surrogate losses that are classification calibrated, meaning they are a relaxation of $L_{\text{ZeroOne}}$, an upper bound, or an approximation of this loss. A significant achievement of recent literature on binary classification has been the identification of necessary and sufficient conditions under which such relaxations yield Fisher consistency\footnote{Fisher consistency, in this context, means that minimizing the surrogate loss also minimizes the expected Zero-One loss under the true data distribution, ensuring reliable classification outcomes.} \cite{Bartlett2006ConvexityCA,f4a7c73671ce457daaae7ceb2c0e1976,Lugosi2003OnTB,Mannor2003GreedyAF,1377497,Zhang01statisticalbehavior}.
All the following losses in this section satisfy such conditions.
\subsubsection{Hinge loss and Perceptron loss (\lcont,\conv)}\label{subsubsec-classification_perceptron_loss}
The most famous surrogated loss is the Hinge loss \cite{NIPS1998_a14ac55a}, which linearly penalizes every prediction where the resulting agreement is $<=1$. 
\begin{equation}\label{eq:hinge}
    L_{\text{Hinge}}(f(\vec{x}), y) = \max(0,1 - (f(\vec{x}) \cdot y))
\end{equation}
The Hinge loss is not strictly convex, but it is Lipschitz continuous and convex, so many of the usual convex optimizers used in machine learning can work with it.
The Hinge loss is commonly employed to optimise the Support Vector Machine (SVM \cite{bosertraining,mathur2008multiclass}).

To train the Perceptron \cite{rosenblatt1958perceptron} a variation of this loss, the Perceptron loss, is employed. This loss slightly differs from the Hinge loss, because it does not penalise samples inside the margin, surrounding the separating hyperplane, but just the ones that are mislabeled by this hyperplane with the same linear penalisation.
\begin{equation}
    L_{\text{Perceptron}}(f(\vec{x}), y) = \max(0, − (f(\vec{x}) \cdot y))
\end{equation}
There are two main drawbacks to using the hinge loss. Firstly, its adoption is used to make the model sensible to outliers in the training data. Secondly, due to the discontinuity of the derivative at  $(f(\vec{x}) \cdot y) = 1$, i.e. the fact that is not continuously differentiable, Hinge loss results be difficult to optimise.

\subsubsection{Smoothed Hinge loss (\lcont,\conv)}
A smoothed version of the Hinge loss was defined in \cite{Rennie2013SmoothHC} with the goal of obtaining a function easier to optimise as shown by the following equation:
\begin{equation}
    L_{\text{SmoothedHinge}}(f(\vec{x}), y) = \begin{cases}
\frac{1}{2} - (f(\vec{x}) \cdot t) &(f(\vec{x}) \cdot y)<=0\\
\frac{1}{2}(1-(f(\vec{x}) \cdot t))^2 &0<(f(\vec{x}) \cdot y)<1\\
0 &(f(\vec{x}) \cdot y)>=1
\end{cases}
\end{equation}
This smoothed version of the Hinge loss is differentiable.
Clearly, this is not the only possible smooth version of the Hinge loss.
However, it is a canonical one that has the important property of being zero for $z >= 1$ and it has a constant (negative) slope
for $z <= 0$. Moreover, for $0 < z < 1$, the loss smoothly transitions from a zero slope to a constant negative one.
This loss inherits sensibility to outliers from the original Hinge loss.
\subsubsection{Quadratically Smoothed Hinge loss (\lcont,\conv,\diff)}
With the same goal of the Smoothed Hinge loss a quadratically smoothed version has been defined in \cite{Zhang04solvinglarge}, to make it easier to optimise:
\begin{equation}
    L_{\text{QSmoothedHinge}}(f(\vec{x}), y) = \begin{cases}
\frac{1}{2\gamma}\max(0, − (f(\vec{x}) \cdot y))^2 &(f(\vec{x}) \cdot y)>= 1 - \gamma\\
1 - \frac{\gamma}{2} - (f(\vec{x}) \cdot y) & \text{otherwise}
\end{cases}
\end{equation}
The hyperparameter $\gamma$ determines the degree of smoothing, for $\gamma \rightarrow 0$ the loss becomes the original hinge. In contrast with the Smoothed Hinge loss, this version is not differentiable in the whole domain.
\subsubsection{Modified Huber loss (\lcont, \diff, \sconv)}
The Modified Huber loss is a slight variation of the Huber loss for regression and a special case of the Quadratic Smoothed Hinge loss with $\gamma=2$ (For more details refer to section \ref{subsubsec-regression_Huber}):
\begin{equation}
    L_{\text{ModHuber}}(f(\vec{x}), y) = \begin{cases}
\frac{1}{4}\max(0, − (f(\vec{x}) \cdot y))^2 &(f(\vec{x}) \cdot y)>= -1\\
 - (f(\vec{x}) \cdot y) & \text{otherwise}
\end{cases}
\end{equation}
\\
\subsubsection{Ramp loss (\cont,\conv)}\label{ramp_loss}
The Ramp loss, or Truncated Hinge, is a piece-wise linear, continuous, and convex loss that has been presented in \cite{Wu2007RobustTH}.  Under a multi-class setting, this loss is more robust to outliers. When employed in SVM, it produces more accurate classifiers using a smaller, and more stable, set of support vectors than the multi-class SVM that employes $L_{\text{Hinge}}$ \cite{Lee02multicategorysupport}.
\begin{equation}
L_{\text{Ramp}}(f(\vec{x}), y) = \begin{cases}
L_{\text{Hinge}}(f(\vec{x}), y)) &(f(\vec{x}) \cdot y)<= 1\\
 1 & \text{otherwise}
\end{cases}
\end{equation}
\subsubsection{Cosine Similarity loss (\lcont,\diff)}
Cosine similarity is generally used as a metric to measure distance when the magnitude of vectors is not important \cite{ML_3}. A typical example is related to text data representation by means of word counts \cite{ML_2,ML_3}. When the label and output can be interpreted  as vectors it is possible to derive a distance metric between them, which can be adapted into a loss function as follows:
\begin{equation}
    L_{cos-sim}(f(\vec{x}), \vec{y}) = 1 - \frac{\vec{y} \cdot \vec{f(x)}}{\norm{\vec{y}}\norm{\vec{f(x)}}}
\end{equation}
It is important to underline that when using Cosine Similarity loss, the range of possible values is restricted to the interval [-1, 1], which may not be suitable for all types of data or applications, particularly when interpretability is a key requirement.

\subsection{Probabilistic loss Functions}\label{probabilistc_loss_functions}
Let $q$ be the probability distribution underlying the dataset and $f_{\vec{\Theta}}$ the function generating the output, probabilistic loss functions provide some distance function between $q$ and $f_{\vec{\Theta}}$. By minimizing that distance, the model output distribution converges to the ground truth one. Usually, models trained with probabilistic loss functions can provide a measure of how likely a sample is labeled with one class instead of another \cite{ML_2,ML_3,generative_models} providing richer information w.r.t. margin based.
\subsubsection{Cross-Entropy loss and Negative Log-Likelihood loss (\cont,\diff,\conv)}\label{sec:cross-entropy}
Maximum likelihood estimation (MLE) is a method to estimate the parameters of a probability distribution by maximizing the likelihood \cite{ML_2,ML_3,MLE_tutorial}. 
From the point of view of Bayesian inference, MLE can be considered a special case of maximum a-posteriori estimation (MAP) that assumes a uniform prior distribution of the parameters.
Formally, it means that, given a dataset of samples $\mathcal{D}$, we are maximizing the following quantity:
\begin{equation}\label{eq:mle}
    P(\mathcal{D}|\vec{\Theta}) = \prod_{n=1}^{N}f_{\vec{\Theta}}(\vec{x}_i)^{y_i}\cdot(1 - f_{\vec{\Theta}}(\vec{x}_i))^{1-y_i}
\end{equation}
The aim is to find the maximum likelihood estimate by minimizing a loss function. 
To maximize Eq.~\ref{eq:mle}, we can turn it into a minimisation problem by employing the negative log-likelihood. To achieve this goal we need to define the following quantity:
\begin{equation}\label{eq:log_likelihood}
    log(P(\mathcal{D}|\vec{\Theta})) = \sum_{i=1}^{N}(y_i \log(f_{\vec{\Theta}}(\vec{x}_i)) + (1-y_i) \log(1 - f_{\vec{\Theta}}(\vec{x}_i))))
\end{equation}
and we can obtain the loss function by taking the negative of the log:
\begin{equation}\label{NLL}
        \mathcal{L}_{NLL} = - \sum_{i=1}^{N}(y_i \log(f_{\vec{\Theta}}(\vec{x}_i)) + (1-y_i) \log(1 - f_{\vec{\Theta}}(\vec{x}_i)))
\end{equation}
\\
Often, the above loss is also called the cross-entropy loss, because it can be derived by minimising the cross entropy between $f_{\vec{\Theta}}$ and $q$.

\begin{equation}
     H(q,f_{\vec{\Theta}}) = - \int q(\vec{x})\log(f_{\vec{\Theta}}(\vec{x}))d\vec{x}
\end{equation}
For the discrete case (which is the one we are interested in) the definition of the cross entropy is:
\begin{equation}
    H(q,f_{\vec{\Theta}}) = - \sum_{i=1}^{N} q(\vec{x}_i)\log(f_{\vec{\Theta}}(\vec{x}_i))
\end{equation}
Maximizing the likelihood for the parameters $\vec{\Theta}$ is the same as minimizing the cross-entropy as shown by the following equations:
\begin{align}\label{eq:cross_entropy_nll}
        \sum_{i=1}^{N}(y_i \log(f_{\vec{\Theta}}(\vec{x}_i)) + (1-y_i) \log(1 - f_{\vec{\Theta}}(\vec{x}_i)))
 = &\frac{1}{N} \prod_{n=1}^{N}f_{\vec{\Theta}}(\vec{x}_i)^{Ny_i}\\
 = & \sum_{i=1}^{N}q(\vec{x}_i)\log(f_{\vec{\Theta}}(\vec{x}_i)\\ = & -H(q,f_{\vec{\Theta}})
 \end{align}
\\
The classical approach to extend this loss to the multi-class scenario is to add as a final activation of the model a softmax function, defined accordingly to the number of ($K$) classes considered. 
Given a score for each class $f_k(\vec{x}) = s$, its output can be squashed to sum up to $1$ by mean of a softmax function $f_S$ obtaining:
\begin{equation}
    \widehat{f}_k(\vec{x}_i) = f_S(f_k(\vec{x}))
\end{equation}
where the softmax is defined as follows:
\begin{equation}\label{eq:softmax}
    f_S(s_i) = \frac{e^{s_i}}{\sum_{j=1}^{K} e^{s_j}}
\end{equation}
\\
The final loss (usually named categorical cross-entropy) is:
\begin{equation}
    L_{CCE} = -\frac{1}{K}\sum_{j=1}^{K}\log(\widehat{f}_k(\vec{x}))
\end{equation}
\subsubsection{Kullback-Leibler divergence (\cont, \conv, \diff)}\label{sec:kl-div}
The Kullback-Leibler (\(\mathcal{L}_{\text{KL}}\)) divergence is an information-based measure of disparity among probability distributions. Precisely, it is a non-symmetrical measurement of how one probability distribution differs from another one \cite{kl_div,ML_2,ML_3}. Technically speaking, \(\mathcal{L}_{\text{KL}}\) divergence is not a distance metric because it doesn’t obey the triangle inequality (\(\mathcal{L}_{\text{KL}}(q||f_{\vec{\Theta}})\) is not equal to \(\mathcal{L}_{\text{KL}}(f_{\vec{\Theta}}||q)\)). It is important to notice that, in the classification use case, minimizing the \(\mathcal{L}_{\text{KL}}\) divergence is the same as minimising the cross-entropy.

Precisely, the \(\mathcal{L}_{\text{KL}}\) divergence between two continuous distributions is defined as:
\begin{equation}\label{eq:kl-divergence}
    \mathcal{L}_{\text{KL}}(q||f_{\vec{\Theta}}) = \int q(\vec{x})\log\left(\frac{q(\vec{x})}{f_{\vec{\Theta}}(\vec{x})}\right) d\vec{x} = - \int q(\vec{x})\log(f_{\vec{\Theta}}(\vec{x}))d\vec{x} + \int q(\vec{x})\log(q(\vec{x}))d\vec{x}
\end{equation}
If we want to minimise \(\mathcal{L}_{\text{KL}}\) on the parameter \(\vec{\Theta}\), since the second integral is non-dependent on \(\vec{\Theta}\), we obtain:
\begin{equation}
\min_{\vec{\Theta}} \mathcal{L}_{\text{KL}}(q||f_{\vec{\Theta}}) = \min_{\vec{\Theta}} - \int q(\vec{x})\log(f_{\vec{\Theta}}(\vec{x}))d\vec{x} = \min_{\vec{\Theta}} H(q, f_{\vec{\Theta}})
\end{equation}

In general, cross-entropy is preferable to \(\mathcal{L}_{\text{KL}}\) divergence because it is typically easier to compute and optimize. Cross-entropy only involves a single sum over the data, whereas \(\mathcal{L}_{\text{KL}}\) divergence involves a double sum. This can make it more computationally efficient, especially when working with large datasets.

\subsubsection{Focal Loss (\lcont, \conv)}\label{subsubsec-classification_focal_loss}
Focal Loss \cite{lin2017focal} is a modification of the cross-entropy loss designed to address class imbalance by down-weighting the contribution of easy examples and focusing more on hard, misclassified examples. This is particularly effective in tasks such as object detection, where imbalanced datasets are common.

The Focal Loss is defined as:
\begin{equation}\label{eq:focal_loss}
    \mathcal{L}_{\text{Focal}} = - (1 - f_{\vec{\Theta}}(\mathbf{x}_i))^\gamma \log(f_{\vec{\Theta}}(\mathbf{x}_i)),
\end{equation}
where \( f_{\vec{\Theta}}(\mathbf{x}_i) \) is the predicted probability for the true class, and \( \gamma \geq 0 \) is the focusing parameter that controls the strength of the modulation. When \( \gamma = 0 \), Focal Loss reduces to the standard cross-entropy loss in the binary scenario.

Focal Loss is Lipschitz continuous (\lcont) and convex (\conv) for the predicted probabilities \( f_{\vec{\Theta}}(\mathbf{x}_i) \).

\subsubsection{Dice Loss (\cont, \diff)}\label{subsubsec-classification_dice_loss}
Dice Loss \cite{sudre2017generalised} is widely used in image segmentation tasks to handle class imbalance. It is based on the Dice coefficient, a measure of overlap between the predicted and true segmentation. Dice Loss is particularly effective when there is a significant imbalance between foreground and background classes.

The Dice Loss is defined as:
\begin{equation}\label{eq:dice_loss}
    \mathcal{L}_{\text{Dice}} = 1 - \frac{2 \sum_i f_{\vec{\Theta}}(\mathbf{x}_i) y_i}{\sum_i f_{\vec{\Theta}}(\mathbf{x}_i) + \sum_i y_i},
\end{equation}
where \( f_{\vec{\Theta}}(\mathbf{x}_i) \) is the predicted probability for pixel \( i \), and \( y_i \) is the corresponding ground truth.

Dice Loss is continuous (\cont) and differentiable (\diff), but not convex. It is commonly employed in medical image segmentation tasks, where accurately measuring the overlap between the predicted and true segmentation is crucial.

\subsubsection{Tversky Loss (\cont, \diff)}\label{subsubsec-classification_tversky_loss}
Tversky Loss \cite{salehi2017tversky} extends the Dice Loss by introducing parameters that control the trade-off between false positives and false negatives. This makes Tversky Loss especially useful in segmentation tasks with high class imbalance, where one class dominates the other.

The Tversky Loss is defined as:
\begin{equation}\label{eq:tversky_loss}
    \mathcal{L}_{\text{Tversky}} = 1 - \frac{\sum_i f_{\vec{\Theta}}(\mathbf{x}_i) y_i}{\sum_i f_{\vec{\Theta}}(\mathbf{x}_i) y_i + \alpha \sum_i f_{\vec{\Theta}}(\mathbf{x}_i) (1 - y_i) + \beta \sum_i (1 - f_{\vec{\Theta}}(\mathbf{x}_i)) y_i},
\end{equation}
where \( f_{\vec{\Theta}}(\mathbf{x}_i) \) is the predicted probability for pixel \( i \), \( y_i \) is the corresponding ground truth, and \( \alpha \) and \( \beta \) are parameters that balance false positives and false negatives.

Tversky Loss is continuous (\cont) and differentiable (\diff), and by adjusting \( \alpha \) and \( \beta \), it can emphasize reducing false positives or false negatives, depending on the task's requirements. This flexibility makes it a powerful tool in imbalanced segmentation problems.

\section{Generative losses} \label{sec:generative}
\begin{figure}[H]
    \centering
    \includegraphics[width=0.8\linewidth]{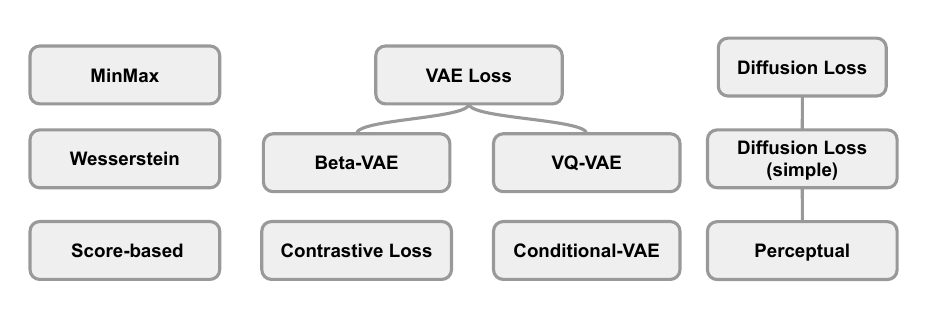}
    \caption{Overview of the generative losses.}
    \label{fig:generative_losses_focus}
\end{figure}

In recent years, generative models have become particularly valuable for modeling complex data distributions and regenerating realistic samples from them~\cite{goodfellow2020generative,goodfellow2014generative}. In this section, as illustrated in Fig.~\ref{fig:generative_losses_focus}, we describe the primary losses associated with Variational Autoencoders (VAEs) (Section~\ref{sec:vae}), Generative Adversarial Networks (GANs) (Section~\ref{sec:gan}), Diffusion Models (Section~\ref{sec:diffusion}) and Transformers with a focus on LLMs (Section~\ref{sec:transformers}). 

While this survey focuses on the core generative models, other architectures like Pixel RNNs~\cite{van2016pixel}, realNVP~\cite{dinh2016density}, Flow-based models~\cite{rezende2015variational}, and WaveNet~\cite{oord2016wavenet} are also impactful but fall beyond the scope of this work.

\subsection{Variational Autoencoders (VAEs)}\label{sec:vae}
Variational Autoencoders (VAEs) \cite{kingma2013auto,rezende2014stochastic} are generative models that learn a latent representation of data through probabilistic encoding and decoding. By modeling the underlying structure of data, VAEs aim to produce a latent space that follows a known prior distribution, typically Gaussian.

A VAE consists of two main components: an encoder and a decoder. The encoder maps the input data $\mathbf{x}$ to a latent variable $\mathbf{z}$, capturing key data characteristics in a compact form through the approximate posterior distribution $q(\mathbf{z}|\mathbf{x})$. The decoder then reconstructs the original data $\mathbf{x}$ from $\mathbf{z}$, modeling the conditional distribution $p(\mathbf{x}|\mathbf{z})$. This probabilistic framework allows VAEs to generate new samples by sampling from the learned latent space.

VAEs find applications in image generation, semi-supervised learning, and anomaly detection. In image generation, they facilitate smooth latent representations, useful for denoising and sample synthesis \cite{hou2017deep,bengio2013representation}. In semi-supervised learning, VAEs leverage both labeled and unlabeled data to improve classification \cite{kingma2014semi}. For anomaly detection, VAEs model normal data distributions, identifying anomalies via high reconstruction errors for out-of-distribution samples \cite{an2015variational}. Despite successes, VAEs face challenges such as image blurriness, often addressed by discrete latent spaces like those in VQ-VAE \cite{van2017neural}.

\subsubsection{VAE Loss (ELBO) (\cont, \diff, \lcont)}\label{subsubsec-generative_vae_loss}
The VAE loss function is derived from the Evidence Lower Bound (ELBO), which provides a lower bound on the data log-likelihood. It comprises two main components: the reconstruction loss $\mathcal{L}_{\text{recon}}$ and the Kullback-Leibler (KL) divergence $\mathcal{L}_{\text{KL}}$ (see section \ref{sec:kl-div}).

The reconstruction loss encourages the decoder to accurately reproduce the input data $\mathbf{x}$ from the latent variable $\mathbf{z}$ by minimizing the difference between the original data and its reconstruction. Depending on the data type, $\mathcal{L}_{\text{recon}}$ can be the mean squared error (MSE, see section \ref{subsubsec-regression_MSE}) for real-valued data or the binary cross-entropy (negative log-likelihood, see section \ref{sec:cross-entropy}) for binary data. In a unified expression, we can represent $\mathcal{L}_{\text{recon}}$ as the expected log-likelihood with respect to the approximate posterior distribution of the latent variable:
\[
\mathcal{L}_{\text{recon}} = - \mathbb{E}_{q(\mathbf{z}|\mathbf{x})}[\log f_{\vec{\Theta}}(\mathbf{x}|\mathbf{z})].
\]

The full VAE loss, often referred to as the negative ELBO, is defined as:
\begin{equation}\label{eq:vae_loss}
    \mathcal{L}_{VAE} = -\mathbb{E}_{q(\mathbf{z}|\mathbf{x})}[\log f_{\vec{\Theta}}(\mathbf{x}|\mathbf{z})] + \mathcal{L}_{\text{KL}}(q(\mathbf{z}|\mathbf{x}) \| f_{\vec{\Theta}}(\mathbf{z})),
\end{equation}
where the first term is the reconstruction loss and the second term, the KL divergence, regularizes the latent space to follow the prior distribution.

While $\mathcal{L}_{\text{recon}}$ and $\mathcal{L}_{\text{KL}}$ may individually exhibit convexity under certain conditions, their combined interaction, particularly with neural network parameterization, results in an overall non-convex loss function \cite{ML_3,rezende2014stochastic,kingma2013auto}. 

\bigskip

The reconstruction loss alone focuses on achieving high-fidelity reconstructions of the input data. However, without the KL divergence term, the model does not impose any structure on the latent space, leading to unstructured representations that generalize poorly to generating new samples \cite{kingma2013auto,rezende2014stochastic}.

In contrast, the ELBO, which combines the reconstruction loss with the KL divergence term, introduces a trade-off between accurate reconstruction and regularization of the latent space. The KL divergence term encourages the latent space to follow a smooth prior distribution, facilitating sample generation by ensuring structured, regularized representations. However, if the KL divergence term dominates, it can lead to poorer reconstructions as the model prioritizes matching the prior over precise data recovery~\cite{higgins2017beta}. Excessive weight on the KL term can also cause posterior collapse, where the model ignores the latent variables entirely, resulting in suboptimal generative performance~\cite{bowman2015generating,alemi2018fixing}.

In summary, while $\mathcal{L}_{\text{recon}}$ alone enhances data fidelity, combining it with $\mathcal{L}_{\text{KL}}$ (in the ELBO) enables proper generative modeling. Achieving realistic data generation requires balancing accurate reconstruction with a regularized latent space~\cite{burgess2018understanding}.

\subsubsection{Extensions of VAE Losses}
Several extensions of the VAE loss have been proposed to improve the model's performance or flexibility. The most notable variants include:

\begin{itemize}
    \item \textbf{Beta-VAE} (\cont, \diff, \lcont,): The Beta-VAE \cite{higgins2017beta} introduces a hyperparameter $\beta$ to weight the KL divergence term. The loss function becomes:
    \begin{equation}
        \mathcal{L}_{\beta\text{-VAE}} = \mathcal{L}_{\text{recon}} + \beta \mathcal{L}_{\text{KL}}.
    \end{equation}
    By adjusting $\beta$, the balance between reconstruction accuracy and latent space regularization can be controlled \cite{burgess2018understanding}. This variant retains the continuity, differentiability, and Lipschitz continuity but it is non-convex due to the interaction between terms.

    Beta-VAE improves interpretability by encouraging the model to learn more disentangled representations. This is particularly useful in applications where distinct, independent latent factors are beneficial, as in unsupervised learning tasks or when a well-structured latent space is desired \cite{higgins2017beta}. The introduction of $\beta$ can lead to a trade-off where increasing regularization may harm reconstruction quality. Excessively high values of $\beta$ can also cause posterior collapse, where the model ignores the latent variables, resulting in reduced generative performance \cite{alemi2018fixing,bowman2015generating,burgess2018understanding}.

    \item \textbf{VQ-VAE} (\cont): In Vector Quantized VAEs \cite{van2017neural}, the latent space is discrete, rather than continuous, and a codebook is used to quantize the latent variables. The key difference in VQ-VAE compared to traditional VAEs is the use of a discrete latent representation, which introduces the following steps:

Given an input $\mathbf{x}$, the encoder produces a continuous latent variable $\mathbf{z}_e(\mathbf{x}) \in \mathbb{R}^D$. However, instead of passing $\mathbf{z}_e(\mathbf{x})$ directly to the decoder, VQ-VAE performs a vector quantization by mapping $\mathbf{z}_e(\mathbf{x})$ to the nearest vector in a learned codebook $\mathcal{E} = \{\mathbf{e}_k\}_{k=1}^K$, where each $\mathbf{e}_k \in \mathbb{R}^D$ is a learned embedding vector. This process can be written as:
\begin{equation}
    \mathbf{z}_q(\mathbf{x}) = \mathbf{e}_k, \quad \text{where} \quad k = \arg \min_j \|\mathbf{z}_e(\mathbf{x}) - \mathbf{e}_j\|_2.
\end{equation}

The quantized latent variable $\mathbf{z}_q(\mathbf{x})$ is then passed to the decoder, which reconstructs the input as $\hat{\mathbf{x}} = p(\mathbf{x}|\mathbf{z}_q)$.

The VQ-VAE loss function consists of three terms:
\begin{equation}
    \mathcal{L}_{VQ-VAE} = \mathcal{L}_{\text{recon}} + \|\text{sg}[\mathbf{z}_e(\mathbf{x})] - \mathbf{e}_k\|_2^2 + \beta \|\mathbf{z}_e(\mathbf{x}) - \text{sg}[\mathbf{e}_k]\|_2^2,
\end{equation}
where:
\begin{itemize}
    \item $\mathcal{L}_{\text{recon}}$ is the reconstruction loss (typically mean squared error or binary cross-entropy).
    \item The second term $\|\text{sg}[\mathbf{z}_e(\mathbf{x})] - \mathbf{e}_k\|_2^2$ (the codebook loss) ensures that the codebook vector $\mathbf{e}_k$ is close to the encoder output $\mathbf{z}_e(\mathbf{x})$.
    \item The third term $\|\mathbf{z}_e(\mathbf{x}) - \text{sg}[\mathbf{e}_k]\|_2^2$ (the commitment loss) encourages the encoder to commit to a particular codebook vector, where $\text{sg}[\cdot]$ denotes the **stop gradient** operator, ensuring gradients only flow through the appropriate part of the network.
    \item $\beta$ is a hyperparameter controlling the weight of the commitment loss.
\end{itemize}

While VQ-VAE remains continuous (\cont) overall, the discrete quantization step introduces non-differentiability in the loss function, as the mapping from encoder outputs to codebook vectors is non-differentiable. 

The key advantage of VQ-VAE is that the use of a discrete latent space can produce sharper and higher-quality generated samples, addressing some of the issues with blurry outputs often observed in continuous VAEs \cite{van2017neural}. VQ-VAE is particularly beneficial in tasks where the data have inherently discrete characteristics, such as in audio and image generation \cite{dhariwal2020jukebox,razavi2019generating}.

    \item \textbf{Conditional VAE (CVAE)} (\cont, \diff, \lcont): Conditional VAEs \cite{sohn2015learning} condition both the encoder and decoder on auxiliary information (e.g., class labels), modifying the ELBO to learn the conditional distribution $p(\mathbf{x}|\mathbf{y})$. The CVAE loss is:
    \begin{equation}
        \mathcal{L}_{CVAE} = -\mathbb{E}_{q(\mathbf{z}|\mathbf{x}, \mathbf{y})}[\log f_{\vec{\Theta}}(\mathbf{x}|\mathbf{z}, \mathbf{y})] + \mathcal{L}_{\text{KL}}(q(\mathbf{z}|\mathbf{x}, \mathbf{y}) \| f_{\vec{\Theta}}(\mathbf{z}|\mathbf{y}))
    \end{equation}
    The CVAE loss is continuous, differentiable, and Lipschitz continuous, and the KL divergence remains convex, but the overall loss is still non-convex due to the interaction between terms. 
    
    The advantage of using CVAE is its ability to generate conditional outputs based on specific attributes or labels. This makes CVAE particularly useful in scenarios where controlled generation is required, such as in image generation conditioned on class labels or text generation based on input attributes \cite{kingma2014semi,yan2016attribute2image}. By incorporating auxiliary information, CVAEs allow for more structured and interpretable latent spaces, improving the ability to generate targeted samples in a variety of applications \cite{kingma2014semi}.
\end{itemize}

\color{black}
    \subsection{Generative Adversarial Networks}\label{sec:gan}
    Generative Adversarial Networks (GANs) are used to create new data instances that are sampled from the training data. GANs have two main components: 
    \begin{itemize}
        \item The \textbf{generator}, referred as $G(\svec{z})$, which generates data starting from random noise and tries to replicate real data distributions
        \item The \textbf{discriminator}, referred to as $D(\svec{x})$, learns to distinguish the generator's fake data from the real one. It applies penalties in the generator loss for producing distinguishable fake data compared with real data.
    \end{itemize}
    The GAN architecture is relatively straightforward, although one aspect remains challenging: GAN loss functions. Precisely, the discriminator is trained to provide the loss function for the generator. If generator training goes well, the discriminator gets worse at telling the difference between real and fake samples. It starts to classify fake data as real, and its accuracy decreases. 
    
    Both the generator and the discriminator components are typically neural networks, where the generator output is connected directly to the discriminator input. The discriminator's classification provides a signal that the generator uses to update its weights through back-propagation.\par
    As GANs try to replicate a probability distribution, they should use loss functions that reflect the distance between the distribution of the data generated by the GAN and the distribution of the real data. \par
    Two common GAN loss functions are typically used: \textbf{minimax loss}~\cite{minimax} and \textbf{Wasserstein loss}~\cite{Wasserstein}.
    The generator and discriminator losses derive from a single distance measure between the two aforementioned probability distributions. The generator can only affect one term in the distance measure: the term that reflects the distribution of the fake data. During generator training, we drop the other term, which reflects the real data distribution.
    The generator and discriminator losses look different, even though they derive from a single formula.
    
    In the following, both minimax and Wasserstein losses are written in a general form. The properties of the loss function (\cont, \diff, etc.) are identified based on the function chosen for the generator or discriminator. 
\subsubsection{Minimax loss}
    The generative model $G$ learns the data distributions and is trained simultaneously with the discriminative model $D$. The latter estimates the probability that a given sample is identical to the training data rather than $G$. $G$ is trained to maximize the likelihood of tricking $D$~\cite{minimax}. 
    In other words, the generator tries to minimize the following function while the discriminator tries to maximize it:
    \begin{equation}
        \begin{aligned}
            \mathcal{L}_{\text{minimax}}(D, G) &= E_{\svec{x}}[\log(D(\svec{x})) + \\
            &\quad  E_{\svec{z}}[\log(1 - D(G(\svec{z})))],
        \end{aligned}
    \end{equation}
    
    where: 
    \begin{itemize}
        \item $D(\svec{x})$ is the discriminator's estimate of the probability that real data instance \svec{x} is real, 
        \item $E_{\svec{x}}$ is the expected value over all real data instances,
        \item $G(\svec{z})$ is the generator's output when given noise $\svec{z}$,
        \item $D(G(z))$ is the discriminator's estimate of the probability that a fake instance is real, 
        \item $E_{\svec{z}}$ is the expected value over all random inputs to the generator (in effect, the expected value over all generated fake instances $G(\svec{z}))$.
    \end{itemize}
    The loss function above directly represents the cross-entropy between real and generated data distributions. The generator can not directly affect the $\log(D(\svec{x}))$ term in the function, and it only minimizes the term $\log(1 - D(G(\svec{z})))$. A disadvantage of this formulation of the loss function is that the above minimax loss function can cause the GAN to get stuck in the early stages of the training when the discriminator received trivial tasks. Therefore, a suggested modification to the loss~\cite{minimax} is to allow the generator to maximize $\log (D(G(\svec{z})))$.
\subsubsection{Wasserstein loss}
    The Wasserstein distance gives an alternative method of training the generator to better approximate the distribution of the training dataset. In this setup, the training of the generator itself is responsible for minimizing the distance between the distribution of the training and generated datasets. The possible solutions are to use distribution distance measures, like Kullback-Leibler (KL) divergence, Jensen-Shannon (JS) divergence, and the Earth-Mover (EM) distance (also called Wasserstein distance). The main advantage of using Wasserstein distance is due to its differentiability and having a continuous linear gradient~\cite{Wasserstein}.
    
    
    A GAN that uses a Wasserstein loss, known as a WGAN, does not discriminate between real and generated distributions in the same way as other GANs. Instead, the WGAN discriminator is called a "critic," and it scores each instance with a real-valued score rather than predicting the probability that it is fake. This score is calculated so that the distance between scores for real and fake data is maximised.
    
    The advantage of the WGAN is that the training procedure is more stable and less sensitive to model architecture and selection of hyperparameters.
    
    The two loss functions can be written as: 

\begin{equation}
    \begin{aligned}
        \mathcal{L}_{\text{critic}} &= D(\svec{x}) - D(G(\svec{z})) \\
        \mathcal{L}_{\text{generator}} &= D(G(\svec{z}))
    \end{aligned}
\end{equation}

    \bigskip
    The discriminator tries to maximize $\mathcal{L}_{critic}$. In other words, it tries to maximize the difference between its output on real instances and its output on fake instances. The generator tries to maximize $\mathcal{L}_{Generator}$. In other words, It tries to maximize the discriminator's output for its fake instances.

    The benefit of Wasserstein loss is that it provides a useful gradient almost everywhere, allowing for the continued training of the models. It also means that a lower Wasserstein loss correlates with better generator image quality, meaning that it explicitly seeks a minimization of generator loss. Finally, it is less vulnerable to getting stuck in a local minimum than minimax-based GANs~\cite{Wasserstein}. However, accurately estimating the Wasserstein distance using batches requires unaffordable batch size, which significantly increases the amount of data needed~\cite{stanczuk2021wasserstein}.

\subsection{Diffusion models}\label{sec:diffusion}
Diffusion Models are generative models that rely on probabilistic likelihood estimation to generate data samples. Originally inspired by the physical phenomenon of diffusion, these models systematically add noise to the data and then learn to reverse this process. The forward process corrupts the data by progressively adding noise, and the reverse process reconstructs the data by removing the noise, a procedure learned through a neural network that models conditional probability densities~\cite{sohl2015deep,song2019generative,ho2020denoising}.

The forward diffusion process adds Gaussian noise to the data over $T$ time steps, transforming it into pure noise. The reverse diffusion process then uses a neural network to learn the conditional probabilities of each time step, gradually denoising the data to reconstruct the original input.

\paragraph{Forward diffusion process}
Given a data point $\mathbf{x}_0$ sampled from a real data distribution $q(\mathbf{x})$, the forward process adds small amounts of Gaussian noise iteratively $T$ times, resulting in a sequence of noisy samples $\mathbf{x}_1, \cdots, \mathbf{x}_T$. The noise distribution is Gaussian, and because the forecasted probability density at time $t$ depends only on the previous state $t-1$, the conditional probability density is:
\begin{equation}
    q(\mathbf{x}_t \vert \mathbf{x}_{t-1}) = \mathcal{N}(\mathbf{x}_t; \sqrt{1 - \beta_t} \mathbf{x}_{t-1}, \beta_t \mathbf{I}),
\end{equation}
with $\beta_t \in (0,1)$, a hyperparameter that can be constant or variable, and $t \in [1, T]$. As $T \rightarrow \infty$, $\mathbf{x}_T$ converges to a pure Gaussian distribution. The forward process does not require neural network training but involves iteratively applying Gaussian noise to the data.

\paragraph{Reverse diffusion process}
In the reverse process, given the noisy state $\mathbf{x}_T$, the goal is to estimate the probability density of the original data by gradually removing the noise. The neural network, parameterized by $\vec{\Theta}$, is trained to approximate $f_\vec{\Theta}(\mathbf{x}_{t-1} \vert \mathbf{x}_t)$, enabling the reconstruction of $\mathbf{x}_{0}$ from $\mathbf{x}_T$. The reverse diffusion process is modeled as:
\begin{equation}
    f_\vec{\Theta}(\mathbf{x}_{t-1} \vert \mathbf{x}_t) = \mathcal{N}(\mathbf{x}_{t-1}; \boldsymbol{\mu}_\vec{\Theta}(\mathbf{x}_t, t), \boldsymbol{\Sigma}_\vec{\Theta}(\mathbf{x}_t, t)),
\end{equation}
where $\boldsymbol{\mu}_\vec{\Theta}(\mathbf{x}_t, t)$ is a learned mean function, and $\boldsymbol{\Sigma}_\vec{\Theta}(\mathbf{x}_t, t)$ is a learned variance term. Sampling from this distribution allows the model to reconstruct the data by successively denoising the noisy sample from time step $T$ down to $t = 0$~\cite{ho2020denoising}.

Recent work has demonstrated that diffusion models can outperform other generative models, such as GANs, in certain tasks~\cite{dhariwal2021diffusion,song2020improved}, though they tend to be computationally more expensive due to the large number of forward and reverse steps involved.

\subsubsection{Diffusion Model Loss Function (\cont, \diff)}
The objective of diffusion models is to minimize the variational lower bound (VLB) on the data likelihood, which is formulated as:
\begin{equation}
    \mathcal{L}_{\text{diffusion}} = \mathbb{E}_t \left[\log p(\mathbf{x}_T) - \sum_{t \geq 1} \log \frac{f_\vec{\Theta}(\mathbf{x}_{t-1}|\mathbf{x}_t)}{q(\mathbf{x}_t|\mathbf{x}_{t-1})}\right].
\end{equation}
A simplified version of this loss is often used to train Denoising Diffusion Probabilistic Models (DDPMs)~\cite{ho2020denoising}:
\begin{equation}
    \mathcal{L}_{\text{diffusion}}^\text{simple} = \mathbb{E}_{t, \mathbf{x}_0, \boldsymbol{\epsilon}_t} \left[\|\boldsymbol{\epsilon}_t - \boldsymbol{\epsilon}_\vec{\Theta}(\sqrt{\bar{\alpha}_t} \mathbf{x}_0 + \sqrt{1 - \bar{\alpha}_t} \boldsymbol{\epsilon}_t, t)\|^2 \right],
\end{equation}
where the model learns to predict the noise $\boldsymbol{\epsilon}_\vec{\Theta}$ added at each step. This simplified loss connects closely to score matching, as the model learns to predict the noise, which is essential for reconstructing the data.

The simplified loss function (also callsed Noise Prediction Loss) in DDPMs is computationally efficient and relatively straightforward to implement. It uses a mean squared error (MSE) objective, which is cheap to compute. Each training iteration requires the model to predict the added noise and minimize the difference between the actual and predicted noise. This formulation is easy to optimize with standard gradient-based methods, leading to faster convergence and more stable training compared to more complex alternatives. As a result, DDPM's simplified loss provides a good trade-off between computational efficiency and sample quality, making it suitable for a wide range of generative tasks~\cite{ho2020denoising}.

\subsubsection{Other MSE-Based Losses in Diffusion Models (\cont, \diff)}\label{subsubsec-generative_other_mse_losses}
While the Noise Prediction Loss is central to diffusion models, other MSE-based losses are also employed to optimize different aspects of the generation process. These losses extend the basic MSE formulation to handle higher-level tasks and representations:

\paragraph{Perceptual Loss} 
In tasks such as super-resolution and image synthesis, where pixel-level differences may fail to capture perceptual quality, Perceptual Loss \cite{rombach2022high} is applied. This loss leverages high-level feature representations extracted from a pre-trained convolutional neural network (e.g., VGG). The MSE objective is then applied to these feature maps, rather than pixel values, ensuring that the generated images align more closely with human perception.

\paragraph{Latent Space Regularization} 
In latent diffusion models \cite{rombach2022high}, MSE is applied to regularize the latent space during the diffusion process. This regularization ensures that the latent representations of the input data remain smooth and structured, improving the stability and consistency of the generation process. Latent space regularization helps to prevent issues such as mode collapse and ensures higher-quality outputs.

\subsubsection{Score-based generative model loss (\cont, \diff)}\label{subsubsec-sb_model_loss}
In Score-Based Generative Models (SBGMs)~\cite{song2020score}, the diffusion process is generalized to a continuous-time framework using stochastic differential equations (SDEs). The model learns the score function, i.e., the gradient of the log probability density at different noise levels.

The loss function for learning the score function is derived from denoising score matching:
\begin{equation}
    \mathcal{L}_{\text{score}} = \mathbb{E}_{\mathbf{x}_0, \sigma} \left[ \lambda(\sigma) \| \nabla_{\mathbf{x}} \log f_\vec{\Theta}(\mathbf{x}_t | \sigma) + \frac{\mathbf{x}_t - \mathbf{x}_0}{\sigma^2} \|^2 \right],
\end{equation}
where $\mathbf{x}_t$ is the noisy sample, $\log f_\vec{\Theta}(\mathbf{x}_t | \sigma)$ is the predicted score function, and $\lambda(\sigma)$ is a weighting function for different noise levels.

In contrast with the simplified loss function in DDPMs, the score-based loss used in Score-Based Generative Models (SBGMs) is computationally more intensive. The model must learn the gradient of the log probability (the score function), which involves higher-order derivatives and sampling from various noise scales. Each iteration requires calculating and backpropagating through the score function, significantly increasing the computational cost. While this approach can provide richer generative performance and flexibility, particularly in continuous-time frameworks, the added complexity often results in slower training and higher resource requirements~\cite{song2020score}.

\subsubsection{Cosine Similarity in Multimodal Context (\cont, \diff)}\label{subsubsec-generative_clip_loss}
Cosine similarity plays a crucial role in aligning representations across different modalities, particularly in models like CLIP (Contrastive Language-Image Pretraining \cite{radford2021learning}), which embeds both text and images in a shared feature space. In multimodal tasks, such as text-to-image generation, this approach ensures that representations from different domains are semantically consistent.

In text-to-image diffusion models, a common use of cosine similarity is the CLIP Guidance Loss \cite{rombach2022high}, where the goal is to align the generated image with a given text prompt. The loss leverages a pre-trained CLIP model that computes embeddings for both the text and the generated image. By maximizing the cosine similarity between the text and image embeddings, the model enhances the semantic coherence of the generated images with respect to the input text.

The CLIP guidance loss is defined as:
\begin{equation}\label{eq:clip_loss}
    \mathcal{L}_{\text{CLIP}} = 1 - \cos(\phi_{\text{text}}(\mathbf{y}), \phi_{\text{image}}(\hat{\mathbf{x}})),
\end{equation}
where $\phi_{\text{text}}(\mathbf{y})$ is the CLIP embedding of the input text $\mathbf{y}$, and $\phi_{\text{image}}(\hat{\mathbf{x}})$ is the CLIP embedding of the generated image $\hat{\mathbf{x}}$.

This loss is continuous (\cont) and differentiable (\diff) with respect to the generated image, as both the CLIP encoders and cosine similarity function are composed of standard neural network operations that are differentiable. Typically, the pre-trained CLIP model is frozen during the training of diffusion models, allowing gradients to flow back through the generated image while keeping the CLIP parameters unchanged.

\subsection{Transformers and LLM Loss Functions}\label{sec:transformers}
Transformer has revolutionized machine learning, particularly in natural language processing (NLP), by enabling the parallel processing of sequential data using self-attention mechanisms \cite{vaswani2017attention}. This architecture has been the foundation for many powerful models, including Large Language Models (LLMs) such as GPT \cite{radford2019language} and BERT \cite{devlin2018bert}, which excel at generative tasks like text generation, translation, and summarization.

The Transformer model is built around an encoder-decoder structure, with the encoder processing the input data and the decoder generating the output sequence. An attention mechanism is a technique used in machine learning and artificial intelligence to improve model performance by focusing on relevant information. In Transformers, the attention mechanism allows the model to selectively assign varying importance to different elements in the sequence, enabling it to capture long-range dependencies and context effectively. Within each encoder and decoder layer, attention mechanisms assign scores to tokens based on relevance, ensuring that the model attends to the most pertinent information for each output. After attention, feedforward layers further refine these representations, capturing complex patterns and allowing for high parallelization, which enables Transformers to efficiently handle long sequences and excel at tasks involving sequential data.

Transformer-based models, particularly LLMs, are trained using a variety of loss functions depending on the task. This section explores the primary loss functions used in transformer and LLM training. 

\subsubsection{Probabilistic Losses in LLMs}\label{subsec-transformer_probabilistic_losses}
Probabilistic losses are central to training LLMs, as they help models learn from probability distributions over sequences, tokens, and classes. These losses are primarily extensions or modifications of cross-entropy and KL divergence, as introduced in Sections \ref{sec:cross-entropy} and \ref{sec:kl-div}.  Even though the original losses are generally convex, the overall optimization landscape becomes non-convex due to the complex architecture of Transformers \cite{goodfellow2016deep,choi2020fine}. 
Below, we outline key probabilistic losses used in training popular LLM architectures.

\paragraph{Autoregressive Language Modeling Loss (\cont, \diff, \conv)}\label{subsubsec-transformer_autoregressive_loss}
In autoregressive language models, such as GPT \cite{radford2019language}, the model is trained to predict the next token in a sequence based on the preceding tokens. This approach is crucial for generative tasks like text generation, where the model generates text one token at a time.

The loss function used in autoregressive models is typically the cross-entropy loss, which measures the negative log-likelihood of the correct token given the model's predictions. For a sequence of tokens $(\mathbf{x}_1, \mathbf{x}_2, \cdots, \mathbf{x}_T)$, the loss is defined as:
\begin{equation}
    \mathcal{L}_{\text{AR}} = - \sum_{t=1}^{T} \log p_\theta(\mathbf{x}_t | \mathbf{x}_{<t}),
\end{equation}
where \( p_\theta(\mathbf{x}_t | \mathbf{x}_{<t}) \) is the predicted probability of token \( \mathbf{x}_t \) conditioned on previous tokens \( \mathbf{x}_{<t} \).

Autoregressive cross-entropy loss is continuous (\cont), differentiable (\diff), and convex (\conv) when considered in isolation. In LLMs, it effectively aligns model outputs with target sequences but, due to architectural complexity, contributes to an overall non-convex optimization landscape.

\paragraph{Masked Language Modeling (MLM) Loss (\cont, \diff, \conv)}\label{subsubsec-transformer_mlm_loss}
Masked language modeling, used in models like BERT \cite{devlin2018bert}, trains the model to predict randomly masked tokens within a sequence. This allows the model to leverage bidirectional context, as it relies on both left and right tokens for prediction.

The loss function for masked language modeling is also cross-entropy. For an input sequence $\mathbf{x} = (\mathbf{x}_1, \mathbf{x}_2, \cdots, \mathbf{x}_T)$ with masked tokens \( \mathbf{x}_m \), the MLM loss is:
\begin{equation}
    \mathcal{L}_{\text{MLM}} = - \sum_{m \in \text{Masked}} \log p_\theta(\mathbf{x}_m | \mathbf{x}_{\backslash m}),
\end{equation}
where \( p_\theta(\mathbf{x}_m | \mathbf{x}_{\backslash m}) \) denotes the predicted probability of a masked token given the unmasked context.

The cross-entropy loss applied here is continuous (\cont), differentiable (\diff), and convex (\conv), which suits MLM tasks well for aligning token predictions with target distributions. 

\paragraph{Label Smoothing Loss (\cont, \diff, \conv)}\label{subsubsec-transformer_label_smoothing_loss}
Label smoothing \cite{vaswani2017attention} is a regularization technique that distributes a small probability mass over incorrect classes, thus preventing the model from becoming overconfident. It’s particularly helpful in LLMs for generative tasks like machine translation and summarization.

The adjusted target labels \( y_i \) in label smoothing are given by:
\begin{equation}
    y_i = (1 - \epsilon) \cdot y_i + \frac{\epsilon}{K},
\end{equation}
where \( \epsilon \) is the smoothing parameter and \( K \) is the number of classes. The loss can then be expressed as:
\begin{equation}
    \mathcal{L}_{\text{label\_smoothing}} = - \sum_{i=1}^{K} y_i \log f_{\vec{\Theta}}(\vec{x}_i).
\end{equation}

The label-smoothing cross-entropy remains continuous (\cont), differentiable (\diff), and convex (\conv) in itself. By moderating confidence, it helps improve model generalization in tasks prone to overfitting.

\paragraph{KL Divergence Loss for Knowledge Distillation (\cont, \diff)}\label{subsubsec-transformer_kl_distillation_loss}
Knowledge distillation, commonly applied in transformer-based models like DistilBERT \cite{sanh2019distilbert}, transfers knowledge from a larger teacher model to a smaller student model. The loss function combines standard cross-entropy with a KL divergence term that minimizes the difference between the output distributions of the teacher and student models:
\begin{equation}
    \mathcal{L}_{\text{KD}} = (1 - \alpha) \mathcal{L}_{\text{CE}}(p_{\text{student}}, \vec{y}) + \alpha \mathcal{L}_{\text{KL}}(p_{\text{student}}, p_{\text{teacher}}),
\end{equation}
where \( \mathcal{L}_{\text{CE}} \) is the cross-entropy loss, \( \mathcal{L}_{\text{KL}} \) is the KL divergence between the student and teacher outputs, and \( \alpha \) controls the balance between the two terms.

The KL divergence loss is continuous (\cont) and differentiable (\diff) in itself. In knowledge distillation, it allows smaller models to approximate the behavior of larger models, improving efficiency while maintaining performance.

\subsubsection{Ranking Losses in LLM (\cont, \diff)}\label{subsubsec-transformer_ranking_losses}
In Large Language Models (LLMs), ranking losses such as Triplet Loss and Contrastive Loss are widely used to structure the embedding space for tasks like semantic search, question answering, and sentence clustering. These losses aim to pull semantically similar sentences closer together while pushing dissimilar sentences farther apart, thus learning meaningful sentence embeddings. They will be discussed in the ranking section \ref{sec:ranking} in particular in subsections \ref{sec:triplet-loss-ranking} and \ref{sec:contrastive-ranking-loss}.





\color{black}
    
\section{Ranking losses} \label{sec:ranking}  

    Machine learning can be employed to solve ranking problems, which have important industrial applications, especially in information retrieval systems. These problems can be typically solved by employing supervised, semi-supervised, or reinforcement learning~\cite{bromley1993signature,hoffer2015deep}.
    
    The goal of ranking losses, in contrast to other loss functions like the cross-entropy loss or MSE loss, is to anticipate the relative distances between inputs rather than learning to predict a label, a value, or a set of values given an input. This is also sometimes called \emph{metric learning}. Nevertheless, the cross-entropy loss can be used in the top-one probability ranking. In this scenario, given the scores of all the objects, the top-one probability of an object in this model indicates the likelihood that it will be ranked first~\cite{cao2007learning}. 

  Ranking loss functions for training data can be highly customizable because they require only a method to measure the similarity between two data points, i.e., similarity score. For example, consider a face verification dataset, pairs of photographs that belong to the same person will have a high similarity score, whereas those that don’t will have a low score ~\cite{wang2014learning}.\footnote{Different tasks, applications, and neural network configurations use ranking losses (like Siamese Nets or Triplet Nets). Because of this, there are various losses can be used, including Contrastive loss, Margin loss, Hinge loss, and Triplet loss.}

    
   In general, ranking loss functions require a feature extraction for two (or three) data instances, which returns an embedded representation for each of them. A metric function can then be defined to measure the similarity between those representations, such as the euclidean distance. Finally, the feature extractors are trained to produce similar representations for both inputs in case the inputs are similar or distant representations in case of dissimilarity.
    
\begin{figure}[H]
        \centering
        \includegraphics[width=0.8\linewidth]{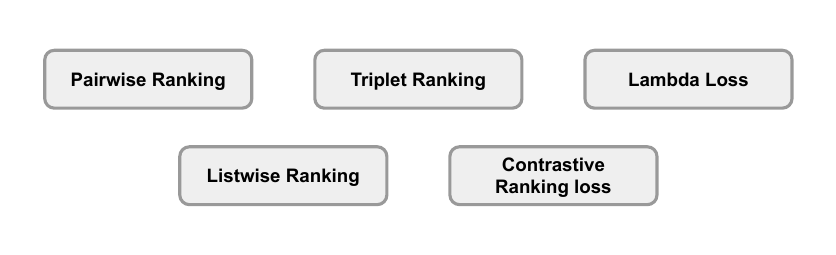}
        \caption{Overview of the ranking losses.}
        \label{fig:ranking_losses_focus}
    \end{figure}

    Similar to Section~\ref{sec:generative}, both pairwise and triplet ranking losses are presented in a general form, as shown in Fig.~\ref{fig:ranking_losses_focus}. The properties of the loss function (\cont, \diff, etc.) are identified based on the metric function chosen. 

\subsection{Pairwise Ranking loss}
    
    In the context of Pairwise Ranking loss, positive and negative pairs of training data points are used ~\cite{bromley1993signature,chopra2005learning,hadsell2006dimensionality,li2017improving}. Positive pairs are composed of an anchor sample  $\vec{x}_a$ and a positive sample $\vec{x}_p$, which is similar to $\vec{x}_a$ in the metric. Negative pairs are composed of an anchor sample $\vec{x}_a$ and a negative sample $\vec{x}_n$, which is dissimilar to $\vec{x}_a$ in that metric. The objective is to learn representations with a small distance $d$ between them for positive pairs and a greater distance than some margin value $m$ for negative pairs. Pairwise Ranking loss forces representations to have a 0 distance for positive pairs and a distance greater than a margin for negative pairs.
    
    Given $\vec{r}_a$, $\vec{r}_p$, and $\vec{r}_n$ the embedded representations (the output of a feature extractor) of the input samples $\vec{x}_a$, $\vec{x}_p$, $\vec{x}_n$ respectively and $d$ as a distance function the loss function can be written as:
    
    \begin{equation}
        {L}_{pairwise} = \begin{cases}
                                d(\vec{r}_a, \vec{r}_b) & \text{if positive pair}\\
                                \max(0,m-d(\vec{r}_a,\vec{r}_n)) &\text{if negative pair}
                                \end{cases} 
    \end{equation}
    
    For positive pairs, the loss will vanish if the distance between the embedding representations of the two elements in the pair is 0; instead, the loss will increase as the distance between the two representations increases. For negative pairs, the loss will vanish if the distance between the embedding representations of the two elements is greater than the margin $m$. However, if the distance is less than $m$, the loss will be positive, and the model parameters will be updated to provide representations for the two items that are farther apart. When the distance between $\vec{r}_a$ and $\vec{r}_n$ is 0, the loss value will be at most $m$. The purpose of the margin is to create representations for negative pairs that are far enough, thus implicitly stopping the training on these pairs and allowing the model to focus on more challenging ones. If $\vec{r}_0$ and $\vec{r}_1$ are the pair elements representations, $y$ is a binary flag equal to 0 for a negative pair and to 1 for a positive pair, and the distance $d$ is the euclidean distance:

    
    \begin{equation}
        {L}_{pairwise}(\vec{r}_0,\vec{r}_1,y) = y||\vec{r}_0-\vec{r}_1||+(1-y)\max(0,m-||\vec{r}_0-\vec{r}_1||)
    \end{equation}
Unlike typical classification learning, this loss requires more training data and time because it requires access to all the data of all potential pairs during training. Additionally, because training involves pairwise learning, it will output the binary distance from each class, which is more computationally expensive if there is incorrect classification~\cite{koch2015siamese}. 

\subsection{Triplet loss}\label{sec:triplet-loss-ranking}

Employing triplets of training data instances instead of pairs can produce better performance~\cite{hoffer2015deep,wang2014learning,chechik2010large}. This approach is called Triplet Ranking Loss. A triplet consists of an anchor sample $\vec{x}_a$, a positive sample $\vec{x}_p$, and a negative sample $\vec{x}_n$. The objective is for the distance between the anchor sample and the negative sample representations, $d(\vec{r}_a, \vec{r}_n)$, to be greater (by a margin $m$) than the distance between the anchor and positive representations $d(\vec{r}_a, \vec{r}_p)$. Here, $d$ typically represents the Euclidean distance, but other distance metrics can be applied.

The Triplet Ranking Loss is defined as:
\begin{equation}
    \mathcal{L}_{\text{triplet}}(\vec{r}_a, \vec{r}_p, \vec{r}_n) = \max(0, m + d(\vec{r}_a, \vec{r}_p) - d(\vec{r}_a, \vec{r}_n))        
\end{equation}

This loss function considers three types of triplets based on the values of $\vec{r}_a$, $\vec{r}_p$, $\vec{r}_n$, and $m$:
\begin{itemize}
    \item \textbf{Easy Triplets}: $d(\vec{r}_a, \vec{r}_n) > d(\vec{r}_a, \vec{r}_p) + m$. Here, the distance between the negative sample and the anchor sample is already large enough. The model parameters are not updated, and the loss is $0$.
    \item \textbf{Hard Triplets}: $d(\vec{r}_a, \vec{r}_n) < d(\vec{r}_a, \vec{r}_p)$. In this case, the negative sample is closer to the anchor than the positive sample. The loss is positive (and $> m$), leading to updates in the model's parameters.
    \item \textbf{Semi-Hard Triplets}: $d(\vec{r}_a, \vec{r}_p) < d(\vec{r}_a, \vec{r}_n) < d(\vec{r}_a, \vec{r}_p) + m$. Here, the negative sample is further away from the anchor than the positive sample, but the margin constraint is not yet satisfied. The loss remains positive (and $< m$), prompting parameter updates.
\end{itemize}

Triplet Ranking Loss is sensitive to small changes in the input samples, making it less generalizable across datasets~\cite{koch2015siamese}. This sensitivity arises because the model learns relative distances based on the specific sample distribution in the training data, meaning that transferring these learned distances to new data may not always yield the same effectiveness.

Triplet Loss is also used in models like Sentence-BERT (SBERT) \cite{reimers2019sentence} to learn sentence embeddings that preserve semantic similarity. A triplet consists of an anchor sentence, a positive sentence (semantically similar to the anchor), and a negative sentence (semantically dissimilar). The objective is for the anchor’s embedding to be closer to the positive sentence than to the negative sentence by a margin \( m \). This loss improves the alignment of sentence embeddings with semantic meaning, making it particularly useful for sentence similarity tasks.

\subsection{Listwise Ranking loss (\cont, \diff)}
Unlike pairwise and triplet ranking losses, which operate on pairs or triplets of data points, listwise ranking losses focus on ranking the entire list of items. One common listwise ranking loss is based on the softmax cross-entropy loss, which is used to predict the top-one ranking probability~\cite{cao2007learning}.

Given a list of items with scores, the softmax cross-entropy loss aims to maximize the probability of ranking the correct item at the top:
\begin{equation}
    L_{listwise} = - \sum_{i=1}^{N} y_i \log \frac{e^{s_i}}{\sum_{j=1}^{N} e^{s_j}},
\end{equation}
where $y_i$ is the ground truth relevance for item $i$, and $s_i$ is the predicted score for item $i$ in a list of $N$ items.

This loss is continuous (\cont) and differentiable (\diff), and it is particularly useful in information retrieval tasks, where the quality of the entire ranking list matters.

\subsection{Contrastive Ranking loss: NT-Xent (\cont, \diff, \lcont)}\label{sec:contrastive-ranking-loss}
The Normalized Temperature-Scaled Cross Entropy Loss (NT-Xent) is a generalization of pairwise ranking losses, typically used in self-supervised learning tasks~\cite{chen2020simple}. This loss enables efficient learning by comparing the similarity between an anchor and multiple negative samples in a batch.

Given an anchor sample $\vec{x}_a$ and a positive sample $\vec{x}_p$, along with $N$ negative samples $\vec{x}_n^i$, the NT-Xent loss is computed as:
\begin{equation}
    L_{\text{NT-Xent}} = - \log \frac{e^{\text{sim}(\vec{x}_a, \vec{x}_p) / \tau}}{\sum_{i=1}^{N} e^{\text{sim}(\vec{x}_a, \vec{x}_n^i) / \tau}},
\end{equation}
where $\text{sim}(\cdot, \cdot)$ is a similarity measure (e.g., cosine similarity), and $\tau$ is a temperature scaling parameter.

This loss is continuous (\cont), differentiable (\diff), and Lipschitz continuous (\lcont), and it is often used in contrastive learning tasks where learning representations from large-scale data with many negatives is necessary.

This loss is also been used in language models like SimCSE \cite{gao2021simcse} to maximize the similarity between an anchor and a positive sample while minimizing the similarity with negative samples. 
One of the primary advantages of contrastive losses in the context of language modelling is their ability to improve the generalization of sentence embeddings by providing a strong supervision signal, particularly in tasks requiring semantic similarity \cite{gao2021simcse}. Additionally, using multiple negatives in the mini-batch, as with NT-Xent, improves training efficiency by leveraging more negative examples \cite{reimers2019sentence}, leading to faster convergence and more stable training compared to methods with fewer negatives.

\subsubsection{LambdaLoss (\cont, \diff)}\label{subsubsec-ranking_lambda_loss}
LambdaLoss \cite{burges2010ranknet,burges2006learning} is a specialized loss function designed for learning-to-rank tasks, where the objective is to optimize ranking metrics such as Normalized Discounted Cumulative Gain (NDCG) or Mean Reciprocal Rank (MRR). It is particularly effective in applications like search engines and recommendation systems, where errors in top-ranked items are more costly.

LambdaLoss differs from traditional pairwise losses by directly optimizing ranking metrics like NDCG, and weighting errors by their impact on the final ranking. The loss function considers the relative change in the ranking metric when swapping the predicted rankings of two items.

Given two documents \( i \) and \( j \), with ground truth labels \( y_i \) and \( y_j \), and predicted scores \( f(\mathbf{x}_i) \) and \( f(\mathbf{x}_j) \), LambdaLoss is defined as:
\begin{equation}
    L_{\text{Lambda}} = |\Delta \text{NDCG}_{ij}| \cdot \sigma(f(\mathbf{x}_i) - f(\mathbf{x}_j)),
\end{equation}
where:
\begin{itemize}
    \item \( \Delta \text{NDCG}_{ij} \) represents the change in NDCG if the documents \( i \) and \( j \) are swapped in the ranking.
    \item \( \sigma(f(\mathbf{x}_i) - f(\mathbf{x}_j)) \) is the pairwise ranking loss, typically modeled as a logistic function \( \sigma(z) = \frac{1}{1 + e^{-z}} \) or a hinge loss.
\end{itemize}

The total LambdaLoss \( \mathcal{L}_{\text{Lambda}} \) for a dataset is computed as the sum of all pairwise losses across all document pairs:
\begin{equation}
    \mathcal{L}_{\text{Lambda}} = \sum_{i < j} L_{\text{Lambda}},
\end{equation}
where \( i \) and \( j \) index the document pairs in the dataset.

\( L_{\text{Lambda}} \) is continuous (\cont) and differentiable (\diff), making it suitable for gradient-based optimization. The loss function directly optimizes ranking metrics such as NDCG, ensuring that the model's training objective aligns with the final evaluation metric. LambdaLoss emphasizes higher-ranked items, as errors in top positions are penalized more heavily due to their greater impact on the overall ranking.

LambdaLoss is commonly used in systems that require high-quality rankings, such as search engines, ad placement, and recommendation systems.

\color{black}
\section{Energy-based losses}   
\label{sec:ebm}

An Energy-Based Model (EBM) is a probabilistic model that uses a scalar energy function to describe the dependencies of the model variables~\cite{lecun2006tutorial,FRISTON200670,Friston2009,finn2016connection,haarnoja2017reinforcement,grathwohl2019your,Du2019ModelBP}. An EBM can be formalised as ${F}: \mathcal{X} \times \mathcal{Y}\rightarrow \mathbb{R}$, where ${F}(\vec{x},\vec{y})$ stands for the relationship between the $(\vec{x},\vec{y})$ pairings. 

Given an energy function and the input $\vec{x}$, the best fitting value of $\vec{y}$ is computed with the following inference procedure: 
\begin{equation}\label{EMB_def}
    \tilde{\vec{y}} = argmin_{\vec{y}} \{F(\vec{x},\svec{y})\}
\end{equation}



Energy-based models provide fully generative models that can be used as an alternative to probabilistic estimation for prediction, classification, or decision-making tasks~\cite{osadchy2004synergistic,lecun2006tutorial,haarnoja2017reinforcement,du2019implicit,grathwohl2019your}.

The energy function $E(\vec{x},\vec{y}) \equiv F(\vec{x},\vec{y})$ can be explicitly defined for all the values of $\vec{y}\in\mathcal{Y}=\svec{y}$ if and only if the size of the set $\mathcal{Y}$ is small enough. In contrast, when the space of $\mathcal{Y}$ is sufficiently large, a specific strategy, known as the inference procedure, must be employed to find the $\vec{y}$ that minimizes $E(\vec{x}, \svec{y})$. 

In many real situations, the inference procedure can produce an approximate result, which may or may not be the global minimum of $E(\vec{x}, \svec{y})$ for a given $\vec{x}$. Moreover, it is possible that $E(\vec{x}, \svec{y})$ has several equivalent minima. The best inference procedure to use often depends on the internal structure of the model. For example, if $\mathcal{Y}$ is continuous and $E(\vec{x}, \svec{y})$ is smooth and differentiable everywhere concerning $\vec{y}$, a gradient-based optimization algorithm can be employed ~\cite{bengio2000gradient}. 

In general, any probability density function $p(\vec{x})$ for $\vec{x}\in\mathbb{R}^D$ can be rewritten as an EBM:
\begin{equation}
\label{eq:ebm_basic}
 p_\vtheta(\vec{x}) = \frac{\exp(-E_\vtheta(\vec{x}))}{\int_{\vec{x}’} \exp(-E_\vtheta(\vec{x}’)) d\vec{x}’},
\end{equation}
where the energy function ($E_\vtheta$) can be any function parameterised by $\vtheta \in \vec{\Theta}$ (such as a  neural network). In these models, a prediction (e.g. finding $p(\vec{x_0}| \vec{x_1},\vec{x_2},...)$) is done by fixing the values of the conditional variables, and estimating the remaining variables, (e.g. $\vec{x_0}$), by minimizing the energy function \cite{teh2003energy,swersky2011autoencoders,zhai2016deep}. 

An EBM is trained by finding an energy function that associates low energies to values of $\vec{x}$ drawn from the underlying data distribution, $p_\theta(\vec{x})\sim p_D(\vec{x})$, and high energies for values of $\vec{x}$ not close to the underlying distribution.


\subsection{Training EBM}
Given the aforementioned conceptual framework, the training can be thought as finding the model parameters that define the good match between the output ($\vec{y}\in\mathcal{Y}$) and the input ($\vec{x}\in \mathcal{X}$) for every step. This is done by estimating the best energy function from the set of energy functions ($\mathcal{E}$) by scanning all the model parameters $\vec{\vec{\Theta}}$ \cite{kumar2019maximum,song2021train}, where $\mathcal{E} = \{F(\vec{\Theta},\vec{X},\vec{Y}): \vec{\Theta}\in\mathcal{W}=\vec{\Theta}\}$.
A loss function should behave similarly to the energy function described in Equation~\ref{EMB_def}, i.e., lower energy for a correct answer must be modeled by low losses, instead, higher energy, to all incorrect answers, by a higher loss.

Considering a set of training samples $\mathcal{S} = \{(\vec{x}_i, \vec{y}_i): i=1, \cdots, N\}$, during the training procedure, the loss function should have the effect of pushing-down $E(\vec{\Theta},\vec{y}_i,\vec{x}_i)$ and pulling-up $E(\vec{\Theta}, \tilde{\vec{y}}_i,\vec{x}_i)$, i.e., finding the parameters $\vec{\Theta}$ that minimize the loss: 
\begin{equation}
    \vec{\Theta}^* = min_{\vec{\Theta}\in\mathcal{W}}L_{ebm}(\vec{\Theta},\mathcal{S})
\end{equation}

The general form of the loss function $\mathcal{L}_{ebm}$ is defined as: 

\begin{equation}
    \mathcal{L}_{ebm}(\vec{\Theta},\mathcal{S}) = \frac{1}{N}\sum_{i=1}^N L_{ebm}(\vec{y}_i, E(\vec{\Theta},\mathcal{Y},\vec{x}_i)) + \text{regularizer term}
\end{equation}

Where: 
\begin{itemize}
    \item $L_{ebm}(\vec{y}_i, E(\vec{\Theta},\mathcal{Y},\vec{x}_i))$ is the per-sample loss
    \item $\vec{y}_i$ is the desired output
    \item $E(\vec{\Theta},\mathcal{Y},\vec{x}_i))$ is energy surface for a given $\vec{x}_i$ as $\vec{y} \in \mathcal{Y}$ varies
\end{itemize}

This is an average of a per-sample loss functional, denoted $L_{ebm}(\vec{y}_i, E(\vec{\Theta},\mathcal{Y},\vec{x}_i))$, over the training set. This function depends on the desired output $\vec{y}_i$ and on the energies derived by holding the input sample $\vec{x}_i$ constant and changing the output scanning over the sample $\mathcal{Y}$. 
With this definition, the loss remains unchanged when the training samples are mixed up and when the training set is repeated numerous times~\cite{lecun2006tutorial}. As the size of the training set grows, the model is less likely to overfit~\cite{vapnik1999nature}.

\subsection{Loss Functions for EBMs}
Following Fig.~\ref{fig:energy_based_losses_focus}, in this section, we introduce the energy loss first since it is the most straightforward. Then we present common losses in machine learning that can be adapted to the energy-based models, such as the Negative Log-Likelihood loss, the Hinge loss, and the Log loss. Subsequently, we introduce more sophisticated losses like the Generalized Perceptron loss and the Generalized Margin loss. Finally, the Minimum classification error loss, the Square-square loss, and its variation Square-exponential loss are presented.

\begin{figure}[H]
    \centering
    \includegraphics[width=0.6\linewidth]{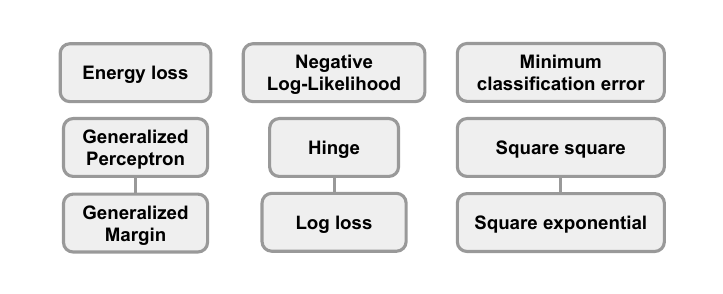}
    \caption{Schematic overview of the energy-based losses with their connection.}
    \label{fig:energy_based_losses_focus}
\end{figure}

\subsubsection{Energy loss}
The so-called energy loss is the most straightforward loss due to its simplicity. It can be simply defined by using the energy function as the per-sample loss: 

\begin{equation}\label{Eenegy_loss}
    {L}_{energy} (\vec{y}_i, E(\vec{\Theta}, \mathcal{Y}, \vec{x}_i)) = E(\vec{\Theta}, \vec{x}_i, \vec{y}_i)
\end{equation} 

This loss is often used in regression tasks. According to its definition, it pulls the energy function down for values that are close to the correct data distribution. However, the energy function is not pulled up for incorrect values. The assumption is that, by lowering the energy function in the correct location, the energy for incorrect values is left higher as a result. Due to this assumption, the training is sensitive to the model design and may result in energy collapse, leading to a largely flat energy function.


\subsubsection{Generalized Perceptron loss (\lcont, \conv)}
The Generalized Perceptron loss is defined as: 

\begin{equation}\label{GPL}
    {L}_{perceptron} (\vec{y}_i, E(\vec{\Theta}, \mathcal{Y}, \vec{x}_i)) = E(\vec{\Theta}, \vec{y}_i, \vec{x}_i) - [\min_{\vec{y}\in \mathcal{Y}}] \{E(\vec{\Theta},\svec{y},\vec{x}_i)\}
\end{equation} 
This loss is positive definite as the second term is the lower bound of the first one, i.e., $E(\vec{\Theta}, \vec{y}_i, \vec{x}_i) - [min_{\vec{y}\in \mathcal{Y}}] \{E(\vec{\Theta},\svec{y},\vec{x}_i)\} \ge 0$. By minimizing this loss, the first term is pushed down and the energy of the model prediction is raised. Although it is widely used~\cite{lecun1998gradient,collins2002discriminative}, this loss is suboptimal as it does not detect the gap between the correct output and the incorrect ones, and it does not restrict the function from assigning the same value to each wrong output $\vec{y}_i$ and it may produce flat energy distributions ~\cite{lecun2006tutorial}.

\subsubsection{Negative Log-Likelihood loss (\cont,\diff,\conv)}
In analogy with the description in Section~\ref{sec:cross-entropy}, the Negative Log-Likelihood loss (NLL) in the energy-based context is defined as: 

\begin{equation}
    \label{E_nll}
    \mathcal{L}_{NLL} (\vec{\Theta}, \mathcal{S}) = \frac{1}{N} \sum_{i=1}^N \left(E(\vec{\Theta}, \vec{y}_i, \vec{x}_i) + \frac{1}{\beta} \log \int_{\vec{y}\in\mathcal{Y}} e^{\beta E(\vec{\Theta},\vec{y},\vec{x}_i)}\right)
\end{equation}

where $\mathcal{S}$ is the training set.

This loss reduces to the perceptron loss when $\beta\rightarrow\infty$ and to the log loss in case $\mathcal{Y}$ has only two labels (i.e., binary classification). Since the integral above is intractable, considerable efforts have been devoted to finding approximation methods, including Monte-Carlo sampling methods~\cite{shapiro2003monte}, and variational methods~\cite{jordan1999introduction}. While these methods have been devised as approximate ways of minimizing the NLL loss, they can be viewed in the energy-based framework as different strategies for choosing the $\vec{y}$’s whose energies will be pulled up~\cite{lecun2005loss,lecun2006tutorial}. 
The NLL is also known as the cross-entropy~\cite{levin1988accelerated} loss and is widely used in many applications, including energy-based models~\cite{bengio2000neural,bengio1992global}.
This loss function formulation is subject to the same limitations listed in section~\ref{sec:cross-entropy}.

\subsubsection{Generalized Margin loss}
The generalized margin loss is a more reliable version of the generalized perceptron loss. The general form of the generalized margin loss in the context of energy-based training is defined as:
\begin{equation}
    L_{margin}(\vec{\Theta},\vec{x}_i,\vec{y}_i) = {Q}_m(E(\vec{\Theta},\vec{x}_i,\vec{y}_i), E(\vec{\Theta},\vec{x}_i,\bar{\vec{y}}_i))
\end{equation}
Where $\bar{\vec{y}}_i$ is the so-called \textit{"most-offending incorrect output"} which is
the output that has the lowest energy among all possible outputs that are incorrect~\cite{lecun2006tutorial}, $m$ is a positive margin parameter, and $Q_m$ is a convex function which ensures that the loss receives low values for $E(\vec{\Theta},\vec{x}_i,\vec{y}_i)$ and high values for $E(\vec{\Theta},\vec{x}_i,\bar{\vec{y}}_i)$. In other words, the loss function can ensure that the energy of the most offending incorrect output is greater by some arbitrary margin than the energy of the correct output.

This loss function is written in the general form and a wide variety of losses that use specific margin function $Q_m$ to produce a gap between the correct output and the wrong output are formalised in the following part of the section.
\paragraph{Hinge loss (\lcont,\conv)}
Already explained in section~\ref{subsubsec-classification_perceptron_loss}, the hinge loss can be rewritten as: 
\begin{equation}
    L_{hinge}(\vec{\Theta},\vec{x}_i,\vec{y}_i) = max(0, m+ E(\vec{\Theta},\vec{x}_i,\vec{y}_i) -  E(\vec{\Theta},\vec{x}_i,\bar{\vec{y}}_i))
\end{equation}
This loss enforces that the difference between the correct answer and the most offending incorrect answer be at least $m$~\cite{taskar2003max,altun2003hidden}. Individual energies are not required to take a specific value because the hinge loss depends on energy differences.
This loss function shares limitations with the original Hinge loss defined in eq. \ref{eq:hinge}.
\paragraph{Log loss (\diff,\cont,\conv)}
This loss is similar to the hinge loss, but it sets a softer margin between the correct output and the most offending outputs. The log loss is defined as: 
\begin{equation}
    L_{log}(\vec{\Theta},\vec{x}_i,\vec{y}_i) = \log (1+e^{E(\vec{\Theta},\vec{x}_i,\vec{y}_i) -  E(\vec{\Theta},\vec{x}_i,\bar{\vec{y}}_i)})
\end{equation}
This loss is also called soft hinge and it may produce overfitting on high dimensional datasets ~\cite{kleinbaum2002logistic}.

\paragraph{Minimum classification error loss (\cont, \diff, \conv)}
A straightforward function that roughly counts the total number of classification errors while being smooth and differentiable is known as the Minimum Classification Error (MCE) loss~\cite{juang1997minimum}. The MCE is written as a sigmoid function: 
\begin{equation}
    L_{mce}(\vec{\Theta},\vec{x}_i,\vec{y}_i) = \sigma(E(\vec{\Theta},\vec{x}_i,\vec{y}_i) -  E(\vec{\Theta},\vec{x}_i,\bar{\vec{y}}_i))
\end{equation}
Where $\sigma$ is defined as $\sigma(x)=(1+e^{-x})^{-1}$. While this function lacks an explicit margin, it nevertheless produces an energy difference between the most offending incorrect output and the correct output.

\paragraph{Square-square loss (\cont,\conv)}
Square-square loss deals differently with the energy of the correct output $E(\vec{\Theta},\vec{x}_i,\vec{y}_i)$ and the energy of the most offensive output $E(\vec{\Theta},\vec{x}_i,\bar{\vec{y}}_i)$ as:
\begin{equation}
    L_{sq-sq} (\vec{\Theta},\vec{x}_i,\vec{y}_i) = E(\vec{\Theta},\vec{x}_i,\vec{y}_i)^2 + (\max(0,m-E(\vec{\Theta},\vec{x}_i,\bar{\vec{y}}_i)))^2
\end{equation}
The combination aims to minimize the energy of the correct output while enforcing a margin of at least $m$ on the most offending incorrect outputs.
This loss is a modified version of the margin loss. This loss can be only used when there is a lower bound on the energy function ~\cite{lecun2005loss,hadsell2006dimensionality}.
\paragraph{Square-exponential loss (\cont, \diff, \conv)}
This loss is similar to the square-square loss function, and it only differs in the second term: 
\begin{equation}
    L_{sq-exp} (\vec{\Theta},\vec{x}_i,\vec{y}_i) = E(\vec{\Theta},\vec{x}_i,\vec{y}_i)^2 + \gamma e^{-E(\vec{\Theta},\vec{x}_i,\bar{\vec{y}}_i)}
\end{equation}
While $\gamma$ is a positive constant, the combination aims to minimize the energy of the correct output while pushing the energy of the most offending incorrect output to an infinite margin~\cite{lecun2005loss,chopra2005learning,osadchy2004synergistic}.
This loss is considered a regularized version of the aforementioned square-square loss. This loss, as for the Square-square loss, can be only used when there is a lower bound on the energy function ~\cite{lecun2005loss,hadsell2006dimensionality}.
\section{Conclusion} \label{sec:conclusion}

The choice of an appropriate loss function is fundamental to the success of machine learning applications. In this survey, we have comprehensively reviewed and detailed 43 of the most commonly used loss functions, covering a diverse range of machine learning tasks, including classification, regression, generative modeling, ranking, and energy-based modeling. Each loss function has been contextualized within its domain of application, highlighting both its advantages and limitations to guide informed selection based on specific problem requirements.

This survey is intended as a resource for machine learning practitioners at all levels, including undergraduate, graduate, and Ph.D. students, as well as researchers working to deepen their understanding of loss functions or to develop new ones. By structuring each loss function within a novel taxonomy, presented in Sec.~\ref{sec:definition}, we have created an intuitive framework that categorizes these functions based on their underlying principles, problem domains, and methodological approaches. This taxonomy provides readers with a clear view of the relationships among different techniques, enabling them to quickly identify the most suitable loss function for their task while also facilitating comparison across methods.

We hope that this structured view of loss functions will not only streamline the selection process for practitioners but also support the conceptual development of this field.

\bibliography{main.bib}

\end{document}

%% file: utils.tex
\usepackage{xfrac}
\usepackage{bm}
\usepackage{xspace}
\usepackage{mathrsfs}
\usepackage{amsmath,amsfonts,bm}
\usepackage{mathtools}

\newcommand{\norm}[1]{\left\lVert#1\right\rVert}

\def\eqref#1{equation~(\ref{#1})}



%% file: main.bbl

\begin{thebibliography}{160}
\ifx \bisbn   \undefined \def \bisbn  #1{ISBN #1}\fi
\ifx \binits  \undefined \def \binits#1{#1}\fi
\ifx \bauthor  \undefined \def \bauthor#1{#1}\fi
\ifx \batitle  \undefined \def \batitle#1{#1}\fi
\ifx \bjtitle  \undefined \def \bjtitle#1{#1}\fi
\ifx \bvolume  \undefined \def \bvolume#1{\textbf{#1}}\fi
\ifx \byear  \undefined \def \byear#1{#1}\fi
\ifx \bissue  \undefined \def \bissue#1{#1}\fi
\ifx \bfpage  \undefined \def \bfpage#1{#1}\fi
\ifx \blpage  \undefined \def \blpage #1{#1}\fi
\ifx \burl  \undefined \def \burl#1{\textsf{#1}}\fi
\ifx \doiurl  \undefined \def \doiurl#1{\url{https://doi.org/#1}}\fi
\ifx \betal  \undefined \def \betal{\textit{et al.}}\fi
\ifx \binstitute  \undefined \def \binstitute#1{#1}\fi
\ifx \binstitutionaled  \undefined \def \binstitutionaled#1{#1}\fi
\ifx \bctitle  \undefined \def \bctitle#1{#1}\fi
\ifx \beditor  \undefined \def \beditor#1{#1}\fi
\ifx \bpublisher  \undefined \def \bpublisher#1{#1}\fi
\ifx \bbtitle  \undefined \def \bbtitle#1{#1}\fi
\ifx \bedition  \undefined \def \bedition#1{#1}\fi
\ifx \bseriesno  \undefined \def \bseriesno#1{#1}\fi
\ifx \blocation  \undefined \def \blocation#1{#1}\fi
\ifx \bsertitle  \undefined \def \bsertitle#1{#1}\fi
\ifx \bsnm \undefined \def \bsnm#1{#1}\fi
\ifx \bsuffix \undefined \def \bsuffix#1{#1}\fi
\ifx \bparticle \undefined \def \bparticle#1{#1}\fi
\ifx \barticle \undefined \def \barticle#1{#1}\fi
\bibcommenthead
\ifx \bconfdate \undefined \def \bconfdate #1{#1}\fi
\ifx \botherref \undefined \def \botherref #1{#1}\fi
\ifx \url \undefined \def \url#1{\textsf{#1}}\fi
\ifx \bchapter \undefined \def \bchapter#1{#1}\fi
\ifx \bbook \undefined \def \bbook#1{#1}\fi
\ifx \bcomment \undefined \def \bcomment#1{#1}\fi
\ifx \oauthor \undefined \def \oauthor#1{#1}\fi
\ifx \citeauthoryear \undefined \def \citeauthoryear#1{#1}\fi
\ifx \endbibitem  \undefined \def \endbibitem {}\fi
\ifx \bconflocation  \undefined \def \bconflocation#1{#1}\fi
\ifx \arxivurl  \undefined \def \arxivurl#1{\textsf{#1}}\fi
\csname PreBibitemsHook\endcsname

\bibitem[\protect\citeauthoryear{Mitchell et~al.}{1990}]{ML_trends}
\begin{barticle}
\bauthor{\bsnm{Mitchell}, \binits{T.}},
\bauthor{\bsnm{Buchanan}, \binits{B.}},
\bauthor{\bsnm{DeJong}, \binits{G.}},
\bauthor{\bsnm{Dietterich}, \binits{T.}},
\bauthor{\bsnm{Rosenbloom}, \binits{P.}},
\bauthor{\bsnm{Waibel}, \binits{A.}}:
\batitle{Machine learning}.
\bjtitle{Annual Review of Computer Science}
\bvolume{4}(\bissue{1}),
\bfpage{417}--\blpage{433}
(\byear{1990})
\doiurl{10.1146/annurev.cs.04.060190.002221}
\end{barticle}
\endbibitem

\bibitem[\protect\citeauthoryear{Jordan and Mitchell}{2015}]{ML_trends2}
\begin{barticle}
\bauthor{\bsnm{Jordan}, \binits{M.I.}},
\bauthor{\bsnm{Mitchell}, \binits{T.M.}}:
\batitle{Machine learning: Trends, perspectives, and prospects}.
\bjtitle{Science}
\bvolume{349}(\bissue{6245}),
\bfpage{255}--\blpage{260}
(\byear{2015})
\end{barticle}
\endbibitem

\bibitem[\protect\citeauthoryear{Mitchell}{1997}]{ML_2}
\begin{bbook}
\bauthor{\bsnm{Mitchell}, \binits{T.M.}}:
\bbtitle{Machine Learning}
vol. \bseriesno{1}.
\bpublisher{McGraw-Hill Education},
\blocation{New York}
(\byear{1997})
\end{bbook}
\endbibitem

\bibitem[\protect\citeauthoryear{Mahesh}{2020}]{ML_1}
\begin{barticle}
\bauthor{\bsnm{Mahesh}, \binits{B.}}:
\batitle{Machine learning algorithms-a review}.
\bjtitle{International Journal of Science and Research (IJSR).[Internet]}
\bvolume{9},
\bfpage{381}--\blpage{386}
(\byear{2020})
\end{barticle}
\endbibitem

\bibitem[\protect\citeauthoryear{Quinlan}{1986}]{decision_trees}
\begin{barticle}
\bauthor{\bsnm{Quinlan}, \binits{J.R.}}:
\batitle{Induction of decision trees}.
\bjtitle{Machine learning}
\bvolume{1}(\bissue{1}),
\bfpage{81}--\blpage{106}
(\byear{1986})
\end{barticle}
\endbibitem

\bibitem[\protect\citeauthoryear{Bishop et~al.}{1995}]{bishop1995neural}
\begin{bbook}
\bauthor{\bsnm{Bishop}, \binits{C.M.}}, \betal:
\bbtitle{Neural Networks for Pattern Recognition}.
\bpublisher{Oxford university press},
\blocation{Oxford}
(\byear{1995})
\end{bbook}
\endbibitem

\bibitem[\protect\citeauthoryear{Shally and RR}{}]{survey_biomed}
\begin{botherref}
\oauthor{\bsnm{Shally}, \binits{H.}},
\oauthor{\bsnm{RR}, \binits{R.R.}}:
Survey for mining biomedic
\end{botherref}
\endbibitem

\bibitem[\protect\citeauthoryear{Patil et~al.}{2020}]{ml_biomed}
\begin{barticle}
\bauthor{\bsnm{Patil}, \binits{S.}},
\bauthor{\bsnm{Patil}, \binits{K.R.}},
\bauthor{\bsnm{Patil}, \binits{C.R.}},
\bauthor{\bsnm{Patil}, \binits{S.S.}}:
\batitle{Performance overview of an artificial intelligence in biomedics: a
  systematic approach}.
\bjtitle{International Journal of Information Technology}
\bvolume{12}(\bissue{3}),
\bfpage{963}--\blpage{973}
(\byear{2020})
\end{barticle}
\endbibitem

\bibitem[\protect\citeauthoryear{Zhang et~al.}{2019}]{brain_interface_survey}
\begin{botherref}
\oauthor{\bsnm{Zhang}, \binits{X.}},
\oauthor{\bsnm{Yao}, \binits{L.}},
\oauthor{\bsnm{Wang}, \binits{X.}},
\oauthor{\bsnm{Monaghan}, \binits{J.}},
\oauthor{\bsnm{Mcalpine}, \binits{D.}},
\oauthor{\bsnm{Zhang}, \binits{Y.}}:
A survey on deep learning based brain computer interface: Recent advances and
  new frontiers.
arXiv preprint arXiv:1905.04149
\textbf{66}
(2019)
\end{botherref}
\endbibitem

\bibitem[\protect\citeauthoryear{Kourou et~al.}{2015}]{cancer_ml}
\begin{barticle}
\bauthor{\bsnm{Kourou}, \binits{K.}},
\bauthor{\bsnm{Exarchos}, \binits{T.P.}},
\bauthor{\bsnm{Exarchos}, \binits{K.P.}},
\bauthor{\bsnm{Karamouzis}, \binits{M.V.}},
\bauthor{\bsnm{Fotiadis}, \binits{D.I.}}:
\batitle{Machine learning applications in cancer prognosis and prediction}.
\bjtitle{Computational and structural biotechnology journal}
\bvolume{13},
\bfpage{8}--\blpage{17}
(\byear{2015})
\end{barticle}
\endbibitem

\bibitem[\protect\citeauthoryear{Chowdhary}{2020}]{nlp_1}
\begin{botherref}
\oauthor{\bsnm{Chowdhary}, \binits{K.}}:
Natural language processing.
Fundamentals of artificial intelligence,
603--649
(2020)
\end{botherref}
\endbibitem

\bibitem[\protect\citeauthoryear{Otter et~al.}{2020}]{nlp_2}
\begin{barticle}
\bauthor{\bsnm{Otter}, \binits{D.W.}},
\bauthor{\bsnm{Medina}, \binits{J.R.}},
\bauthor{\bsnm{Kalita}, \binits{J.K.}}:
\batitle{A survey of the usages of deep learning for natural language
  processing}.
\bjtitle{IEEE transactions on neural networks and learning systems}
\bvolume{32}(\bissue{2}),
\bfpage{604}--\blpage{624}
(\byear{2020})
\end{barticle}
\endbibitem

\bibitem[\protect\citeauthoryear{Shaukat
  et~al.}{2020}]{computer_security_survey}
\begin{barticle}
\bauthor{\bsnm{Shaukat}, \binits{K.}},
\bauthor{\bsnm{Luo}, \binits{S.}},
\bauthor{\bsnm{Varadharajan}, \binits{V.}},
\bauthor{\bsnm{Hameed}, \binits{I.A.}},
\bauthor{\bsnm{Xu}, \binits{M.}}:
\batitle{A survey on machine learning techniques for cyber security in the last
  decade}.
\bjtitle{IEEE Access}
\bvolume{8},
\bfpage{222310}--\blpage{222354}
(\byear{2020})
\end{barticle}
\endbibitem

\bibitem[\protect\citeauthoryear{Chandola
  et~al.}{2009}]{anomaly_detection_survey}
\begin{barticle}
\bauthor{\bsnm{Chandola}, \binits{V.}},
\bauthor{\bsnm{Banerjee}, \binits{A.}},
\bauthor{\bsnm{Kumar}, \binits{V.}}:
\batitle{Anomaly detection: A survey}.
\bjtitle{ACM computing surveys (CSUR)}
\bvolume{41}(\bissue{3}),
\bfpage{1}--\blpage{58}
(\byear{2009})
\end{barticle}
\endbibitem

\bibitem[\protect\citeauthoryear{Lu and
  Weng}{2007}]{image_classification_survey}
\begin{barticle}
\bauthor{\bsnm{Lu}, \binits{D.}},
\bauthor{\bsnm{Weng}, \binits{Q.}}:
\batitle{A survey of image classification methods and techniques for improving
  classification performance}.
\bjtitle{International journal of Remote sensing}
\bvolume{28}(\bissue{5}),
\bfpage{823}--\blpage{870}
(\byear{2007})
\end{barticle}
\endbibitem

\bibitem[\protect\citeauthoryear{Frawley et~al.}{1992}]{frawley1992knowledge}
\begin{barticle}
\bauthor{\bsnm{Frawley}, \binits{W.J.}},
\bauthor{\bsnm{Piatetsky-Shapiro}, \binits{G.}},
\bauthor{\bsnm{Matheus}, \binits{C.J.}}:
\batitle{Knowledge discovery in databases: An overview}.
\bjtitle{AI magazine}
\bvolume{13}(\bissue{3}),
\bfpage{57}--\blpage{57}
(\byear{1992})
\end{barticle}
\endbibitem

\bibitem[\protect\citeauthoryear{Argall et~al.}{2009}]{argall2009survey}
\begin{barticle}
\bauthor{\bsnm{Argall}, \binits{B.D.}},
\bauthor{\bsnm{Chernova}, \binits{S.}},
\bauthor{\bsnm{Veloso}, \binits{M.}},
\bauthor{\bsnm{Browning}, \binits{B.}}:
\batitle{A survey of robot learning from demonstration}.
\bjtitle{Robotics and autonomous systems}
\bvolume{57}(\bissue{5}),
\bfpage{469}--\blpage{483}
(\byear{2009})
\end{barticle}
\endbibitem

\bibitem[\protect\citeauthoryear{Perlich et~al.}{2014}]{online_advertising}
\begin{barticle}
\bauthor{\bsnm{Perlich}, \binits{C.}},
\bauthor{\bsnm{Dalessandro}, \binits{B.}},
\bauthor{\bsnm{Raeder}, \binits{T.}},
\bauthor{\bsnm{Stitelman}, \binits{O.}},
\bauthor{\bsnm{Provost}, \binits{F.}}:
\batitle{Machine learning for targeted display advertising: Transfer learning
  in action}.
\bjtitle{Machine learning}
\bvolume{95}(\bissue{1}),
\bfpage{103}--\blpage{127}
(\byear{2014})
\end{barticle}
\endbibitem

\bibitem[\protect\citeauthoryear{Bontempi
  et~al.}{2012}]{time_series_forecasting_ml}
\begin{bchapter}
\bauthor{\bsnm{Bontempi}, \binits{G.}},
\bauthor{\bsnm{Ben~Taieb}, \binits{S.}},
\bauthor{\bsnm{Borgne}, \binits{Y.-A.L.}}:
\bctitle{Machine learning strategies for time series forecasting}.
In: \bbtitle{European Business Intelligence Summer School},
pp. \bfpage{62}--\blpage{77}
(\byear{2012}).
\bcomment{Springer}
\end{bchapter}
\endbibitem

\bibitem[\protect\citeauthoryear{M{\"u}ller
  et~al.}{2004}]{brain_computer_interfaces}
\begin{barticle}
\bauthor{\bsnm{M{\"u}ller}, \binits{K.-R.}},
\bauthor{\bsnm{Krauledat}, \binits{M.}},
\bauthor{\bsnm{Dornhege}, \binits{G.}},
\bauthor{\bsnm{Curio}, \binits{G.}},
\bauthor{\bsnm{Blankertz}, \binits{B.}}:
\batitle{Machine learning techniques for brain-computer interfaces}.
\bjtitle{Biomed. Tech}
\bvolume{49}(\bissue{1}),
\bfpage{11}--\blpage{22}
(\byear{2004})
\end{barticle}
\endbibitem

\bibitem[\protect\citeauthoryear{Shinde and Shah}{2018}]{other_applications}
\begin{bchapter}
\bauthor{\bsnm{Shinde}, \binits{P.P.}},
\bauthor{\bsnm{Shah}, \binits{S.}}:
\bctitle{A review of machine learning and deep learning applications}.
In: \bbtitle{2018 Fourth International Conference on Computing Communication
  Control and Automation (ICCUBEA)},
pp. \bfpage{1}--\blpage{6}
(\byear{2018}).
\bcomment{IEEE}
\end{bchapter}
\endbibitem

\bibitem[\protect\citeauthoryear{Von~Neumann and
  Morgenstern}{2007}]{von2007theory}
\begin{bbook}
\bauthor{\bsnm{Von~Neumann}, \binits{J.}},
\bauthor{\bsnm{Morgenstern}, \binits{O.}}:
\bbtitle{Theory of Games and Economic Behavior}.
\bpublisher{Princeton University Press},
\blocation{New Jersey}
(\byear{2007})
\end{bbook}
\endbibitem

\bibitem[\protect\citeauthoryear{Wang et~al.}{2020}]{wang2020comprehensive}
\begin{botherref}
\oauthor{\bsnm{Wang}, \binits{Q.}},
\oauthor{\bsnm{Ma}, \binits{Y.}},
\oauthor{\bsnm{Zhao}, \binits{K.}},
\oauthor{\bsnm{Tian}, \binits{Y.}}:
A comprehensive survey of loss functions in machine learning.
Annals of Data Science,
1--26
(2020)
\end{botherref}
\endbibitem

\bibitem[\protect\citeauthoryear{Wang et~al.}{2021}]{wang2021survey}
\begin{barticle}
\bauthor{\bsnm{Wang}, \binits{J.}},
\bauthor{\bsnm{Feng}, \binits{S.}},
\bauthor{\bsnm{Cheng}, \binits{Y.}},
\bauthor{\bsnm{Al-Nabhan}, \binits{N.}}:
\batitle{Survey on the loss function of deep learning in face recognition}.
\bjtitle{Journal of Information Hiding and Privacy Protection}
\bvolume{3}(\bissue{1}),
\bfpage{29}
(\byear{2021})
\end{barticle}
\endbibitem

\bibitem[\protect\citeauthoryear{Jadon}{2020}]{jadon2020survey}
\begin{bchapter}
\bauthor{\bsnm{Jadon}, \binits{S.}}:
\bctitle{A survey of loss functions for semantic segmentation}.
In: \bbtitle{2020 IEEE Conference on Computational Intelligence in
  Bioinformatics and Computational Biology (CIBCB)},
pp. \bfpage{1}--\blpage{7}
(\byear{2020}).
\bcomment{IEEE}
\end{bchapter}
\endbibitem

\bibitem[\protect\citeauthoryear{Virmaux and
  Scaman}{2018}]{virmaux2018lipschitz}
\begin{botherref}
\oauthor{\bsnm{Virmaux}, \binits{A.}},
\oauthor{\bsnm{Scaman}, \binits{K.}}:
Lipschitz regularity of deep neural networks: analysis and efficient
  estimation.
Advances in Neural Information Processing Systems
\textbf{31}
(2018)
\end{botherref}
\endbibitem

\bibitem[\protect\citeauthoryear{Gouk et~al.}{2021}]{gouk2021regularisation}
\begin{barticle}
\bauthor{\bsnm{Gouk}, \binits{H.}},
\bauthor{\bsnm{Frank}, \binits{E.}},
\bauthor{\bsnm{Pfahringer}, \binits{B.}},
\bauthor{\bsnm{Cree}, \binits{M.J.}}:
\batitle{Regularisation of neural networks by enforcing lipschitz continuity}.
\bjtitle{Machine Learning}
\bvolume{110}(\bissue{2}),
\bfpage{393}--\blpage{416}
(\byear{2021})
\end{barticle}
\endbibitem

\bibitem[\protect\citeauthoryear{Pauli
  et~al.}{2021}]{adversarial_robustness_lipschitz_bound}
\begin{barticle}
\bauthor{\bsnm{Pauli}, \binits{P.}},
\bauthor{\bsnm{Koch}, \binits{A.}},
\bauthor{\bsnm{Berberich}, \binits{J.}},
\bauthor{\bsnm{Kohler}, \binits{P.}},
\bauthor{\bsnm{Allg{\"o}wer}, \binits{F.}}:
\batitle{Training robust neural networks using lipschitz bounds}.
\bjtitle{IEEE Control Systems Letters}
\bvolume{6},
\bfpage{121}--\blpage{126}
(\byear{2021})
\end{barticle}
\endbibitem

\bibitem[\protect\citeauthoryear{Kiwiel}{2006}]{gd_approx_1}
\begin{bbook}
\bauthor{\bsnm{Kiwiel}, \binits{K.C.}}:
\bbtitle{Methods of Descent for Nondifferentiable Optimization}
vol. \bseriesno{1133}.
\bpublisher{Springer},
\blocation{Berlin}
(\byear{2006})
\end{bbook}
\endbibitem

\bibitem[\protect\citeauthoryear{Shor}{2012}]{gd_approx_2}
\begin{bbook}
\bauthor{\bsnm{Shor}, \binits{N.Z.}}:
\bbtitle{Minimization Methods for Non-differentiable Functions}
vol. \bseriesno{3}.
\bpublisher{Springer},
\blocation{Berlin}
(\byear{2012})
\end{bbook}
\endbibitem

\bibitem[\protect\citeauthoryear{Conn et~al.}{2009}]{derivative_free_opt_1}
\begin{bbook}
\bauthor{\bsnm{Conn}, \binits{A.R.}},
\bauthor{\bsnm{Scheinberg}, \binits{K.}},
\bauthor{\bsnm{Vicente}, \binits{L.N.}}:
\bbtitle{Introduction to Derivative-free Optimization}.
\bpublisher{SIAM},
\blocation{Philadelphia}
(\byear{2009})
\end{bbook}
\endbibitem

\bibitem[\protect\citeauthoryear{Rios and
  Sahinidis}{2013}]{derivative_free_opt_2}
\begin{barticle}
\bauthor{\bsnm{Rios}, \binits{L.M.}},
\bauthor{\bsnm{Sahinidis}, \binits{N.V.}}:
\batitle{Derivative-free optimization: a review of algorithms and comparison of
  software implementations}.
\bjtitle{Journal of Global Optimization}
\bvolume{56}(\bissue{3}),
\bfpage{1247}--\blpage{1293}
(\byear{2013})
\end{barticle}
\endbibitem

\bibitem[\protect\citeauthoryear{Liu et~al.}{2020}]{zeroth_optimisation_3}
\begin{barticle}
\bauthor{\bsnm{Liu}, \binits{S.}},
\bauthor{\bsnm{Chen}, \binits{P.-Y.}},
\bauthor{\bsnm{Kailkhura}, \binits{B.}},
\bauthor{\bsnm{Zhang}, \binits{G.}},
\bauthor{\bsnm{Hero~III}, \binits{A.O.}},
\bauthor{\bsnm{Varshney}, \binits{P.K.}}:
\batitle{A primer on zeroth-order optimization in signal processing and machine
  learning: Principals, recent advances, and applications}.
\bjtitle{IEEE Signal Processing Magazine}
\bvolume{37}(\bissue{5}),
\bfpage{43}--\blpage{54}
(\byear{2020})
\end{barticle}
\endbibitem

\bibitem[\protect\citeauthoryear{Chen et~al.}{2017}]{zeroth_optimisation_1}
\begin{bchapter}
\bauthor{\bsnm{Chen}, \binits{P.-Y.}},
\bauthor{\bsnm{Zhang}, \binits{H.}},
\bauthor{\bsnm{Sharma}, \binits{Y.}},
\bauthor{\bsnm{Yi}, \binits{J.}},
\bauthor{\bsnm{Hsieh}, \binits{C.-J.}}:
\bctitle{Zoo: Zeroth order optimization based black-box attacks to deep neural
  networks without training substitute models}.
In: \bbtitle{Proceedings of the 10th ACM Workshop on Artificial Intelligence
  and Security},
pp. \bfpage{15}--\blpage{26}
(\byear{2017})
\end{bchapter}
\endbibitem

\bibitem[\protect\citeauthoryear{Dhurandhar
  et~al.}{2019}]{zeroth_optimisation_2}
\begin{botherref}
\oauthor{\bsnm{Dhurandhar}, \binits{A.}},
\oauthor{\bsnm{Pedapati}, \binits{T.}},
\oauthor{\bsnm{Balakrishnan}, \binits{A.}},
\oauthor{\bsnm{Chen}, \binits{P.-Y.}},
\oauthor{\bsnm{Shanmugam}, \binits{K.}},
\oauthor{\bsnm{Puri}, \binits{R.}}:
Model agnostic contrastive explanations for structured data.
arXiv preprint arXiv:1906.00117
(2019)
\end{botherref}
\endbibitem

\bibitem[\protect\citeauthoryear{Efron and
  Hastie}{2016}]{regularisation-cambridge-book}
\begin{bbook}
\bauthor{\bsnm{Efron}, \binits{B.}},
\bauthor{\bsnm{Hastie}, \binits{T.}}:
\bbtitle{Computer Age Statistical Inference: Algorithms, Evidence, and Data
  Science}.
\bsertitle{Institute of Mathematical Statistics Monographs}.
\bpublisher{Cambridge University Press},
\blocation{Cambridge}
(\byear{2016}).
\doiurl{10.1017/CBO9781316576533}
\end{bbook}
\endbibitem

\bibitem[\protect\citeauthoryear{Kukačka
  et~al.}{}]{regularisation-for-deep-learning}
\begin{botherref}
\oauthor{\bsnm{Kukačka}, \binits{J.}},
\oauthor{\bsnm{Golkov}, \binits{V.}},
\oauthor{\bsnm{Cremers}, \binits{D.}}
\doiurl{10.48550/ARXIV.1710.10686} .
\url{https://arxiv.org/abs/1710.10686}
\end{botherref}
\endbibitem

\bibitem[\protect\citeauthoryear{Bartlett
  et~al.}{2002}]{model-selection-and-error-estimation}
\begin{barticle}
\bauthor{\bsnm{Bartlett}, \binits{P.}},
\bauthor{\bsnm{Boucheron}, \binits{S.}},
\bauthor{\bsnm{Lugosi}, \binits{G.}}:
\batitle{Model selection and error estimation}.
\bjtitle{Machine Learning}
\bvolume{48},
\bfpage{85}--\blpage{113}
(\byear{2002})
\doiurl{10.1023/A:1013999503812}
\end{barticle}
\endbibitem

\bibitem[\protect\citeauthoryear{Myung}{2000}]{importance-of-complexity-in-model-selection}
\begin{barticle}
\bauthor{\bsnm{Myung}, \binits{I.J.}}:
\batitle{The importance of complexity in model selection}.
\bjtitle{Journal of Mathematical Psychology}
\bvolume{44}(\bissue{1}),
\bfpage{190}--\blpage{204}
(\byear{2000})
\doiurl{10.1006/jmps.1999.1283}
\end{barticle}
\endbibitem

\bibitem[\protect\citeauthoryear{Bishop and Nasrabadi}{2006}]{ML_3}
\begin{bbook}
\bauthor{\bsnm{Bishop}, \binits{C.M.}},
\bauthor{\bsnm{Nasrabadi}, \binits{N.M.}}:
\bbtitle{Pattern Recognition and Machine Learning}
vol. \bseriesno{4}.
\bpublisher{Springer},
\blocation{Berlin}
(\byear{2006})
\end{bbook}
\endbibitem

\bibitem[\protect\citeauthoryear{Hoerl and Kennard}{2000}]{ridge_regression}
\begin{barticle}
\bauthor{\bsnm{Hoerl}, \binits{A.E.}},
\bauthor{\bsnm{Kennard}, \binits{R.W.}}:
\batitle{Ridge regression: Biased estimation for nonorthogonal problems}.
\bjtitle{Technometrics}
\bvolume{42}(\bissue{1}),
\bfpage{80}--\blpage{86}
(\byear{2000})
\end{barticle}
\endbibitem

\bibitem[\protect\citeauthoryear{Tibshirani}{1996}]{lasso_regression}
\begin{barticle}
\bauthor{\bsnm{Tibshirani}, \binits{R.}}:
\batitle{Regression shrinkage and selection via the lasso}.
\bjtitle{Journal of the Royal Statistical Society: Series B (Methodological)}
\bvolume{58}(\bissue{1}),
\bfpage{267}--\blpage{288}
(\byear{1996})
\end{barticle}
\endbibitem

\bibitem[\protect\citeauthoryear{Ng}{2004}]{andrew-ng-l1-l2}
\begin{bchapter}
\bauthor{\bsnm{Ng}, \binits{A.Y.}}:
\bctitle{Feature selection, l1 vs. l2 regularization, and rotational
  invariance}.
In: \bbtitle{Proceedings of the Twenty-First International Conference on
  Machine Learning}.
\bsertitle{ICML '04},
p. \bfpage{78}.
\bpublisher{Association for Computing Machinery},
\blocation{New York, NY, USA}
(\byear{2004}).
\doiurl{10.1145/1015330.1015435} .
\burl{https://doi.org/10.1145/1015330.1015435}
\end{bchapter}
\endbibitem

\bibitem[\protect\citeauthoryear{Bekta{\c{s}} and
  {\c{S}}i{\c{s}}man}{2010}]{l1_vs_l2}
\begin{barticle}
\bauthor{\bsnm{Bekta{\c{s}}}, \binits{S.}},
\bauthor{\bsnm{{\c{S}}i{\c{s}}man}, \binits{Y.}}:
\batitle{The comparison of l1 and l2-norm minimization methods}.
\bjtitle{International Journal of the Physical Sciences}
\bvolume{5}(\bissue{11}),
\bfpage{1721}--\blpage{1727}
(\byear{2010})
\end{barticle}
\endbibitem

\bibitem[\protect\citeauthoryear{Tsuruoka et~al.}{2009}]{l1_sgd}
\begin{bchapter}
\bauthor{\bsnm{Tsuruoka}, \binits{Y.}},
\bauthor{\bsnm{Tsujii}, \binits{J.}},
\bauthor{\bsnm{Ananiadou}, \binits{S.}}:
\bctitle{Stochastic gradient descent training for l1-regularized log-linear
  models with cumulative penalty.},
pp. \bfpage{477}--\blpage{485}
(\byear{2009}).
\doiurl{10.3115/1687878.1687946}
\end{bchapter}
\endbibitem

\bibitem[\protect\citeauthoryear{Ullah et~al.}{2019}]{ullah2019short}
\begin{barticle}
\bauthor{\bsnm{Ullah}, \binits{F.U.M.}},
\bauthor{\bsnm{Ullah}, \binits{A.}},
\bauthor{\bsnm{Haq}, \binits{I.U.}},
\bauthor{\bsnm{Rho}, \binits{S.}},
\bauthor{\bsnm{Baik}, \binits{S.W.}}:
\batitle{Short-term prediction of residential power energy consumption via cnn
  and multi-layer bi-directional lstm networks}.
\bjtitle{IEEE Access}
\bvolume{8},
\bfpage{123369}--\blpage{123380}
(\byear{2019})
\end{barticle}
\endbibitem

\bibitem[\protect\citeauthoryear{Krishnaiah
  et~al.}{2007}]{krishnaiah2007neural}
\begin{barticle}
\bauthor{\bsnm{Krishnaiah}, \binits{T.}},
\bauthor{\bsnm{Rao}, \binits{S.S.}},
\bauthor{\bsnm{Madhumurthy}, \binits{K.}},
\bauthor{\bsnm{Reddy}, \binits{K.}}:
\batitle{Neural network approach for modelling global solar radiation}.
\bjtitle{Journal of Applied Sciences Research}
\bvolume{3}(\bissue{10}),
\bfpage{1105}--\blpage{1111}
(\byear{2007})
\end{barticle}
\endbibitem

\bibitem[\protect\citeauthoryear{Valipour
  et~al.}{2013}]{valipour2013comparison}
\begin{barticle}
\bauthor{\bsnm{Valipour}, \binits{M.}},
\bauthor{\bsnm{Banihabib}, \binits{M.E.}},
\bauthor{\bsnm{Behbahani}, \binits{S.M.R.}}:
\batitle{Comparison of the arma, arima, and the autoregressive artificial
  neural network models in forecasting the monthly inflow of dez dam
  reservoir}.
\bjtitle{Journal of hydrology}
\bvolume{476},
\bfpage{433}--\blpage{441}
(\byear{2013})
\end{barticle}
\endbibitem

\bibitem[\protect\citeauthoryear{Willmott and
  Matsuura}{2005}]{willmott2005advantages}
\begin{barticle}
\bauthor{\bsnm{Willmott}, \binits{C.J.}},
\bauthor{\bsnm{Matsuura}, \binits{K.}}:
\batitle{Advantages of the mean absolute error (mae) over the root mean square
  error (rmse) in assessing average model performance}.
\bjtitle{Climate research}
\bvolume{30}(\bissue{1}),
\bfpage{79}--\blpage{82}
(\byear{2005})
\end{barticle}
\endbibitem

\bibitem[\protect\citeauthoryear{Li and Shi}{2010}]{li2010comparing}
\begin{barticle}
\bauthor{\bsnm{Li}, \binits{G.}},
\bauthor{\bsnm{Shi}, \binits{J.}}:
\batitle{On comparing three artificial neural networks for wind speed
  forecasting}.
\bjtitle{Applied Energy}
\bvolume{87}(\bissue{7}),
\bfpage{2313}--\blpage{2320}
(\byear{2010})
\end{barticle}
\endbibitem

\bibitem[\protect\citeauthoryear{Murphy}{2013}]{murphy2013machine}
\begin{bbook}
\bauthor{\bsnm{Murphy}, \binits{K.P.}}:
\bbtitle{Machine Learning : a Probabilistic Perspective}.
\bpublisher{MIT Press},
\blocation{Cambridge, Mass. [u.a.]}
(\byear{2013}).
\burl{https://www.amazon.com/Machine-Learning-Probabilistic-Perspective-Computation/dp/0262018020/ref=sr_1_2?ie=UTF8&qid=1336857747&sr=8-2}
\end{bbook}
\endbibitem

\bibitem[\protect\citeauthoryear{Huber}{1965}]{huber1965robust}
\begin{botherref}
\oauthor{\bsnm{Huber}, \binits{P.J.}}:
A robust version of the probability ratio test.
The Annals of Mathematical Statistics,
1753--1758
(1965)
\end{botherref}
\endbibitem

\bibitem[\protect\citeauthoryear{Girshick}{2015}]{girshick2015fast}
\begin{bchapter}
\bauthor{\bsnm{Girshick}, \binits{R.}}:
\bctitle{Fast r-cnn}.
In: \bbtitle{Proceedings of the IEEE International Conference on Computer
  Vision},
pp. \bfpage{1440}--\blpage{1448}
(\byear{2015})
\end{bchapter}
\endbibitem

\bibitem[\protect\citeauthoryear{Ren et~al.}{2015}]{ren2015faster}
\begin{bchapter}
\bauthor{\bsnm{Ren}, \binits{S.}},
\bauthor{\bsnm{He}, \binits{K.}},
\bauthor{\bsnm{Girshick}, \binits{R.}},
\bauthor{\bsnm{Sun}, \binits{J.}}:
\bctitle{Faster r-cnn: Towards real-time object detection with region proposal
  networks}.
In: \bbtitle{Advances in Neural Information Processing Systems},
pp. \bfpage{91}--\blpage{99}
(\byear{2015})
\end{bchapter}
\endbibitem

\bibitem[\protect\citeauthoryear{Chicco et~al.}{2021}]{chicco2021advantages}
\begin{barticle}
\bauthor{\bsnm{Chicco}, \binits{D.}},
\bauthor{\bsnm{Warrens}, \binits{M.J.}},
\bauthor{\bsnm{Jurman}, \binits{G.}}:
\batitle{The advantages of the matthews correlation coefficient (mcc) over f1
  score and accuracy in binary classification evaluation}.
\bjtitle{BMC Genomics}
\bvolume{21}(\bissue{1}),
\bfpage{6}--\blpage{16}
(\byear{2021})
\end{barticle}
\endbibitem

\bibitem[\protect\citeauthoryear{Semeniuta}{2017a}]{regression_rmsle}
\begin{bchapter}
\bauthor{\bsnm{Semeniuta}, \binits{A.}}:
\bctitle{A handy approximation of the rmsle loss function}.
In: \bbtitle{European Conference on Machine Learning},
pp. \bfpage{114}--\blpage{125}
(\byear{2017})
\end{bchapter}
\endbibitem

\bibitem[\protect\citeauthoryear{Semeniuta}{2017b}]{semeniuta2017handy}
\begin{botherref}
\oauthor{\bsnm{Semeniuta}, \binits{A.}}:
Handy approximation of the rmsle loss.
arXiv preprint arXiv:1711.04077
(2017)
\end{botherref}
\endbibitem

\bibitem[\protect\citeauthoryear{Wang}{2006}]{wang2006log}
\begin{botherref}
\oauthor{\bsnm{Wang}, \binits{H.}}:
Logarithmic loss functions.
Journal of Machine Learning Research,
45--62
(2006)
\end{botherref}
\endbibitem

\bibitem[\protect\citeauthoryear{Breiman}{1997}]{breiman1997arcing}
\begin{barticle}
\bauthor{\bsnm{Breiman}, \binits{L.}}:
\batitle{Arcing classifiers}.
\bjtitle{Annals of Statistics}
\bvolume{25}(\bissue{2}),
\bfpage{17}--\blpage{26}
(\byear{1997})
\end{barticle}
\endbibitem

\bibitem[\protect\citeauthoryear{Lin}{2020}]{lin2020frequency}
\begin{botherref}
\oauthor{\bsnm{Lin}, \binits{H.}}:
Frequency and magnitude adjustment for rmsle in predictive modeling.
IEEE Transactions on Knowledge and Data Engineering,
118--128
(2020)
\end{botherref}
\endbibitem

\bibitem[\protect\citeauthoryear{Bartlett
  et~al.}{2006}]{Bartlett2006ConvexityCA}
\begin{barticle}
\bauthor{\bsnm{Bartlett}, \binits{P.L.}},
\bauthor{\bsnm{Jordan}, \binits{M.I.}},
\bauthor{\bsnm{McAuliffe}, \binits{J.D.}}:
\batitle{Convexity, classification, and risk bounds}.
\bjtitle{Journal of the American Statistical Association}
\bvolume{101},
\bfpage{138}--\blpage{156}
(\byear{2006})
\end{barticle}
\endbibitem

\bibitem[\protect\citeauthoryear{Jiang}{2004}]{f4a7c73671ce457daaae7ceb2c0e1976}
\begin{barticle}
\bauthor{\bsnm{Jiang}, \binits{W.}}:
\batitle{Process consistency for adaboost}.
\bjtitle{Annals of Statistics}
\bvolume{32}(\bissue{1}),
\bfpage{13}--\blpage{29}
(\byear{2004})
\doiurl{10.1214/aos/1079120128}
\end{barticle}
\endbibitem

\bibitem[\protect\citeauthoryear{Lugosi and Vayatis}{2003}]{Lugosi2003OnTB}
\begin{barticle}
\bauthor{\bsnm{Lugosi}, \binits{G.}},
\bauthor{\bsnm{Vayatis}, \binits{N.}}:
\batitle{On the bayes-risk consistency of regularized boosting methods}.
\bjtitle{Annals of Statistics}
\bvolume{32},
\bfpage{30}--\blpage{55}
(\byear{2003})
\end{barticle}
\endbibitem

\bibitem[\protect\citeauthoryear{Mannor et~al.}{2003}]{Mannor2003GreedyAF}
\begin{barticle}
\bauthor{\bsnm{Mannor}, \binits{S.}},
\bauthor{\bsnm{Meir}, \binits{R.}},
\bauthor{\bsnm{Zhang}, \binits{T.}}:
\batitle{Greedy algorithms for classification -- consistency, convergence
  rates, and adaptivity}.
\bjtitle{J. Mach. Learn. Res.}
\bvolume{4},
\bfpage{713}--\blpage{741}
(\byear{2003})
\end{barticle}
\endbibitem

\bibitem[\protect\citeauthoryear{Steinwart}{2005}]{1377497}
\begin{barticle}
\bauthor{\bsnm{Steinwart}, \binits{I.}}:
\batitle{Consistency of support vector machines and other regularized kernel
  classifiers}.
\bjtitle{IEEE Transactions on Information Theory}
\bvolume{51}(\bissue{1}),
\bfpage{128}--\blpage{142}
(\byear{2005})
\doiurl{10.1109/TIT.2004.839514}
\end{barticle}
\endbibitem

\bibitem[\protect\citeauthoryear{Zhang}{2001}]{Zhang01statisticalbehavior}
\begin{botherref}
\oauthor{\bsnm{Zhang}, \binits{T.}}:
Statistical Behavior and Consistency of Classification Methods based on Convex
  Risk Minimization
(2001)
\end{botherref}
\endbibitem

\bibitem[\protect\citeauthoryear{Gentile and Warmuth}{1998}]{NIPS1998_a14ac55a}
\begin{bchapter}
\bauthor{\bsnm{Gentile}, \binits{C.}},
\bauthor{\bsnm{Warmuth}, \binits{M.K.K.}}:
\bctitle{Linear hinge loss and average margin}.
In: \beditor{\bsnm{Kearns}, \binits{M.}},
\beditor{\bsnm{Solla}, \binits{S.}},
\beditor{\bsnm{Cohn}, \binits{D.}} (eds.)
\bbtitle{Advances in Neural Information Processing Systems},
vol. \bseriesno{11}.
\bpublisher{MIT Press},
\blocation{Massachusetts}
(\byear{1998}).
\burl{https://proceedings.neurips.cc/paper/1998/file/a14ac55a4f27472c5d894ec1c3c743d2-Paper.pdf}
\end{bchapter}
\endbibitem

\bibitem[\protect\citeauthoryear{Boser et~al.}{}]{bosertraining}
\begin{botherref}
\oauthor{\bsnm{Boser}, \binits{B.E.}},
\oauthor{\bsnm{Guyon}, \binits{I.M.}},
\oauthor{\bsnm{Vapnik}, \binits{V.N.}}:
A training algorithm for optimal margin classifiers.
In: Proceedings of the 5th Annual ACM Workshop on Computational Learning
  Theory,
pp. 144--152
\end{botherref}
\endbibitem

\bibitem[\protect\citeauthoryear{Mathur and Foody}{2008}]{mathur2008multiclass}
\begin{barticle}
\bauthor{\bsnm{Mathur}, \binits{A.}},
\bauthor{\bsnm{Foody}, \binits{G.M.}}:
\batitle{Multiclass and binary svm classification: Implications for training
  and classification users}.
\bjtitle{IEEE Geoscience and remote sensing letters}
\bvolume{5}(\bissue{2}),
\bfpage{241}--\blpage{245}
(\byear{2008})
\end{barticle}
\endbibitem

\bibitem[\protect\citeauthoryear{Rosenblatt}{1958}]{rosenblatt1958perceptron}
\begin{barticle}
\bauthor{\bsnm{Rosenblatt}, \binits{F.}}:
\batitle{The perceptron: a probabilistic model for information storage and
  organization in the brain.}
\bjtitle{Psychological review}
\bvolume{65}(\bissue{6}),
\bfpage{386}
(\byear{1958})
\end{barticle}
\endbibitem

\bibitem[\protect\citeauthoryear{Rennie}{2013}]{Rennie2013SmoothHC}
\begin{bchapter}
\bauthor{\bsnm{Rennie}, \binits{J.D.M.}}:
\bctitle{Smooth hinge classication}.
(\byear{2013})
\end{bchapter}
\endbibitem

\bibitem[\protect\citeauthoryear{Zhang}{2004}]{Zhang04solvinglarge}
\begin{bchapter}
\bauthor{\bsnm{Zhang}, \binits{T.}}:
\bctitle{Solving large scale linear prediction problems using stochastic
  gradient descent algorithms}.
In: \bbtitle{ICML 2004: PROCEEDINGS OF THE TWENTY-FIRST INTERNATIONAL
  CONFERENCE ON MACHINE LEARNING. OMNIPRESS},
pp. \bfpage{919}--\blpage{926}
(\byear{2004})
\end{bchapter}
\endbibitem

\bibitem[\protect\citeauthoryear{Wu and Liu}{2007}]{Wu2007RobustTH}
\begin{barticle}
\bauthor{\bsnm{Wu}, \binits{Y.}},
\bauthor{\bsnm{Liu}, \binits{Y.}}:
\batitle{Robust truncated hinge loss support vector machines}.
\bjtitle{Journal of the American Statistical Association}
\bvolume{102},
\bfpage{974}--\blpage{983}
(\byear{2007})
\end{barticle}
\endbibitem

\bibitem[\protect\citeauthoryear{Lee et~al.}{2002}]{Lee02multicategorysupport}
\begin{botherref}
\oauthor{\bsnm{Lee}, \binits{Y.}},
\oauthor{\bsnm{Lee}, \binits{Y.}},
\oauthor{\bsnm{Lee}, \binits{Y.}},
\oauthor{\bsnm{Lin}, \binits{Y.}},
\oauthor{\bsnm{Lin}, \binits{Y.}},
\oauthor{\bsnm{Wahba}, \binits{G.}},
\oauthor{\bsnm{Wahba}, \binits{G.}}:
Multicategory Support Vector Machines, Theory, and Application to the
  Classification of Microarray Data and Satellite Radiance Data
(2002)
\end{botherref}
\endbibitem

\bibitem[\protect\citeauthoryear{Harshvardhan et~al.}{2020}]{generative_models}
\begin{barticle}
\bauthor{\bsnm{Harshvardhan}, \binits{G.}},
\bauthor{\bsnm{Gourisaria}, \binits{M.K.}},
\bauthor{\bsnm{Pandey}, \binits{M.}},
\bauthor{\bsnm{Rautaray}, \binits{S.S.}}:
\batitle{A comprehensive survey and analysis of generative models in machine
  learning}.
\bjtitle{Computer Science Review}
\bvolume{38},
\bfpage{100285}
(\byear{2020})
\end{barticle}
\endbibitem

\bibitem[\protect\citeauthoryear{Myung}{2003}]{MLE_tutorial}
\begin{barticle}
\bauthor{\bsnm{Myung}, \binits{I.J.}}:
\batitle{Tutorial on maximum likelihood estimation}.
\bjtitle{Journal of mathematical Psychology}
\bvolume{47}(\bissue{1}),
\bfpage{90}--\blpage{100}
(\byear{2003})
\end{barticle}
\endbibitem

\bibitem[\protect\citeauthoryear{Joyce}{2011}]{kl_div}
\begin{bchapter}
\bauthor{\bsnm{Joyce}, \binits{J.M.}}:
\bctitle{Kullback-leibler divergence}.
In: \bbtitle{International Encyclopedia of Statistical Science},
pp. \bfpage{720}--\blpage{722}.
\bpublisher{Springer},
\blocation{Berlin}
(\byear{2011})
\end{bchapter}
\endbibitem

\bibitem[\protect\citeauthoryear{Lin}{2017}]{lin2017focal}
\begin{botherref}
\oauthor{\bsnm{Lin}, \binits{T.}}:
Focal loss for dense object detection.
arXiv preprint arXiv:1708.02002
(2017)
\end{botherref}
\endbibitem

\bibitem[\protect\citeauthoryear{Sudre et~al.}{2017}]{sudre2017generalised}
\begin{bchapter}
\bauthor{\bsnm{Sudre}, \binits{C.H.}},
\bauthor{\bsnm{Li}, \binits{W.}},
\bauthor{\bsnm{Vercauteren}, \binits{T.}},
\bauthor{\bsnm{Ourselin}, \binits{S.}},
\bauthor{\bsnm{Jorge~Cardoso}, \binits{M.}}:
\bctitle{Generalised dice overlap as a deep learning loss function for highly
  unbalanced segmentations}.
In: \bbtitle{Deep Learning in Medical Image Analysis and Multimodal Learning
  for Clinical Decision Support: Third International Workshop, DLMIA 2017, and
  7th International Workshop, ML-CDS 2017, Held in Conjunction with MICCAI
  2017, Qu{\'e}bec City, QC, Canada, September 14, Proceedings 3},
pp. \bfpage{240}--\blpage{248}
(\byear{2017}).
\bcomment{Springer}
\end{bchapter}
\endbibitem

\bibitem[\protect\citeauthoryear{Salehi et~al.}{2017}]{salehi2017tversky}
\begin{bchapter}
\bauthor{\bsnm{Salehi}, \binits{S.S.M.}},
\bauthor{\bsnm{Erdogmus}, \binits{D.}},
\bauthor{\bsnm{Gholipour}, \binits{A.}}:
\bctitle{Tversky loss function for image segmentation using 3d fully
  convolutional deep networks}.
In: \bbtitle{International Workshop on Machine Learning in Medical Imaging},
pp. \bfpage{379}--\blpage{387}
(\byear{2017}).
\bcomment{Springer}
\end{bchapter}
\endbibitem

\bibitem[\protect\citeauthoryear{Goodfellow
  et~al.}{2020}]{goodfellow2020generative}
\begin{barticle}
\bauthor{\bsnm{Goodfellow}, \binits{I.}},
\bauthor{\bsnm{Pouget-Abadie}, \binits{J.}},
\bauthor{\bsnm{Mirza}, \binits{M.}},
\bauthor{\bsnm{Xu}, \binits{B.}},
\bauthor{\bsnm{Warde-Farley}, \binits{D.}},
\bauthor{\bsnm{Ozair}, \binits{S.}},
\bauthor{\bsnm{Courville}, \binits{A.}},
\bauthor{\bsnm{Bengio}, \binits{Y.}}:
\batitle{Generative adversarial networks}.
\bjtitle{Communications of the ACM}
\bvolume{63}(\bissue{11}),
\bfpage{139}--\blpage{144}
(\byear{2020})
\end{barticle}
\endbibitem

\bibitem[\protect\citeauthoryear{Goodfellow
  et~al.}{2014}]{goodfellow2014generative}
\begin{botherref}
\oauthor{\bsnm{Goodfellow}, \binits{I.J.}},
\oauthor{\bsnm{Pouget-Abadie}, \binits{J.}},
\oauthor{\bsnm{Mirza}, \binits{M.}},
\oauthor{\bsnm{Xu}, \binits{B.}},
\oauthor{\bsnm{Warde-Farley}, \binits{D.}},
\oauthor{\bsnm{Ozair}, \binits{S.}},
\oauthor{\bsnm{Courville}, \binits{A.}},
\oauthor{\bsnm{Bengio}, \binits{Y.}}:
Generative adversarial networks.
arXiv preprint arXiv:1406.2661
(2014)
\end{botherref}
\endbibitem

\bibitem[\protect\citeauthoryear{Van~Oord et~al.}{2016}]{van2016pixel}
\begin{bchapter}
\bauthor{\bsnm{Van~Oord}, \binits{A.}},
\bauthor{\bsnm{Kalchbrenner}, \binits{N.}},
\bauthor{\bsnm{Kavukcuoglu}, \binits{K.}}:
\bctitle{Pixel recurrent neural networks}.
In: \bbtitle{International Conference on Machine Learning},
pp. \bfpage{1747}--\blpage{1756}
(\byear{2016}).
\bcomment{PMLR}
\end{bchapter}
\endbibitem

\bibitem[\protect\citeauthoryear{Dinh et~al.}{2016}]{dinh2016density}
\begin{botherref}
\oauthor{\bsnm{Dinh}, \binits{L.}},
\oauthor{\bsnm{Sohl-Dickstein}, \binits{J.}},
\oauthor{\bsnm{Bengio}, \binits{S.}}:
Density estimation using real nvp.
arXiv preprint arXiv:1605.08803
(2016)
\end{botherref}
\endbibitem

\bibitem[\protect\citeauthoryear{Rezende and
  Mohamed}{2015}]{rezende2015variational}
\begin{bchapter}
\bauthor{\bsnm{Rezende}, \binits{D.J.}},
\bauthor{\bsnm{Mohamed}, \binits{S.}}:
\bctitle{Variational inference with normalizing flows}.
In: \bbtitle{International Conference on Machine Learning (ICML)},
pp. \bfpage{1530}--\blpage{1538}
(\byear{2015})
\end{bchapter}
\endbibitem

\bibitem[\protect\citeauthoryear{van~den Oord et~al.}{2016}]{oord2016wavenet}
\begin{botherref}
\oauthor{\bsnm{Oord}, \binits{A.}},
\oauthor{\bsnm{Dieleman}, \binits{S.}},
\oauthor{\bsnm{Zen}, \binits{H.}},
\oauthor{\bsnm{Simonyan}, \binits{K.}},
\oauthor{\bsnm{Vinyals}, \binits{O.}},
\oauthor{\bsnm{Graves}, \binits{A.}},
\oauthor{\bsnm{Kalchbrenner}, \binits{N.}},
\oauthor{\bsnm{Senior}, \binits{A.}},
\oauthor{\bsnm{Kavukcuoglu}, \binits{K.}}:
Wavenet: A generative model for raw audio.
arXiv preprint arXiv:1609.03499
(2016)
\end{botherref}
\endbibitem

\bibitem[\protect\citeauthoryear{Kingma and Welling}{2013}]{kingma2013auto}
\begin{botherref}
\oauthor{\bsnm{Kingma}, \binits{D.P.}},
\oauthor{\bsnm{Welling}, \binits{M.}}:
Auto-encoding variational bayes.
arXiv preprint arXiv:1312.6114
(2013)
\end{botherref}
\endbibitem

\bibitem[\protect\citeauthoryear{Rezende et~al.}{2014}]{rezende2014stochastic}
\begin{bchapter}
\bauthor{\bsnm{Rezende}, \binits{D.J.}},
\bauthor{\bsnm{Mohamed}, \binits{S.}},
\bauthor{\bsnm{Wierstra}, \binits{D.}}:
\bctitle{Stochastic backpropagation and approximate inference in deep
  generative models}.
In: \bbtitle{International Conference on Machine Learning},
pp. \bfpage{1278}--\blpage{1286}
(\byear{2014}).
\bcomment{PMLR}
\end{bchapter}
\endbibitem

\bibitem[\protect\citeauthoryear{Hou et~al.}{2016}]{hou2017deep}
\begin{botherref}
\oauthor{\bsnm{Hou}, \binits{X.}},
\oauthor{\bsnm{Shen}, \binits{L.}},
\oauthor{\bsnm{Sun}, \binits{K.}},
\oauthor{\bsnm{Qiu}, \binits{G.}}:
Deep feature consistent variational autoencoder.
CoRR
\textbf{abs/1610.00291}
(2016)
{\href{https://arxiv.org/abs/1610.00291}{{1610.00291}}}
\end{botherref}
\endbibitem

\bibitem[\protect\citeauthoryear{Bengio
  et~al.}{2012}]{bengio2013representation}
\begin{botherref}
\oauthor{\bsnm{Bengio}, \binits{Y.}},
\oauthor{\bsnm{Courville}, \binits{A.C.}},
\oauthor{\bsnm{Vincent}, \binits{P.}}:
Unsupervised feature learning and deep learning: {A} review and new
  perspectives.
CoRR
\textbf{abs/1206.5538}
(2012)
{\href{https://arxiv.org/abs/1206.5538}{{1206.5538}}}
\end{botherref}
\endbibitem

\bibitem[\protect\citeauthoryear{Kingma et~al.}{2014}]{kingma2014semi}
\begin{botherref}
\oauthor{\bsnm{Kingma}, \binits{D.P.}},
\oauthor{\bsnm{Rezende}, \binits{D.J.}},
\oauthor{\bsnm{Mohamed}, \binits{S.}},
\oauthor{\bsnm{Welling}, \binits{M.}}:
Semi-supervised learning with deep generative models.
CoRR
\textbf{abs/1406.5298}
(2014)
{\href{https://arxiv.org/abs/1406.5298}{{1406.5298}}}
\end{botherref}
\endbibitem

\bibitem[\protect\citeauthoryear{An and Cho}{2015}]{an2015variational}
\begin{barticle}
\bauthor{\bsnm{An}, \binits{J.}},
\bauthor{\bsnm{Cho}, \binits{S.}}:
\batitle{Variational autoencoder based anomaly detection using reconstruction
  probability}.
\bjtitle{Special Lecture on IE}
\bvolume{2}(\bissue{1}),
\bfpage{1}--\blpage{18}
(\byear{2015})
\end{barticle}
\endbibitem

\bibitem[\protect\citeauthoryear{Van Den~Oord et~al.}{2017}]{van2017neural}
\begin{botherref}
\oauthor{\bsnm{Van Den~Oord}, \binits{A.}},
\oauthor{\bsnm{Vinyals}, \binits{O.}}, et al.:
Neural discrete representation learning.
Advances in neural information processing systems
\textbf{30}
(2017)
\end{botherref}
\endbibitem

\bibitem[\protect\citeauthoryear{Higgins et~al.}{2017}]{higgins2017beta}
\begin{botherref}
\oauthor{\bsnm{Higgins}, \binits{I.}},
\oauthor{\bsnm{Matthey}, \binits{L.}},
\oauthor{\bsnm{Pal}, \binits{A.}},
\oauthor{\bsnm{Burgess}, \binits{C.P.}},
\oauthor{\bsnm{Glorot}, \binits{X.}},
\oauthor{\bsnm{Botvinick}, \binits{M.M.}},
\oauthor{\bsnm{Mohamed}, \binits{S.}},
\oauthor{\bsnm{Lerchner}, \binits{A.}}:
beta-vae: Learning basic visual concepts with a constrained variational
  framework.
ICLR (Poster)
\textbf{3}
(2017)
\end{botherref}
\endbibitem

\bibitem[\protect\citeauthoryear{Bowman et~al.}{2015}]{bowman2015generating}
\begin{botherref}
\oauthor{\bsnm{Bowman}, \binits{S.R.}},
\oauthor{\bsnm{Vilnis}, \binits{L.}},
\oauthor{\bsnm{Vinyals}, \binits{O.}},
\oauthor{\bsnm{Dai}, \binits{A.M.}},
\oauthor{\bsnm{Jozefowicz}, \binits{R.}},
\oauthor{\bsnm{Bengio}, \binits{S.}}:
Generating sentences from a continuous space.
arXiv preprint arXiv:1511.06349
(2015)
\end{botherref}
\endbibitem

\bibitem[\protect\citeauthoryear{Alemi et~al.}{2018}]{alemi2018fixing}
\begin{bchapter}
\bauthor{\bsnm{Alemi}, \binits{A.}},
\bauthor{\bsnm{Poole}, \binits{B.}},
\bauthor{\bsnm{Fischer}, \binits{I.}},
\bauthor{\bsnm{Dillon}, \binits{J.}},
\bauthor{\bsnm{Saurous}, \binits{R.A.}},
\bauthor{\bsnm{Murphy}, \binits{K.}}:
\bctitle{Fixing a broken elbo}.
In: \bbtitle{International Conference on Machine Learning},
pp. \bfpage{159}--\blpage{168}
(\byear{2018}).
\bcomment{PMLR}
\end{bchapter}
\endbibitem

\bibitem[\protect\citeauthoryear{Burgess
  et~al.}{2018}]{burgess2018understanding}
\begin{botherref}
\oauthor{\bsnm{Burgess}, \binits{C.P.}},
\oauthor{\bsnm{Higgins}, \binits{I.}},
\oauthor{\bsnm{Pal}, \binits{A.}},
\oauthor{\bsnm{Matthey}, \binits{L.}},
\oauthor{\bsnm{Watters}, \binits{N.}},
\oauthor{\bsnm{Desjardins}, \binits{G.}},
\oauthor{\bsnm{Lerchner}, \binits{A.}}:
Understanding disentangling in beta-vae.
arXiv preprint arXiv:1804.03599
(2018)
\end{botherref}
\endbibitem

\bibitem[\protect\citeauthoryear{Dhariwal et~al.}{2020}]{dhariwal2020jukebox}
\begin{botherref}
\oauthor{\bsnm{Dhariwal}, \binits{P.}},
\oauthor{\bsnm{Payne}, \binits{H.}},
\oauthor{\bsnm{Kim}, \binits{J.W.}},
\oauthor{\bsnm{Radford}, \binits{A.}},
\oauthor{\bsnm{Sutskever}, \binits{I.}}:
Jukebox: A generative model for music.
arXiv preprint arXiv:2005.00341
(2020)
\end{botherref}
\endbibitem

\bibitem[\protect\citeauthoryear{Razavi et~al.}{2019}]{razavi2019generating}
\begin{bchapter}
\bauthor{\bsnm{Razavi}, \binits{A.}},
\bauthor{\bsnm{Oord}, \binits{A.}},
\bauthor{\bsnm{Vinyals}, \binits{O.}}:
\bctitle{Generating diverse high-fidelity images with {VQ-VAE-2}}.
In: \bbtitle{Advances in Neural Information Processing Systems (NeurIPS)},
pp. \bfpage{14866}--\blpage{14876}
(\byear{2019})
\end{bchapter}
\endbibitem

\bibitem[\protect\citeauthoryear{Sohn et~al.}{2015}]{sohn2015learning}
\begin{botherref}
\oauthor{\bsnm{Sohn}, \binits{K.}},
\oauthor{\bsnm{Lee}, \binits{H.}},
\oauthor{\bsnm{Yan}, \binits{X.}}:
Learning structured output representation using deep conditional generative
  models.
Advances in neural information processing systems
\textbf{28}
(2015)
\end{botherref}
\endbibitem

\bibitem[\protect\citeauthoryear{Yan et~al.}{2016}]{yan2016attribute2image}
\begin{bchapter}
\bauthor{\bsnm{Yan}, \binits{X.}},
\bauthor{\bsnm{Yang}, \binits{J.}},
\bauthor{\bsnm{Sohn}, \binits{K.}},
\bauthor{\bsnm{Lee}, \binits{H.}},
\bauthor{\bsnm{Yang}, \binits{M.-H.}}:
\bctitle{Attribute2image: Conditional image generation from visual attributes}.
In: \bbtitle{European Conference on Computer Vision (ECCV)},
pp. \bfpage{776}--\blpage{791}
(\byear{2016}).
\bcomment{Springer}
\end{bchapter}
\endbibitem

\bibitem[\protect\citeauthoryear{Goodfellow et~al.}{2014}]{minimax}
\begin{botherref}
\oauthor{\bsnm{Goodfellow}, \binits{I.J.}},
\oauthor{\bsnm{Pouget-Abadie}, \binits{J.}},
\oauthor{\bsnm{Mirza}, \binits{M.}},
\oauthor{\bsnm{Xu}, \binits{B.}},
\oauthor{\bsnm{Warde-Farley}, \binits{D.}},
\oauthor{\bsnm{Ozair}, \binits{S.}},
\oauthor{\bsnm{Courville}, \binits{A.}},
\oauthor{\bsnm{Bengio}, \binits{Y.}}:
Generative Adversarial Networks.
arXiv
(2014).
\doiurl{10.48550/ARXIV.1406.2661} .
\url{https://arxiv.org/abs/1406.2661}
\end{botherref}
\endbibitem

\bibitem[\protect\citeauthoryear{Arjovsky et~al.}{2017}]{Wasserstein}
\begin{botherref}
\oauthor{\bsnm{Arjovsky}, \binits{M.}},
\oauthor{\bsnm{Chintala}, \binits{S.}},
\oauthor{\bsnm{Bottou}, \binits{L.}}:
Wasserstein GAN.
arXiv
(2017).
\doiurl{10.48550/ARXIV.1701.07875} .
\url{https://arxiv.org/abs/1701.07875}
\end{botherref}
\endbibitem

\bibitem[\protect\citeauthoryear{Stanczuk
  et~al.}{2021}]{stanczuk2021wasserstein}
\begin{botherref}
\oauthor{\bsnm{Stanczuk}, \binits{J.}},
\oauthor{\bsnm{Etmann}, \binits{C.}},
\oauthor{\bsnm{Kreusser}, \binits{L.M.}},
\oauthor{\bsnm{Sch{\"o}nlieb}, \binits{C.-B.}}:
Wasserstein gans work because they fail (to approximate the wasserstein
  distance).
arXiv preprint arXiv:2103.01678
(2021)
\end{botherref}
\endbibitem

\bibitem[\protect\citeauthoryear{Sohl-Dickstein et~al.}{2015}]{sohl2015deep}
\begin{bchapter}
\bauthor{\bsnm{Sohl-Dickstein}, \binits{J.}},
\bauthor{\bsnm{Weiss}, \binits{E.}},
\bauthor{\bsnm{Maheswaranathan}, \binits{N.}},
\bauthor{\bsnm{Ganguli}, \binits{S.}}:
\bctitle{Deep unsupervised learning using nonequilibrium thermodynamics}.
In: \bbtitle{International Conference on Machine Learning},
pp. \bfpage{2256}--\blpage{2265}
(\byear{2015}).
\bcomment{PMLR}
\end{bchapter}
\endbibitem

\bibitem[\protect\citeauthoryear{Song and Ermon}{2019}]{song2019generative}
\begin{botherref}
\oauthor{\bsnm{Song}, \binits{Y.}},
\oauthor{\bsnm{Ermon}, \binits{S.}}:
Generative modeling by estimating gradients of the data distribution.
Advances in Neural Information Processing Systems
\textbf{32}
(2019)
\end{botherref}
\endbibitem

\bibitem[\protect\citeauthoryear{Ho et~al.}{2020}]{ho2020denoising}
\begin{barticle}
\bauthor{\bsnm{Ho}, \binits{J.}},
\bauthor{\bsnm{Jain}, \binits{A.}},
\bauthor{\bsnm{Abbeel}, \binits{P.}}:
\batitle{Denoising diffusion probabilistic models}.
\bjtitle{Advances in Neural Information Processing Systems}
\bvolume{33},
\bfpage{6840}--\blpage{6851}
(\byear{2020})
\end{barticle}
\endbibitem

\bibitem[\protect\citeauthoryear{Dhariwal and
  Nichol}{2021}]{dhariwal2021diffusion}
\begin{barticle}
\bauthor{\bsnm{Dhariwal}, \binits{P.}},
\bauthor{\bsnm{Nichol}, \binits{A.}}:
\batitle{Diffusion models beat gans on image synthesis}.
\bjtitle{Advances in Neural Information Processing Systems}
\bvolume{34},
\bfpage{8780}--\blpage{8794}
(\byear{2021})
\end{barticle}
\endbibitem

\bibitem[\protect\citeauthoryear{Song and Ermon}{2020}]{song2020improved}
\begin{barticle}
\bauthor{\bsnm{Song}, \binits{Y.}},
\bauthor{\bsnm{Ermon}, \binits{S.}}:
\batitle{Improved techniques for training score-based generative models}.
\bjtitle{Advances in neural information processing systems}
\bvolume{33},
\bfpage{12438}--\blpage{12448}
(\byear{2020})
\end{barticle}
\endbibitem

\bibitem[\protect\citeauthoryear{Rombach et~al.}{2022}]{rombach2022high}
\begin{bchapter}
\bauthor{\bsnm{Rombach}, \binits{R.}},
\bauthor{\bsnm{Blattmann}, \binits{A.}},
\bauthor{\bsnm{Lorenz}, \binits{D.}},
\bauthor{\bsnm{Esser}, \binits{P.}},
\bauthor{\bsnm{Ommer}, \binits{B.}}:
\bctitle{High-resolution image synthesis with latent diffusion models}.
In: \bbtitle{Proceedings of the IEEE/CVF Conference on Computer Vision and
  Pattern Recognition},
pp. \bfpage{10684}--\blpage{10695}
(\byear{2022})
\end{bchapter}
\endbibitem

\bibitem[\protect\citeauthoryear{Song et~al.}{2020}]{song2020score}
\begin{botherref}
\oauthor{\bsnm{Song}, \binits{Y.}},
\oauthor{\bsnm{Sohl-Dickstein}, \binits{J.}},
\oauthor{\bsnm{Kingma}, \binits{D.P.}},
\oauthor{\bsnm{Kumar}, \binits{A.}},
\oauthor{\bsnm{Ermon}, \binits{S.}},
\oauthor{\bsnm{Poole}, \binits{B.}}:
Score-based generative modeling through stochastic differential equations.
arXiv preprint arXiv:2011.13456
(2020)
\end{botherref}
\endbibitem

\bibitem[\protect\citeauthoryear{Radford et~al.}{2021}]{radford2021learning}
\begin{botherref}
\oauthor{\bsnm{Radford}, \binits{A.}},
\oauthor{\bsnm{Kim}, \binits{J.W.}},
\oauthor{\bsnm{Hallacy}, \binits{C.}},
\oauthor{\bsnm{Ramesh}, \binits{A.}},
\oauthor{\bsnm{Goh}, \binits{G.}},
\oauthor{\bsnm{Agarwal}, \binits{S.}},
\oauthor{\bsnm{Sastry}, \binits{G.}},
\oauthor{\bsnm{Askell}, \binits{A.}},
\oauthor{\bsnm{Mishkin}, \binits{P.}},
\oauthor{\bsnm{Clark}, \binits{J.}}, et al.:
Learning transferable visual models from natural language supervision.
arXiv preprint arXiv:2103.00020
(2021)
\end{botherref}
\endbibitem

\bibitem[\protect\citeauthoryear{Vaswani}{2017}]{vaswani2017attention}
\begin{botherref}
\oauthor{\bsnm{Vaswani}, \binits{A.}}:
Attention is all you need.
Advances in Neural Information Processing Systems
(2017)
\end{botherref}
\endbibitem

\bibitem[\protect\citeauthoryear{Radford et~al.}{2019}]{radford2019language}
\begin{barticle}
\bauthor{\bsnm{Radford}, \binits{A.}},
\bauthor{\bsnm{Wu}, \binits{J.}},
\bauthor{\bsnm{Child}, \binits{R.}},
\bauthor{\bsnm{Luan}, \binits{D.}},
\bauthor{\bsnm{Amodei}, \binits{D.}},
\bauthor{\bsnm{Sutskever}, \binits{I.}}, \betal:
\batitle{Language models are unsupervised multitask learners}.
\bjtitle{OpenAI blog}
\bvolume{1}(\bissue{8}),
\bfpage{9}
(\byear{2019})
\end{barticle}
\endbibitem

\bibitem[\protect\citeauthoryear{Devlin}{2018}]{devlin2018bert}
\begin{botherref}
\oauthor{\bsnm{Devlin}, \binits{J.}}:
Bert: Pre-training of deep bidirectional transformers for language
  understanding.
arXiv preprint arXiv:1810.04805
(2018)
\end{botherref}
\endbibitem

\bibitem[\protect\citeauthoryear{Goodfellow et~al.}{2016}]{goodfellow2016deep}
\begin{bbook}
\bauthor{\bsnm{Goodfellow}, \binits{I.}},
\bauthor{\bsnm{Bengio}, \binits{Y.}},
\bauthor{\bsnm{Courville}, \binits{A.}}:
\bbtitle{Deep Learning}.
\bpublisher{MIT Press}, \blocation{???}
(\byear{2016})
\end{bbook}
\endbibitem

\bibitem[\protect\citeauthoryear{Choi et~al.}{2020}]{choi2020fine}
\begin{botherref}
\oauthor{\bsnm{Choi}, \binits{Y.}},
\oauthor{\bsnm{Uh}, \binits{Y.}},
\oauthor{\bsnm{Yoo}, \binits{J.}},
\oauthor{\bsnm{Ha}, \binits{J.-W.}},
\oauthor{\bsnm{Chun}, \binits{S.}}:
Fine-grained image-to-image transformation towards visual recognition.
Proceedings of the IEEE/CVF Conference on Computer Vision and Pattern
  Recognition (CVPR),
3626--3635
(2020)
\end{botherref}
\endbibitem

\bibitem[\protect\citeauthoryear{Sanh et~al.}{2020}]{sanh2019distilbert}
\begin{botherref}
\oauthor{\bsnm{Sanh}, \binits{V.}},
\oauthor{\bsnm{Debut}, \binits{L.}},
\oauthor{\bsnm{Chaumond}, \binits{J.}},
\oauthor{\bsnm{Wolf}, \binits{T.}}:
DistilBERT, a distilled version of BERT: smaller, faster, cheaper and lighter
(2020).
\url{https://arxiv.org/abs/1910.01108}
\end{botherref}
\endbibitem

\bibitem[\protect\citeauthoryear{Bromley et~al.}{1993}]{bromley1993signature}
\begin{botherref}
\oauthor{\bsnm{Bromley}, \binits{J.}},
\oauthor{\bsnm{Guyon}, \binits{I.}},
\oauthor{\bsnm{LeCun}, \binits{Y.}},
\oauthor{\bsnm{S{\"a}ckinger}, \binits{E.}},
\oauthor{\bsnm{Shah}, \binits{R.}}:
Signature verification using a" siamese" time delay neural network.
Advances in neural information processing systems
\textbf{6}
(1993)
\end{botherref}
\endbibitem

\bibitem[\protect\citeauthoryear{Hoffer and Ailon}{2015}]{hoffer2015deep}
\begin{bchapter}
\bauthor{\bsnm{Hoffer}, \binits{E.}},
\bauthor{\bsnm{Ailon}, \binits{N.}}:
\bctitle{Deep metric learning using triplet network}.
In: \bbtitle{International Workshop on Similarity-based Pattern Recognition},
pp. \bfpage{84}--\blpage{92}
(\byear{2015}).
\bcomment{Springer}
\end{bchapter}
\endbibitem

\bibitem[\protect\citeauthoryear{Cao et~al.}{2007}]{cao2007learning}
\begin{bchapter}
\bauthor{\bsnm{Cao}, \binits{Z.}},
\bauthor{\bsnm{Qin}, \binits{T.}},
\bauthor{\bsnm{Liu}, \binits{T.-Y.}},
\bauthor{\bsnm{Tsai}, \binits{M.-F.}},
\bauthor{\bsnm{Li}, \binits{H.}}:
\bctitle{Learning to rank: from pairwise approach to listwise approach}.
In: \bbtitle{Proceedings of the 24th International Conference on Machine
  Learning},
pp. \bfpage{129}--\blpage{136}
(\byear{2007})
\end{bchapter}
\endbibitem

\bibitem[\protect\citeauthoryear{Wang et~al.}{2014}]{wang2014learning}
\begin{bchapter}
\bauthor{\bsnm{Wang}, \binits{J.}},
\bauthor{\bsnm{Song}, \binits{Y.}},
\bauthor{\bsnm{Leung}, \binits{T.}},
\bauthor{\bsnm{Rosenberg}, \binits{C.}},
\bauthor{\bsnm{Wang}, \binits{J.}},
\bauthor{\bsnm{Philbin}, \binits{J.}},
\bauthor{\bsnm{Chen}, \binits{B.}},
\bauthor{\bsnm{Wu}, \binits{Y.}}:
\bctitle{Learning fine-grained image similarity with deep ranking}.
In: \bbtitle{Proceedings of the IEEE Conference on Computer Vision and Pattern
  Recognition},
pp. \bfpage{1386}--\blpage{1393}
(\byear{2014})
\end{bchapter}
\endbibitem

\bibitem[\protect\citeauthoryear{Chopra et~al.}{2005}]{chopra2005learning}
\begin{bchapter}
\bauthor{\bsnm{Chopra}, \binits{S.}},
\bauthor{\bsnm{Hadsell}, \binits{R.}},
\bauthor{\bsnm{LeCun}, \binits{Y.}}:
\bctitle{Learning a similarity metric discriminatively, with application to
  face verification}.
In: \bbtitle{2005 IEEE Computer Society Conference on Computer Vision and
  Pattern Recognition (CVPR'05)},
vol. \bseriesno{1},
pp. \bfpage{539}--\blpage{546}
(\byear{2005}).
\bcomment{IEEE}
\end{bchapter}
\endbibitem

\bibitem[\protect\citeauthoryear{Hadsell
  et~al.}{2006}]{hadsell2006dimensionality}
\begin{bchapter}
\bauthor{\bsnm{Hadsell}, \binits{R.}},
\bauthor{\bsnm{Chopra}, \binits{S.}},
\bauthor{\bsnm{LeCun}, \binits{Y.}}:
\bctitle{Dimensionality reduction by learning an invariant mapping}.
In: \bbtitle{2006 IEEE Computer Society Conference on Computer Vision and
  Pattern Recognition (CVPR'06)},
vol. \bseriesno{2},
pp. \bfpage{1735}--\blpage{1742}
(\byear{2006}).
\bcomment{IEEE}
\end{bchapter}
\endbibitem

\bibitem[\protect\citeauthoryear{Li et~al.}{2017}]{li2017improving}
\begin{bchapter}
\bauthor{\bsnm{Li}, \binits{Y.}},
\bauthor{\bsnm{Song}, \binits{Y.}},
\bauthor{\bsnm{Luo}, \binits{J.}}:
\bctitle{Improving pairwise ranking for multi-label image classification}.
In: \bbtitle{Proceedings of the IEEE Conference on Computer Vision and Pattern
  Recognition},
pp. \bfpage{3617}--\blpage{3625}
(\byear{2017})
\end{bchapter}
\endbibitem

\bibitem[\protect\citeauthoryear{Koch et~al.}{2015}]{koch2015siamese}
\begin{bchapter}
\bauthor{\bsnm{Koch}, \binits{G.}},
\bauthor{\bsnm{Zemel}, \binits{R.}},
\bauthor{\bsnm{Salakhutdinov}, \binits{R.}}, \betal:
\bctitle{Siamese neural networks for one-shot image recognition}.
In: \bbtitle{ICML Deep Learning Workshop},
vol. \bseriesno{2},
p. \bfpage{0}
(\byear{2015}).
\bcomment{Lille}
\end{bchapter}
\endbibitem

\bibitem[\protect\citeauthoryear{Chechik et~al.}{2010}]{chechik2010large}
\begin{botherref}
\oauthor{\bsnm{Chechik}, \binits{G.}},
\oauthor{\bsnm{Sharma}, \binits{V.}},
\oauthor{\bsnm{Shalit}, \binits{U.}},
\oauthor{\bsnm{Bengio}, \binits{S.}}:
Large scale online learning of image similarity through ranking.
Journal of Machine Learning Research
\textbf{11}(3)
(2010)
\end{botherref}
\endbibitem

\bibitem[\protect\citeauthoryear{Reimers and
  Gurevych}{2019}]{reimers2019sentence}
\begin{bchapter}
\bauthor{\bsnm{Reimers}, \binits{N.}},
\bauthor{\bsnm{Gurevych}, \binits{I.}}:
\bctitle{Sentence-bert: Sentence embeddings using siamese bert-networks}.
In: \bbtitle{Proceedings of the 2019 Conference on Empirical Methods in Natural
  Language Processing}.
\bpublisher{Association for Computational Linguistics},
\blocation{Darmstadt}
(\byear{2019}).
\burl{https://arxiv.org/abs/1908.10084}
\end{bchapter}
\endbibitem

\bibitem[\protect\citeauthoryear{Chen et~al.}{2020}]{chen2020simple}
\begin{bchapter}
\bauthor{\bsnm{Chen}, \binits{T.}},
\bauthor{\bsnm{Kornblith}, \binits{S.}},
\bauthor{\bsnm{Norouzi}, \binits{M.}},
\bauthor{\bsnm{Hinton}, \binits{G.}}:
\bctitle{A simple framework for contrastive learning of visual
  representations}.
In: \bbtitle{International Conference on Machine Learning},
pp. \bfpage{1597}--\blpage{1607}
(\byear{2020}).
\bcomment{PMLR}
\end{bchapter}
\endbibitem

\bibitem[\protect\citeauthoryear{Gao et~al.}{2021}]{gao2021simcse}
\begin{botherref}
\oauthor{\bsnm{Gao}, \binits{T.}},
\oauthor{\bsnm{Yao}, \binits{X.}},
\oauthor{\bsnm{Chen}, \binits{D.}}:
Simcse: Simple contrastive learning of sentence embeddings.
arXiv preprint arXiv:2104.08821
(2021)
\end{botherref}
\endbibitem

\bibitem[\protect\citeauthoryear{Burges}{2010}]{burges2010ranknet}
\begin{barticle}
\bauthor{\bsnm{Burges}, \binits{C.J.}}:
\batitle{From ranknet to lambdarank to lambdamart: An overview}.
\bjtitle{Learning}
\bvolume{11}(\bissue{23-581}),
\bfpage{81}
(\byear{2010})
\end{barticle}
\endbibitem

\bibitem[\protect\citeauthoryear{Burges et~al.}{2006}]{burges2006learning}
\begin{botherref}
\oauthor{\bsnm{Burges}, \binits{C.}},
\oauthor{\bsnm{Ragno}, \binits{R.}},
\oauthor{\bsnm{Le}, \binits{Q.}}:
Learning to rank with nonsmooth cost functions.
Advances in neural information processing systems
\textbf{19}
(2006)
\end{botherref}
\endbibitem

\bibitem[\protect\citeauthoryear{LeCun et~al.}{2006}]{lecun2006tutorial}
\begin{botherref}
\oauthor{\bsnm{LeCun}, \binits{Y.}},
\oauthor{\bsnm{Chopra}, \binits{S.}},
\oauthor{\bsnm{Hadsell}, \binits{R.}},
\oauthor{\bsnm{Ranzato}, \binits{M.}},
\oauthor{\bsnm{Huang}, \binits{F.}}:
A tutorial on energy-based learning.
Predicting structured data
\textbf{1}(0)
(2006)
\end{botherref}
\endbibitem

\bibitem[\protect\citeauthoryear{Friston et~al.}{2006}]{FRISTON200670}
\begin{barticle}
\bauthor{\bsnm{Friston}, \binits{K.}},
\bauthor{\bsnm{Kilner}, \binits{J.}},
\bauthor{\bsnm{Harrison}, \binits{L.}}:
\batitle{A free energy principle for the brain}.
\bjtitle{Journal of Physiology-Paris}
\bvolume{100}(\bissue{1}),
\bfpage{70}--\blpage{87}
(\byear{2006})
\doiurl{10.1016/j.jphysparis.2006.10.001} .
\bcomment{Theoretical and Computational Neuroscience: Understanding Brain
  Functions}
\end{barticle}
\endbibitem

\bibitem[\protect\citeauthoryear{Friston}{2009}]{Friston2009}
\begin{barticle}
\bauthor{\bsnm{Friston}, \binits{K.}}:
\batitle{The free-energy principle: a rough guide to the brain?}
\bjtitle{Trends in Cognitive Sciences}
\bvolume{13}(\bissue{7}),
\bfpage{293}--\blpage{301}
(\byear{2009})
\doiurl{10.1016/j.tics.2009.04.005}
\end{barticle}
\endbibitem

\bibitem[\protect\citeauthoryear{Finn et~al.}{2016}]{finn2016connection}
\begin{botherref}
\oauthor{\bsnm{Finn}, \binits{C.}},
\oauthor{\bsnm{Christiano}, \binits{P.}},
\oauthor{\bsnm{Abbeel}, \binits{P.}},
\oauthor{\bsnm{Levine}, \binits{S.}}:
A connection between generative adversarial networks, inverse reinforcement
  learning, and energy-based models.
arXiv preprint arXiv:1611.03852
(2016)
\end{botherref}
\endbibitem

\bibitem[\protect\citeauthoryear{Haarnoja
  et~al.}{2017}]{haarnoja2017reinforcement}
\begin{bchapter}
\bauthor{\bsnm{Haarnoja}, \binits{T.}},
\bauthor{\bsnm{Tang}, \binits{H.}},
\bauthor{\bsnm{Abbeel}, \binits{P.}},
\bauthor{\bsnm{Levine}, \binits{S.}}:
\bctitle{Reinforcement learning with deep energy-based policies}.
In: \bbtitle{International Conference on Machine Learning},
pp. \bfpage{1352}--\blpage{1361}
(\byear{2017}).
\bcomment{PMLR}
\end{bchapter}
\endbibitem

\bibitem[\protect\citeauthoryear{Grathwohl et~al.}{2019}]{grathwohl2019your}
\begin{botherref}
\oauthor{\bsnm{Grathwohl}, \binits{W.}},
\oauthor{\bsnm{Wang}, \binits{K.-C.}},
\oauthor{\bsnm{Jacobsen}, \binits{J.-H.}},
\oauthor{\bsnm{Duvenaud}, \binits{D.}},
\oauthor{\bsnm{Norouzi}, \binits{M.}},
\oauthor{\bsnm{Swersky}, \binits{K.}}:
Your classifier is secretly an energy based model and you should treat it like
  one.
arXiv preprint arXiv:1912.03263
(2019)
\end{botherref}
\endbibitem

\bibitem[\protect\citeauthoryear{Du et~al.}{2019}]{Du2019ModelBP}
\begin{botherref}
\oauthor{\bsnm{Du}, \binits{Y.}},
\oauthor{\bsnm{Lin}, \binits{T.}},
\oauthor{\bsnm{Mordatch}, \binits{I.}}:
Model based planning with energy based models.
ArXiv
\textbf{abs/1909.06878}
(2019)
\end{botherref}
\endbibitem

\bibitem[\protect\citeauthoryear{Osadchy et~al.}{2004}]{osadchy2004synergistic}
\begin{botherref}
\oauthor{\bsnm{Osadchy}, \binits{M.}},
\oauthor{\bsnm{Miller}, \binits{M.}},
\oauthor{\bsnm{Cun}, \binits{Y.}}:
Synergistic face detection and pose estimation with energy-based models.
Advances in neural information processing systems
\textbf{17}
(2004)
\end{botherref}
\endbibitem

\bibitem[\protect\citeauthoryear{Du and Mordatch}{2019}]{du2019implicit}
\begin{botherref}
\oauthor{\bsnm{Du}, \binits{Y.}},
\oauthor{\bsnm{Mordatch}, \binits{I.}}:
Implicit generation and modeling with energy based models.
Advances in Neural Information Processing Systems
\textbf{32}
(2019)
\end{botherref}
\endbibitem

\bibitem[\protect\citeauthoryear{Bengio}{2000}]{bengio2000gradient}
\begin{barticle}
\bauthor{\bsnm{Bengio}, \binits{Y.}}:
\batitle{Gradient-based optimization of hyperparameters}.
\bjtitle{Neural computation}
\bvolume{12}(\bissue{8}),
\bfpage{1889}--\blpage{1900}
(\byear{2000})
\end{barticle}
\endbibitem

\bibitem[\protect\citeauthoryear{Teh et~al.}{2003}]{teh2003energy}
\begin{barticle}
\bauthor{\bsnm{Teh}, \binits{Y.W.}},
\bauthor{\bsnm{Welling}, \binits{M.}},
\bauthor{\bsnm{Osindero}, \binits{S.}},
\bauthor{\bsnm{Hinton}, \binits{G.E.}}:
\batitle{Energy-based models for sparse overcomplete representations}.
\bjtitle{Journal of Machine Learning Research}
\bvolume{4}(\bissue{Dec}),
\bfpage{1235}--\blpage{1260}
(\byear{2003})
\end{barticle}
\endbibitem

\bibitem[\protect\citeauthoryear{Swersky
  et~al.}{2011}]{swersky2011autoencoders}
\begin{bchapter}
\bauthor{\bsnm{Swersky}, \binits{K.}},
\bauthor{\bsnm{Ranzato}, \binits{M.}},
\bauthor{\bsnm{Buchman}, \binits{D.}},
\bauthor{\bsnm{Freitas}, \binits{N.D.}},
\bauthor{\bsnm{Marlin}, \binits{B.M.}}:
\bctitle{On autoencoders and score matching for energy based models}.
In: \bbtitle{Proceedings of the 28th International Conference on Machine
  Learning (ICML-11)},
pp. \bfpage{1201}--\blpage{1208}
(\byear{2011})
\end{bchapter}
\endbibitem

\bibitem[\protect\citeauthoryear{Zhai et~al.}{2016}]{zhai2016deep}
\begin{bchapter}
\bauthor{\bsnm{Zhai}, \binits{S.}},
\bauthor{\bsnm{Cheng}, \binits{Y.}},
\bauthor{\bsnm{Lu}, \binits{W.}},
\bauthor{\bsnm{Zhang}, \binits{Z.}}:
\bctitle{Deep structured energy based models for anomaly detection}.
In: \bbtitle{International Conference on Machine Learning},
pp. \bfpage{1100}--\blpage{1109}
(\byear{2016}).
\bcomment{PMLR}
\end{bchapter}
\endbibitem

\bibitem[\protect\citeauthoryear{Kumar et~al.}{2019}]{kumar2019maximum}
\begin{botherref}
\oauthor{\bsnm{Kumar}, \binits{R.}},
\oauthor{\bsnm{Ozair}, \binits{S.}},
\oauthor{\bsnm{Goyal}, \binits{A.}},
\oauthor{\bsnm{Courville}, \binits{A.}},
\oauthor{\bsnm{Bengio}, \binits{Y.}}:
Maximum entropy generators for energy-based models.
arXiv preprint arXiv:1901.08508
(2019)
\end{botherref}
\endbibitem

\bibitem[\protect\citeauthoryear{Song and Kingma}{2021}]{song2021train}
\begin{botherref}
\oauthor{\bsnm{Song}, \binits{Y.}},
\oauthor{\bsnm{Kingma}, \binits{D.P.}}:
How to train your energy-based models.
arXiv preprint arXiv:2101.03288
(2021)
\end{botherref}
\endbibitem

\bibitem[\protect\citeauthoryear{Vapnik}{1999}]{vapnik1999nature}
\begin{bbook}
\bauthor{\bsnm{Vapnik}, \binits{V.}}:
\bbtitle{The Nature of Statistical Learning Theory}.
\bpublisher{Springer},
\blocation{Berlin}
(\byear{1999})
\end{bbook}
\endbibitem

\bibitem[\protect\citeauthoryear{LeCun et~al.}{1998}]{lecun1998gradient}
\begin{barticle}
\bauthor{\bsnm{LeCun}, \binits{Y.}},
\bauthor{\bsnm{Bottou}, \binits{L.}},
\bauthor{\bsnm{Bengio}, \binits{Y.}},
\bauthor{\bsnm{Haffner}, \binits{P.}}:
\batitle{Gradient-based learning applied to document recognition}.
\bjtitle{Proceedings of the IEEE}
\bvolume{86}(\bissue{11}),
\bfpage{2278}--\blpage{2324}
(\byear{1998})
\end{barticle}
\endbibitem

\bibitem[\protect\citeauthoryear{Collins}{2002}]{collins2002discriminative}
\begin{bchapter}
\bauthor{\bsnm{Collins}, \binits{M.}}:
\bctitle{Discriminative training methods for hidden markov models: Theory and
  experiments with perceptron algorithms}.
In: \bbtitle{Proceedings of the 2002 Conference on Empirical Methods in Natural
  Language Processing (EMNLP 2002)},
pp. \bfpage{1}--\blpage{8}
(\byear{2002})
\end{bchapter}
\endbibitem

\bibitem[\protect\citeauthoryear{Shapiro}{2003}]{shapiro2003monte}
\begin{barticle}
\bauthor{\bsnm{Shapiro}, \binits{A.}}:
\batitle{Monte carlo sampling methods}.
\bjtitle{Handbooks in operations research and management science}
\bvolume{10},
\bfpage{353}--\blpage{425}
(\byear{2003})
\end{barticle}
\endbibitem

\bibitem[\protect\citeauthoryear{Jordan et~al.}{1999}]{jordan1999introduction}
\begin{barticle}
\bauthor{\bsnm{Jordan}, \binits{M.I.}},
\bauthor{\bsnm{Ghahramani}, \binits{Z.}},
\bauthor{\bsnm{Jaakkola}, \binits{T.S.}},
\bauthor{\bsnm{Saul}, \binits{L.K.}}:
\batitle{An introduction to variational methods for graphical models}.
\bjtitle{Machine learning}
\bvolume{37}(\bissue{2}),
\bfpage{183}--\blpage{233}
(\byear{1999})
\end{barticle}
\endbibitem

\bibitem[\protect\citeauthoryear{LeCun and Huang}{2005}]{lecun2005loss}
\begin{bchapter}
\bauthor{\bsnm{LeCun}, \binits{Y.}},
\bauthor{\bsnm{Huang}, \binits{F.J.}}:
\bctitle{Loss functions for discriminative training of energy-based models}.
In: \bbtitle{International Workshop on Artificial Intelligence and Statistics},
pp. \bfpage{206}--\blpage{213}
(\byear{2005}).
\bcomment{PMLR}
\end{bchapter}
\endbibitem

\bibitem[\protect\citeauthoryear{Levin and
  Fleisher}{1988}]{levin1988accelerated}
\begin{barticle}
\bauthor{\bsnm{Levin}, \binits{E.}},
\bauthor{\bsnm{Fleisher}, \binits{M.}}:
\batitle{Accelerated learning in layered neural networks}.
\bjtitle{Complex systems}
\bvolume{2}(\bissue{625-640}),
\bfpage{3}
(\byear{1988})
\end{barticle}
\endbibitem

\bibitem[\protect\citeauthoryear{Bengio et~al.}{2000}]{bengio2000neural}
\begin{botherref}
\oauthor{\bsnm{Bengio}, \binits{Y.}},
\oauthor{\bsnm{Ducharme}, \binits{R.}},
\oauthor{\bsnm{Vincent}, \binits{P.}}:
A neural probabilistic language model.
Advances in neural information processing systems
\textbf{13}
(2000)
\end{botherref}
\endbibitem

\bibitem[\protect\citeauthoryear{Bengio et~al.}{1992}]{bengio1992global}
\begin{barticle}
\bauthor{\bsnm{Bengio}, \binits{Y.}},
\bauthor{\bsnm{De~Mori}, \binits{R.}},
\bauthor{\bsnm{Flammia}, \binits{G.}},
\bauthor{\bsnm{Kompe}, \binits{R.}}:
\batitle{Global optimization of a neural network-hidden markov model hybrid}.
\bjtitle{IEEE transactions on Neural Networks}
\bvolume{3}(\bissue{2}),
\bfpage{252}--\blpage{259}
(\byear{1992})
\end{barticle}
\endbibitem

\bibitem[\protect\citeauthoryear{Taskar et~al.}{2003}]{taskar2003max}
\begin{botherref}
\oauthor{\bsnm{Taskar}, \binits{B.}},
\oauthor{\bsnm{Guestrin}, \binits{C.}},
\oauthor{\bsnm{Koller}, \binits{D.}}:
Max-margin markov networks.
Advances in neural information processing systems
\textbf{16}
(2003)
\end{botherref}
\endbibitem

\bibitem[\protect\citeauthoryear{Altun et~al.}{2003}]{altun2003hidden}
\begin{bchapter}
\bauthor{\bsnm{Altun}, \binits{Y.}},
\bauthor{\bsnm{Tsochantaridis}, \binits{I.}},
\bauthor{\bsnm{Hofmann}, \binits{T.}}:
\bctitle{Hidden markov support vector machines}.
In: \bbtitle{Proceedings of the 20th International Conference on Machine
  Learning (ICML-03)},
pp. \bfpage{3}--\blpage{10}
(\byear{2003})
\end{bchapter}
\endbibitem

\bibitem[\protect\citeauthoryear{Kleinbaum
  et~al.}{2002}]{kleinbaum2002logistic}
\begin{bbook}
\bauthor{\bsnm{Kleinbaum}, \binits{D.G.}},
\bauthor{\bsnm{Dietz}, \binits{K.}},
\bauthor{\bsnm{Gail}, \binits{M.}},
\bauthor{\bsnm{Klein}, \binits{M.}},
\bauthor{\bsnm{Klein}, \binits{M.}}:
\bbtitle{Logistic Regression}.
\bpublisher{Springer},
\blocation{Berlin}
(\byear{2002})
\end{bbook}
\endbibitem

\bibitem[\protect\citeauthoryear{Juang et~al.}{1997}]{juang1997minimum}
\begin{barticle}
\bauthor{\bsnm{Juang}, \binits{B.-H.}},
\bauthor{\bsnm{Hou}, \binits{W.}},
\bauthor{\bsnm{Lee}, \binits{C.-H.}}:
\batitle{Minimum classification error rate methods for speech recognition}.
\bjtitle{IEEE Transactions on Speech and Audio processing}
\bvolume{5}(\bissue{3}),
\bfpage{257}--\blpage{265}
(\byear{1997})
\end{barticle}
\endbibitem

\end{thebibliography}
